\documentclass[acmsmall]{acmart}
\AtBeginDocument{%
  \providecommand\BibTeX{{%
    \normalfont B\kern-0.5em{\scshape i\kern-0.25em b}\kern-0.8em\TeX}}}

\setcopyright{acmcopyright}
\copyrightyear{2022}
\acmYear{2022}
\acmDOI{10.1145/1122445.1122456}

\acmJournal{JACM}
\acmVolume{0}
\acmNumber{0}
\acmArticle{000}
\acmMonth{0}

\usepackage{caption}
\usepackage{subcaption}
\usepackage{booktabs}
\usepackage{hyperref}
\usepackage{cleveref}
\usepackage[normalem]{ulem}
\DeclareMathOperator{\typeaconf}{type1conf}
\DeclareMathOperator{\typebconf}{type2conf}
\DeclareMathOperator{\mean}{mean}
\DeclareMathOperator{\prob}{P}
\usepackage{xspace}
\newcommand{\rom}[1]{\uppercase\expandafter{\romannumeral #1\relax}}

\newcommand{\etal}{\hbox{\emph{et al.}}\xspace}
\newcommand{\eg}{\hbox{\emph{e.g.,}}\xspace}
\newcommand{\ie}{\hbox{\emph{i.e.}}\xspace}

\newcommand{\wrt}{\hbox{\emph{w.r.t.}}\xspace}

\newcommand{\etc}{\hbox{\emph{etc.}}\xspace}
\usepackage{framed}

\newcommand{\msout}[1]{\text{\sout{\ensuremath{#1}}}}

\begin{document}

\newcommand{\yuchi}[1]{{\scriptsize \todo{Yuchi:  {\Red #1}}}}
\newcommand{\bray}[1]{{\scriptsize \todo{Bray:  {\Red #1}}}}
\newcommand{\vicente}[1]{{\scriptsize \todo{Vicente:  {\green{#1}}}}}
\newcommand{\zz}[1]{{\scriptsize \todo{Ziyuan:  {\orange #1}}}}
\newcommand{\conor}[1]{{\scriptsize \todo{Conor:  {\blue{ #1}}}}}

\newcommand{\new}[1]{\red {#1}}

\renewcommand{\new}[1]{\textcolor{black}{{#1}}}
\renewcommand{\sout}[1]{}

\newcommand{\positive}{left (+ve)~}
\newcommand{\negative}{right (-ve)~}

\newcommand{\iorg}{$I_{o}$}
\newcommand{\itrans}{$I_{t}$}
\newcommand{\iorgo}{$\theta_{o}$}
\newcommand{\itranso}{$\theta_{t}$}
\newcommand{\isusp}{$I_{err}$}
\newcommand{\iorig}{$I_{org}$}

\newcommand{\Comment}[1]{}

\newcommand{\tool}{WR\xspace}
\newcommand{\ta}{Type1 confusions\xspace}
\newcommand{\tb}{Type2 confusions\xspace}
\newcommand{\navd}{\textsc{napvd}\xspace}
\newcommand{\npam}{Neuron Activation Probability Matrix\xspace}
\newcommand{\napm}{Neuron Activation Probability Matrix\xspace}
\newcommand{\nasm}{Neuron Activation Sensitivity Matrix\xspace}

\newcommand{\coco}{COCO~}
\newcommand{\cifars}{CIFAR-10~}
\newcommand{\cifar}{CIFAR-100~}

\newcommand{\ddata}{experimental data\xspace}


\newcommand{\bias}{\textrm{bias}}
\newcommand{\avgbias}{\textrm{average bias}}
\newcommand{\probability}{\textrm{probability}}
\newcommand{\object}{\textrm{object}}
\newcommand{\activates}{\textrm{activates}}
\newcommand{\distance}{\textrm{distance}}
\newcommand{\fairness}{\textrm{fairness}}
\newcommand{\ab}{\textrm{avg\_bias}\xspace}
\newcommand{\cd}{\textrm{cd}\xspace}
\newcommand{\acd}{\textrm{avg\_cd}\xspace}
\newcommand{\error}{\textrm{error}}
\newcommand{\logic}{semantics~}
\newcommand{\Logic}{Semantics~}

\newcommand{\eqnum}{\refstepcounter{equation}\textup{\tagform({\theequation})}}
\makeatother

%
\newcounter{RQCounter}
\newcounter{RQACounter}

%

\newcommand{\RQ}[2]{%
\refstepcounter{RQCounter} \label{#1}
 \noindent
 \textbf{RQ\arabic{RQCounter}.~#2}
}

\newcommand{\RQA}[2]{%
\refstepcounter{RQACounter} \label{#1}
\vspace{0.1in} \noindent\textbf{RQ\arabic{RQACounter}.~#2 \vspace{0.05in}}

}

\newcommand{\RQBoxed}[2]{%
\refstepcounter{RQCounter} \label{#1}
\begin{framed}%
\filbreak
\vspace{0.001in}
	\noindent\textbf{RQ\arabic{RQCounter}: }%
#2\end{framed}
}

%
\newcommand{\RS}[2]{%
\begin{framed}%
\filbreak
\textbf{Result {\ref{#1}}:~}{\emph {#2}}%
\end{framed}
}
\newcommand\at{@}
\newcommand\lb{[}
\newcommand\rb{]}
\newcommand\Red[1]{\textcolor[rgb]{1.00,0.00,0.00}{\textbf{#1}}}
\newcommand\red[1]{\textcolor[rgb]{1.00,0.00,0.00}{#1}}
\newcommand\grayline[1]{\textcolor[rgb]{0.4,0.4,0.4}{{#1}}}
\newcommand\gray[1]{\colorbox[rgb]{0.8,0.8,0.8}{{#1}}}
\newcommand\brown[1]{\textcolor[rgb]{0.55,0.27,0.07}{#1}}
\newcommand\orange[1]{\textcolor[rgb]{1.00,0.55,0.30}{\textbf{#1}}}
\newcommand\lime[1]{\textcolor[rgb]{0.13,0.55,0.13}{{#1}}}
\newcommand\blue[1]{\textcolor[rgb]{0.00,0.00,1.00}{{#1}}}
\newcommand\ColorLine[1]{\textcolor[rgb]{0.00,0.00,1.00}{{#1}}}
\newcommand\javagreen[1]{\textcolor[rgb]{0.25,0.5,0.35}{{#1}}}
\newcommand\javadocblue[1]{\textcolor[rgb]{0.25,0.35,0.75}{{#1}}}
\newcommand\dkgreen[1]{\textcolor[rgb]{0.0,0.6,0}{\textbf{#1}}}
\newcommand\green[1]{\textcolor[rgb]{0.0,0.6,0}{#1}}
\newcommand\mauve[1]{\textcolor[rgb]{0.58,0.0,0.82}{{#1}}}
\newcommand\yellow[1]{\textcolor[rgb]{1,0.9,0.5}{{\bf #1}}}
\newcommand\lblue[1]{\colorbox[rgb]{0.87,1,1}{{#1}}}

\newcommand\PYbg[1]{\textcolor[rgb]{0.00,0.50,0.00}{\textbf{#1}}}
\newcommand\PYbf[1]{\textcolor[rgb]{0.73,0.40,0.53}{\textbf{#1}}}
\newcommand\PYbe[1]{\textcolor[rgb]{0.40,0.40,0.40}{#1}}
\newcommand\PYbd[1]{\textcolor[rgb]{0.73,0.13,0.13}{#1}}
\newcommand\PYbc[1]{\textcolor[rgb]{0.00,0.50,0.00}{\textbf{#1}}}
\newcommand\PYbb[1]{\textcolor[rgb]{0.40,0.40,0.40}{#1}}
\newcommand\PYba[1]{\textcolor[rgb]{0.00,0.00,0.50}{\textbf{#1}}}
\newcommand\PYaJ[1]{\textcolor[rgb]{0.73,0.13,0.13}{#1}}
\newcommand\PYaK[1]{\textcolor[rgb]{0.00,0.00,1.00}{\textbf{#1}}}
\newcommand\PYaH[1]{\fcolorbox[rgb]{1.00,0.00,0.00}{1,1,1}{#1}}
\newcommand\PYaI[1]{\textcolor[rgb]{0.69,0.00,0.25}{#1}}
\newcommand\PYaN[1]{\textcolor[rgb]{0.00,0.00,1.00}{\textbf{#1}}}
\newcommand\PYaO[1]{\textcolor[rgb]{0.00,0.00,0.50}{\textbf{#1}}}
\newcommand\PYaL[1]{\textcolor[rgb]{0.73,0.73,0.73}{#1}}
\newcommand\PYaM[1]{\textcolor[rgb]{0.74,0.48,0.00}{#1}}
\newcommand\PYaB[1]{\textcolor[rgb]{0.00,0.25,0.82}{#1}}
\newcommand\PYaC[1]{\textcolor[rgb]{0.67,0.13,1.00}{#1}}
\newcommand\PYaA[1]{\textcolor[rgb]{0.00,0.50,0.00}{#1}}
\newcommand\PYaF[1]{\textcolor[rgb]{1.00,0.00,0.00}{#1}}
\newcommand\PYaG[1]{\textcolor[rgb]{0.10,0.09,0.49}{#1}}
\newcommand\PYaD[1]{\textcolor[rgb]{0.25,0.50,0.50}{\textit{#1}}}
\newcommand\PYaE[1]{\textcolor[rgb]{0.63,0.00,0.00}{#1}}
\newcommand\PYaZ[1]{\textcolor[rgb]{0.00,0.50,0.00}{\textbf{#1}}}
\newcommand\PYaX[1]{\textcolor[rgb]{0.00,0.50,0.00}{#1}}
\newcommand\PYaY[1]{\textcolor[rgb]{0.73,0.13,0.13}{#1}}
\newcommand\PYaR[1]{\textcolor[rgb]{0.10,0.09,0.49}{#1}}
\newcommand\PYaS[1]{\textcolor[rgb]{0.25,0.50,0.50}{\textit{#1}}}
\newcommand\PYaP[1]{\textcolor[rgb]{0.49,0.56,0.16}{#1}}
\newcommand\PYaQ[1]{\textcolor[rgb]{0.40,0.40,0.40}{#1}}
\newcommand\PYaV[1]{\textcolor[rgb]{0.00,0.00,1.00}{\textbf{#1}}}
\newcommand\PYaW[1]{\textcolor[rgb]{0.73,0.13,0.13}{#1}}
\newcommand\PYaT[1]{\textcolor[rgb]{0.50,0.00,0.50}{\textbf{#1}}}
\newcommand\PYaU[1]{\textcolor[rgb]{0.82,0.25,0.23}{\textbf{#1}}}
\newcommand\PYaj[1]{\textcolor[rgb]{0.00,0.50,0.00}{#1}}
\newcommand\PYak[1]{\textcolor[rgb]{0.73,0.40,0.53}{#1}}
\newcommand\PYah[1]{\textcolor[rgb]{0.63,0.63,0.00}{#1}}
\newcommand\PYai[1]{\textcolor[rgb]{0.10,0.09,0.49}{#1}}
\newcommand\PYan[1]{\textcolor[rgb]{0.67,0.13,1.00}{\textbf{#1}}}
\newcommand\PYao[1]{\textcolor[rgb]{0.73,0.40,0.13}{\textbf{#1}}}
\newcommand\PYal[1]{\textcolor[rgb]{0.25,0.50,0.50}{\textit{#1}}}
\newcommand\PYam[1]{\textbf{#1}}
\newcommand\PYab[1]{\textit{#1}}
\newcommand\PYac[1]{\textcolor[rgb]{0.73,0.13,0.13}{#1}}
\newcommand\PYaa[1]{\textcolor[rgb]{0.50,0.50,0.50}{#1}}
\newcommand\PYaf[1]{\textcolor[rgb]{0.25,0.50,0.50}{\textit{#1}}}
\newcommand\PYag[1]{\textcolor[rgb]{0.40,0.40,0.40}{#1}}
\newcommand\PYad[1]{\textcolor[rgb]{0.73,0.13,0.13}{#1}}
\newcommand\PYae[1]{\textcolor[rgb]{0.40,0.40,0.40}{#1}}
\newcommand\PYaz[1]{\textcolor[rgb]{0.00,0.63,0.00}{#1}}
\newcommand\PYax[1]{\textcolor[rgb]{0.60,0.60,0.60}{\textbf{#1}}}
\newcommand\PYay[1]{\textcolor[rgb]{0.00,0.50,0.00}{\textbf{#1}}}
\newcommand\PYar[1]{\textcolor[rgb]{0.10,0.09,0.49}{#1}}
\newcommand\PYas[1]{\textcolor[rgb]{0.73,0.13,0.13}{\textit{#1}}}
\newcommand\PYap[1]{\textcolor[rgb]{0.00,0.50,0.00}{#1}}
\newcommand\PYaq[1]{\textcolor[rgb]{0.53,0.00,0.00}{#1}}
\newcommand\PYav[1]{\textcolor[rgb]{0.00,0.50,0.00}{\textbf{#1}}}
\newcommand\PYaw[1]{\textcolor[rgb]{0.40,0.40,0.40}{#1}}
\newcommand\PYat[1]{\textcolor[rgb]{0.10,0.09,0.49}{#1}}
\newcommand\PYau[1]{\textcolor[rgb]{0.40,0.40,0.40}{#1}}

\title{Repairing Group-Level Errors for DNNs Using Weighted Regularization}

\author{Ziyuan Zhong}
\authornote{Both authors contributed equally to this research.}
\email{ziyuan.zhong@columbia.edu}
\author{Yuchi Tian}
\authornotemark[1]
\email{yt2667@columbia.edu}
\affiliation{%
  \institution{Columbia University}
  \streetaddress{116th and Broadway}
  \city{New York}
  \state{NY}
  \country{USA}
  \postcode{10027}
}

\author{Conor J.Sweeney}
\email{cjs2201@columbia.edu}
\affiliation{%
  \institution{Columbia University}
  \streetaddress{116th and Broadway}
  \city{New York}
  \state{NY}
  \country{USA}
  \postcode{10027}
}

\author{Vicente Ordonez}
\email{vicenteor@rice.edu}
\affiliation{%
  \institution{Rice University}
  \streetaddress{6100 Main St}
  \city{Houston}
  \state{TX}
  \country{USA}
  \postcode{77005}
}

\author{Baishakhi Ray}
\email{rayb@cs.columbia.edu}
\affiliation{%
  \institution{Columbia University}
  \streetaddress{116th and Broadway}
  \city{New York}
  \state{NY}
  \country{USA}
  \postcode{10027}
}

\renewcommand{\shortauthors}{Zhong and Tian, et al.}

\begin{abstract}

Deep Neural Networks (DNNs) have been widely used in software making decisions impacting people's lives. However, they have been found to exhibit severe erroneous behaviors that may lead to unfortunate outcomes. Previous work shows that such misbehaviors often occur due to class property violations rather than errors on a single image. Although methods for detecting such errors have been proposed, fixing them has not been studied so far. Here, we propose a generic method called Weighted Regularization (WR) consisting of five concrete methods targeting the error-producing classes to fix the DNNs. In particular, it can repair confusion error and bias error of DNN models for both single-label and multi-label image classifications. A confusion error happens when a given DNN model tends to confuse between two classes. Each method in WR assigns more weights at a stage of DNN retraining or inference to mitigate the confusion between target pair. A bias error can be fixed \new{similarly}\sout{in a similar way}. We evaluate and compare the proposed methods \new{along with baselines} on \new{six}\sout{five} widely-used datasets and architecture combinations. The results suggest that WR methods have different trade-offs but under each setting at least one WR method can greatly reduce confusion/bias errors \new{at a very limited cost of} \sout{and improve} the overall performance. 
\end{abstract}




\maketitle

\section{Introduction}

Deep Neural Networks are widely used nowadays as components in many critical applications like self-driving cars, face-recognition, medical diagnosis, \etc 
\sout{Unlike traditional software, although a DNN model has no code logics, it may still suffer from a different type of "serious bugs"}
\new{"Although DNN models do not contain any code logic like more traditional forms of software engineering, these models can still suffer from a different form of "serious bugs"}\cite{repairing_DNNoverview, tian2019deepinspect}.
For example, it has been found that Google photo-tagging app tagged pictures of two dark-skinned people as ``gorillas''~\cite{google_tag}. Analogous to traditional software bugs, previous work in Software Engineering (SE) has denoted classification errors like this as {\em model bugs}~\cite{mode2018} that arise from either biased training data, problematic model architecture, training procedure error or the combination of them.

DNN classification errors fall into three main categories, instance-wise, group-wise, and dataset-wise. Instance-wise error has been well studied in the previous literature. In essence, an instance-wise error happens when a DNN model misclassifies different semantic-preserving transformations of a given input \cite{pei2017deepxplore, tian2017deeptest, deeprobust, madry2018}. Over the years, researchers have found numerous such transformations such as norm-bounded perturbation\cite{madry2018}, natural transformation\cite{engstrom2019exploring}, or physical attack\cite{evtimov2017robust} to fool a well-trained DNN classifier. The fixing strategies such as adversarial training, data augmentation are also widely studied \cite{madry2018, engstrom2019exploring, gao2020fuzz}. Dataset-wise error is essentially the model's overall accuracy being worse than expected. Previous work have shown promising fixing results using strategies like data augmentation \cite{ren20}, weight adaptation \cite{zhang19} and input selection using differential heat maps \cite{mode2018}.

Different from the above two, group-wise error is about the DNN model's weak performance on differentiating among certain classes or has inconsistent performance across classes\cite{tian2019deepinspect}. There are very few work on repairing group-wise errors and it only receives attentions recently \cite{tian2019deepinspect}. This type of bugs is very concerning since it has been found to relate to many real-world notorious errors. Some work have proposed techniques to detect this kind of errors, however, until now, no fixing methods have been proposed for repairing them. To bridge this gap, in this work, we propose a generic fixing method for repairing such errors of any given DNN models.

\begin{figure}[ht]
\centering
\begin{subfigure}[b]{0.36\linewidth}
    \includegraphics[width=\linewidth]{img/intro/confusion_laptop_mouse.jpg}
    \caption{given laptop, a mouse is predicted to be present}
\end{subfigure}
\begin{subfigure}[b]{0.36\linewidth}
    \includegraphics[width=\linewidth]{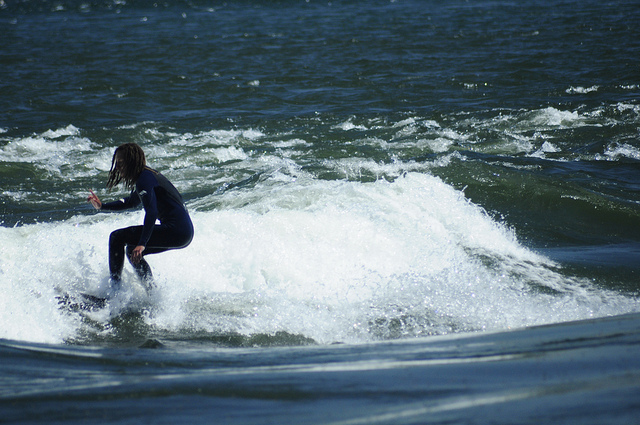}
    \caption{a surfing woman is misclassied as man\label{fig:surfing_woman}}
\end{subfigure}
\caption{\small{\textbf{Examples of confusion and bias errors found in \cite{tian2019deepinspect}}}}
\label{fig:error_examples}
\vspace{-2mm}
\end{figure}

The group-wise errors definition proposed in \cite{tian2019deepinspect} consists of two main types with different root causes: \emph{(i) \textbf{Confusion}: The model cannot differentiate one class from another. For example, Google Photos confuses skier and mountain~\cite{google_photo}.  (ii) \textbf{Bias}: The model shows disparate outcomes between two related groups. For example, Zhao \etal \cite{zhao2017men} find classification bias in favor of women on activities like shopping, cooking, washing, \etc.} \Cref{fig:error_examples} presents two concrete examples of both types of errors from \coco and Image-Net reported in \cite{tian2019deepinspect}. Note that unlike an instance-wise error which affects a particular misclassified image or a dataset-wise error which affects all the misclassified images in the dataset, such group-wise error affects all the images falling into the groups.

\begin{figure}[ht]
\centering
\includegraphics[width=0.48\linewidth]{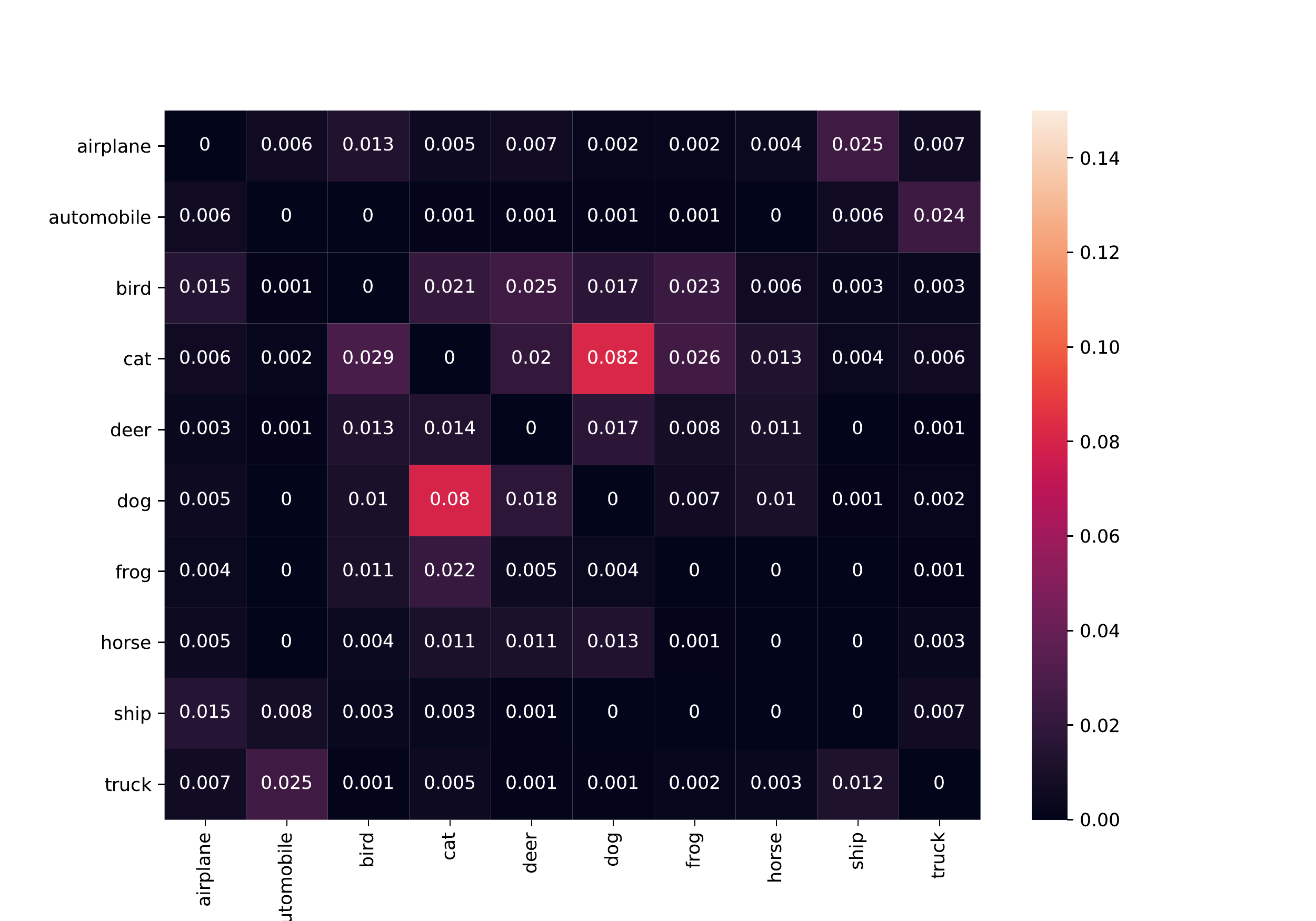}
\caption{\small{\textbf{Confusion matrix of VGG11-BN on \cifars.}}}
\label{fig:confusion_matrix_orig}
\vspace{-2mm}
\end{figure}

The causes of a group-wise error can be that certain classes are harder to be differentiated from each other. For example, in \cifars, dog and cat tend to confuse even a state-of-the-art DNN model \sout{since they share many common semantic features}. \new{\Cref{fig:confusion_matrix_orig} shows the confusion matrix for a well-trained VGG11-BN model on \cifars. The (i,j)-th entry is the probability of the class i being predicted to class j by mistake. It can be seen that the model is much more likely to confuse between dog and cat with a confusion of 0.081 meaning given a uniformly randomly sampled dog or cat image there is a 8.1\% chance for the model to make a mistake by predicting the image to belong to the other class. This is much larger than 0.025 which is the second largest pairwise confusion. One potential cause of the model's high confusion between dog and cat is that these two classes share many common semantic features.} As a result, the two classes tend to be very close to each other in the representation space and the decision boundary between them might not be "fine-grained" enough for correct classification on some dog and cat images. We denote the error-inducing classes as target classes. To fix the errors of the target classes, the model needs to take more effort to learn from them. \new{Note that this is similar to solving the problem of class imbalance\cite{oversampling18, oversampling19} in which case a given model is supposed to take more effort on some minority classes in order to achieve improved class-balanced error. The key differences are two-fold. First, group-wise errors are not necessarily caused by class imbalance. As shown in the example of \cifars in \Cref{fig:confusion_matrix_orig}, although each class in the training data has the same number of samples, the model still suffer from much higher confusion between dog and cat compared with other pairs. Second, the two share different optimization objectives. The objective in our setting is mitigating confusion/bias while maintaining overall accuracy rather than improving class-balanced accuracy. It is possible for a model to have similar class-wise accuracy for each class but still suffer from high pair-wise confusion between some pairs of classes.}

For large and complex DNN models, complete training from scratch may not be possible. Sometimes no extra data can be collected, either. In these cases, fine-tuning with potential data augmentation can be applied to enforce the model to learn to better classify the target classes. When fine-tuning is not possible or training data cannot be accessed, (for example, the user does not have right to access the data) the output can be modified to fix the errors while sacrificing the overall performance to some extent.

\begin{figure}[ht]
\centering

\includegraphics[width=0.8\linewidth]{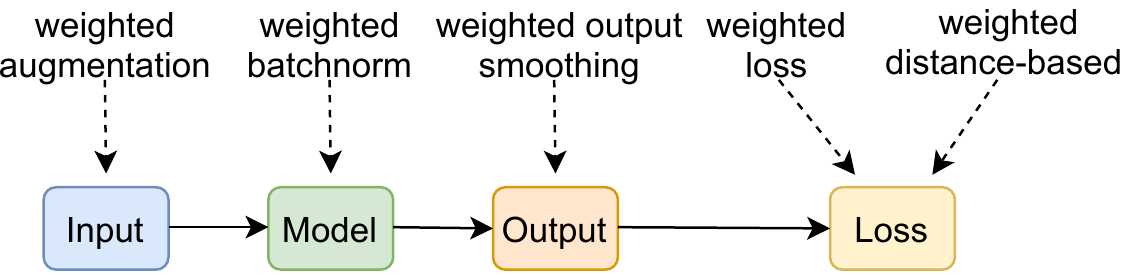}

\caption{\small{\textbf{Overview of Weighted Regularization for Target Fixing}
}}
\label{fig:overview}
\vspace{-2mm}
\end{figure} 

Based on these observations \new{as well as inspiration from existing works on the problem of class imbalance}, we propose a generic method called weighted regularization (\tool). \tool consists of multiple concrete methods including weighted augmentation (w-aug), weighted batch normalization (w-bn), weighted output smoothing (w-os), weighted loss (w-loss), and weighted  distance-based regularization  (w-dbr). These methods function at different stages of a given DNN's training or inference time \Cref{fig:overview}. In particular, if retraining is allowed and training data are provided, w-aug assigns more weights to the target classes during the retraining, w-bn shifts the distribution of the activation values induced by the input at every batchnorm layer(assuming the model has batchnorm layers), and w-loss and w-dbr modify the loss function by assigning more weights to the erroneous instances and regularizing the class centroids in the representation space, respectively. Such fine-tuning strategies enable the model to emphasize more on the instances of the target classes and thus more likely to avoid the errors involving the target classes. If fine-tuning and training data are not provided, w-os multiplies the model's prediction on target classes by a smaller user-specified constant. In other words, making the model predict less the target class. In this way, the group-wise errors on those unsure data points can be avoided.

\begin{figure}[ht]
\centering

\includegraphics[width=\linewidth]{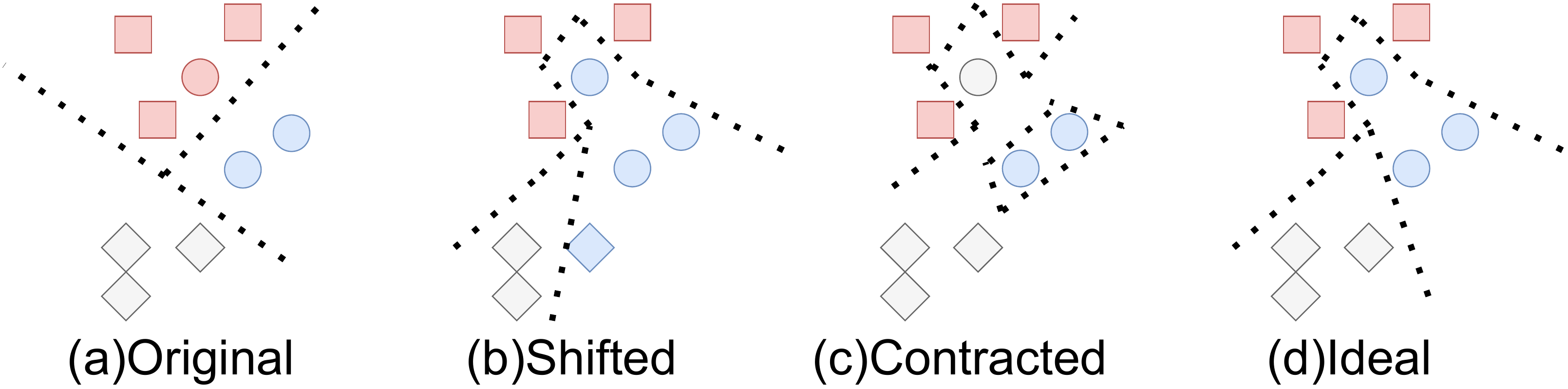}
\caption{\small{\textbf{Illustration of different potential decision boundary before and after applying \tool.}
}}
\label{fig:demonstration}
\vspace{-2mm}
\end{figure}

We illustrate an ideal fixing result and potential fixing results after applying these methods in \Cref{fig:demonstration} with an example consists of three classes (square, circle and diamond). The colors represent the model's prediction while the dashed lines denote the model's decision boundary. \Cref{fig:demonstration}(a) shows that the original model tends to confuse between square and circle since these two classes are very close to each other. Ideally, a fixing method wants to fine-tune the model such that the decision boundary becomes that in \Cref{fig:demonstration}(d). w-os tends to solve the confusion issue by contracting the decision boundary of the target classes as illustrated in \Cref{fig:demonstration}(c). w-aug, w-loss, and w-dbr try to reduce confusion by shifting the decision boundary. They may be able to achieve \Cref{fig:demonstration}(d) but may also sacrifice the decision boundary for other classes sometimes and get the decision boundary in \Cref{fig:demonstration}(b) instead. w-bn comes in between: on the one hand, it tends to contract the decision boundary as w-os; on the other hand, it tends to shift the decision boundary through fine-tuning.

To check what the effectiveness of the proposed methods and what are their fixing performance in practice, we evaluate them on fixing confusion error and bias error for both single-label and multi-label image classification in \new{six}\sout{five} different settings involving four datasets and \new{five} DNN architectures. Our experiments show that in every setting, a subset of our proposed methods can \new{significantly} reduce the error significantly and most of time at least one method can \new{have similar}\sout{achieve better} accuracy (or mean average precision for multi-label image classification) and lower confusion/bias error than the original model at the same time. We also provide some analysis of the proposed methods' performance and applicability. In summary, we make the following contributions:
\begin{itemize}
\item We propose a generic method for targeted group-wise error fixing of DNN models, \tool, which consists of five specific methods.
\item We compare the five proposed methods performance and show their effectiveness on fixing two types of group-wise errors. 
\end{itemize}
Our code is available at \new{\url{https://github.com/yuchi1989/deeprepair}} \sout{\url{https://github.com/deepfixdeepfix/dnnfix}}.

%
%

\section{Background}
\subsection{Image Classification}
This work focus on two types of image classification problems. In a \textbf{single-label classification} problem, each datum is associated with a single label \textit{l} from a set of disjoint labels $L$ where $|L| > 1$. Typical datasets include CIFAR-10/CIFAR-100~\cite{cifar} where an image can be categorized into one of 10/100 classes. In a \textbf{multi-label classification} problem, each datum is associated with a set of labels \textit{Y} where $Y \subseteq L$. MS-COCO\cite{lin2014microsoft} is a commonly used dataset where an image can be labeled as \textit{car, person, traffic light and bicycle}.

Given any single- or multi-label classification task, a DNN classifier software tries to learn decision boundaries that can separate the classes. In particular, all members of one class, say $C_i$, should be categorized identically irrespective of their individual features, while all members of any other class, say $C_j$, should not be categorized to $C_i$~\cite{bengio2013representation}. A DNN represents a given input image in an embedded space with the feature vector at a certain intermediate layer and uses the layers after as a classifier to classify these representations. The {\em class separation} between two classes estimates how well the DNN has learned to separate each class from the other.
If the embedded distance between two classes is too small compared to other classes, or lower than some pre-defined threshold, the DNN is likely to not be able to separate them from each other.

\subsection{Group-wise Error}
\label{sec:group_wise_error}
The proposed method aims to address the two types of group-wise errors proposed in \cite{tian2019deepinspect}. We provide the definitions of them in the following.
\subsubsection{Confusion Error}
\label{sec:group_wise_error_confusion}
A confusion error occurs when a DNN frequently makes mistakes in disambiguating members of two different classes.

\textit{\ta}: In single-label classification, Type1 confusion occurs when an object of class $A$ (\eg {\tt dog}) is misclassified to another class $B$ (\eg {\tt cat}). We call class $A$ and class $B$ as target classes since they are the classes inducing the error.
For all the objects of class $A$ and $B$, it can be quantified as: 
\[
\typeaconf(A, B) = \mean(\prob(A| B), \prob(B| A))
\]
In other words, it is the DNN's probability to misclassify class $B$ as $A$ and vice-versa, and takes the average value between the two. For example, given two classes {\tt cat} and {\tt dog}, $\typeaconf$ estimates the mean probability of {\tt dog} misclassified to {\tt cat} and vice versa. Note that, this is a bi-directional score, \ie misclassification of $B$ as $A$ is the same as misclassification of $A$ as $B$. 

\textit{\tb}: In multi-label classification, Type2 confusion occurs when an input image contains an object of class $A$ (\eg  {\tt person}) and no object of class $B$ (\eg {\tt bus}), but the model predicts both classes. 
 For a pair of classes, this can be quantified as: 
 \[
 \typebconf(A, B) = \mean(\prob((A, B)| A), \prob((B, A)| B))
 \] 
In other words, it is the probability to detect two objects in the presence of only one. For example, given two classes {\tt bus} and {\tt person}, $\typebconf$ estimates the mean probability of {\tt person} being predicted while predicting {\tt bus} and vice versa. 
This is also a bi-directional score.

\subsubsection{Bias Error}
\label{sec:group_wise_error_bias}
A DNN model is {\em biased} if it associates two classes differently with a third class.   
For example, consider three classes: {\tt man}, {\tt woman}, and {\tt skis}. An unbiased model should not have different error rates while classifying {\tt man} or {\tt woman} in the presence of {\tt skis}. 
To measure such bias formally, \textbf{confusion disparity} ($\cd$) is defined to measure differences in error rate between classes $A$ and $C$ and between $B$ and $C$: 
\[
\cd(A, B, C) = |error(A, C) - error(B, C)|,
\]
where the  $error$ measure can be either $\typeaconf$ or $\typebconf$ as defined earlier. $\cd$ essentially estimates the disparity of the model's error between classes $A$, $B$ (\eg~{\tt man}, {\tt woman}) \wrt a third class $C$ (\eg~{\tt skis}).

Without loss of generality, we assume $error(A, C)\geq error(B, C)$. We call class $A$ positive target class, class $B$ negative target class, and class $C$ anchor target class. The three classes $A$, $B$, and $C$ are also collectively called target classes.

\subsubsection{Repairing Group-wise Error} We define if a group-wise confusion/bias error (type1conf, type2conf, or cd) of a given DNN model is repaired if that error is \new{reduced more}\sout{lower} than a desired user-specified threshold $\lambda_1$ after repairing. Note that it is sometimes acceptable to sacrifice the model's overall accuracy to some extent as long as the overall accuracy loss is smaller than a user-specified threshold $\lambda_2$. \new{In this work, if not stated otherwise, we consider the error reduction to be 25\% of the original error and accuracy loss to be 1\% of the original accuracy.} We next provide some usage scenarios. \sout{One example for this trade-off is in fair ML, it is sometimes acceptable to sacrifice the overall accuracy to avoid a fairness property's violation.}

\new{\textbf{Usage scenarios for repairing confusion errors:} one usage scenario is fixing the erroneous model used in google photo which was once reported to tag black people to gorillas by mistakes \cite{google_tag}. This is likely caused by the model’s confusion between black people and gorillas. Given the model was trained using lots of resources, the developer might want to fix this confusion error through fine-tuning or other strategies rather than retrain a new model from scratch. Meanwhile, it might be tolerable for the model to make some less irritating confusions (e.g. confusion between two very similar genres of dogs) as a trade-off. Another usage scenario is when one wants to have the pair-wise confusions among multiple pairs roughly similar. This relates to group-level fairness definitions (e.g., equalized odds, equal opportunity, etc.) which seek for group-level treatment to be roughly similar \cite{fairsurvey21}.}

\new{\textbf{Usage scenarios for repairing bias errors:} one usage scenario is fixing the well-trained but biased multi-label classification models which have been shown to be much more likely to wrongly classify a woman to a man in the context of doing sports like surfing or skiing \cite{tian2019deepinspect}. Such biased behaviors involving gender are undesirable and can also be regarded as a violation of a kind of group-level fairness property. In fair ML, it is sometimes considered acceptable to slightly sacrifice the overall accuracy to avoid certain fairness property's violation. In fact, such trade-offs between group-level fairness and accuracy are unavoidable (except in extreme cases) and the methods trying to achieve the best trade-offs w.r.t. some group-level fairness properties have been studied in previous work \cite{reduction18, fair_postprocess}.}

\subsection{Regularization}

Based on the no free lunch theorem\cite{nofreelunch}, a specific machine learning algorithm needs to be designed in order to perform well on a specific task. One approach is to leverage regularization to give an algorithm a preference for one solution over another solution. As proposed by Goodfellow \etal \cite{goodfellow2016deep}, regularization are all different approaches expressing preference for different solutions. One well-known technique of regularization is by imposing a penalty into an optimization process to prevent models from overfitting. 
For example, the norm penalty of a DNN model's weight can be added into the loss function to encourage smaller weight and to prevent overfitting. In this work, we apply the idea of regularization in the context of fixing group-wise errors. In the current work, the proposed five concrete weighted regularization methods, which are respectively applied in input phase, model layer phase, output phase and loss phase of training or inference, all try to fix a given target confusing pair or bias triplet by forcing the DNN models to take more effort for the target classes.


\section{Methodology}
\label{method}
Our goal is to reduce a DNN model's confusion error on a targeted pair or bias error on a targeted triplet (defined in \Cref{sec:group_wise_error}) while maintaining its overall accuracy. The targeted pair or triplet are user-specified e.g. the laptop,mouse pair and the surfing, man, woman triplet as shown in \Cref{fig:surfing_woman}.
To achieve our goal, an intuitive approach is to let the DNN model focus more on the targeted pair/triplet. All of our proposed methods follow this intuitive approach but differ in at which stage of the DNN retraining/inference they assign more weights to the targeted pair/triplet as shown in \Cref{fig:overview}. There are two reasons of designing five methods. First, there are underlying intuitions for each method and it is not clear which method will perform the best under each setting. Second, since these methods function at different stages, they have different applicable scenarios.

In the following subsections, we will first introduce the loss function of a generic DNN image classifier to provide notations and serve as a baseline. Next, we introduce each method by showing the underlying motivations \new{ and connections with existing literature}, how they are developed to fix the confusion error and the bias error for both single-label classification and multi-label classification, and finally their applicable scenarios. For simplicity, we only explain our methods in fixing confusion error for one pair of classes and bias error for one triplet of classes. However, our method can be easily extended to fix multiple pairs or triplets by treating every \new{target} pair / triplet the same way as the one demonstrated and in \Cref{sec:res}\new{.~To demonstrate the method's generalizability to multiple pairs, we}\sout{, we further} show the effectiveness of applying our methods to fix confusion errors involving multiple pairs.

\subsection{Baseline: Original Model (orig)\new{~and Fine-tuned Original Model (orig-ft)}}
The original DNN model is trained using a standard objective function:
\[
Loss_{orig} = \mathbb{E}_{(x,y)\sim \mathbb{D}} \mathbf{L}(f(x) , y)
\]
where $\mathbb{D}$ is the underlying data distribution of input $x$ and label $y$, and $\mathbf{L}$ is a loss function. Two widely used classification loss functions are cross-entropy loss and L2 loss. \new{We additionally consider a fine-tuned model which applies the same fine-tuning procedures as w-aug, w-bn, w-loss, and w-dbr discussed in the following but using $Loss_{orig}$ as the objective function during the fine-tuning process.}

\subsection{Regularize Input: Weighted Augmentation (w-aug)}
The weighted augmentation method fine-tunes a given DNN model with reweighted sampling probability for images according to their classes. In particular, it samples more from the target classes and less from the non-target classes. With more sampled from the target classes, the DNN model is expected to be able to better identify these target classes. \new{This method is similar to the over-sampling used to addressing class imbalance\cite{oversampling18, oversampling19}. The difference is that in our setting the augmented classes are the target classes while in the class imbalance setting the augmented classes are the minority classes. However, target classes are not necessarily the minority classes. For example, in \cifars, although all the classes have the same number of samples, large confusion error exist for the dog and cat pair.} The loss function \new{for w-aug} is defined as:
$$Loss_{aug} = \mathbb{E}_{(x,y)\sim \mathbb{D}'} \mathbf{L}(f(x), y)$$ where the probability density function for the weighted distribution $\mathbb{D}'$ is 
\[ 
pdf'(X, Y) = 
\begin{cases} 
\mbox{\sout{$\rho \times$}$pdf(X,Y)$,} & \mbox{if } y\in Y_{target} \\ 
\mbox{$\new{\rho \times} pdf(X,Y)$,} & \mbox{otherwise} 
\end{cases} 
\]
where $pdf$ is the probability density function of the original data distribution $D$. In essence, the images that have labels belonging to the target classes are oversampled by \new{scaling the weights of the non-target classes by} a user specified constant \new{$\rho \in [0,1]$} \sout{$\rho \geq 1$} \sout{times} during fine-tuning. The \new{smaller}\sout{larger} $\rho$ is, \new{the less}\sout{more} effort the DNN model will spend on the \new{non-}target classes compared with \new{the }\sout{other non-}target classes during the fine-tuning.

\subsubsection{Fixing confusion error} The target classes are the confused pair of classes $A$ and $B$ so the DNN should be able to better distinguish one from the other.

\subsubsection{Fixing bias error} The target classes are the biased triplet of classes $A$, $B$ and $C$ so the DNN should be able to better differentiate the three.

\subsubsection{Applicable Scenario} Fine-tuning is allowed.

\subsubsection{Extension to multiple pairs/triplets} \new{We include all the involved classes in the pairs/triplets into $Y_{target}$.}

\subsection{Regularize Model: Weighted Batch Normalization (w-bn)}
\begin{figure}[!htpb]
\centering
    \begin{subfigure}[b]{0.32\linewidth}
        \includegraphics[width=\linewidth]{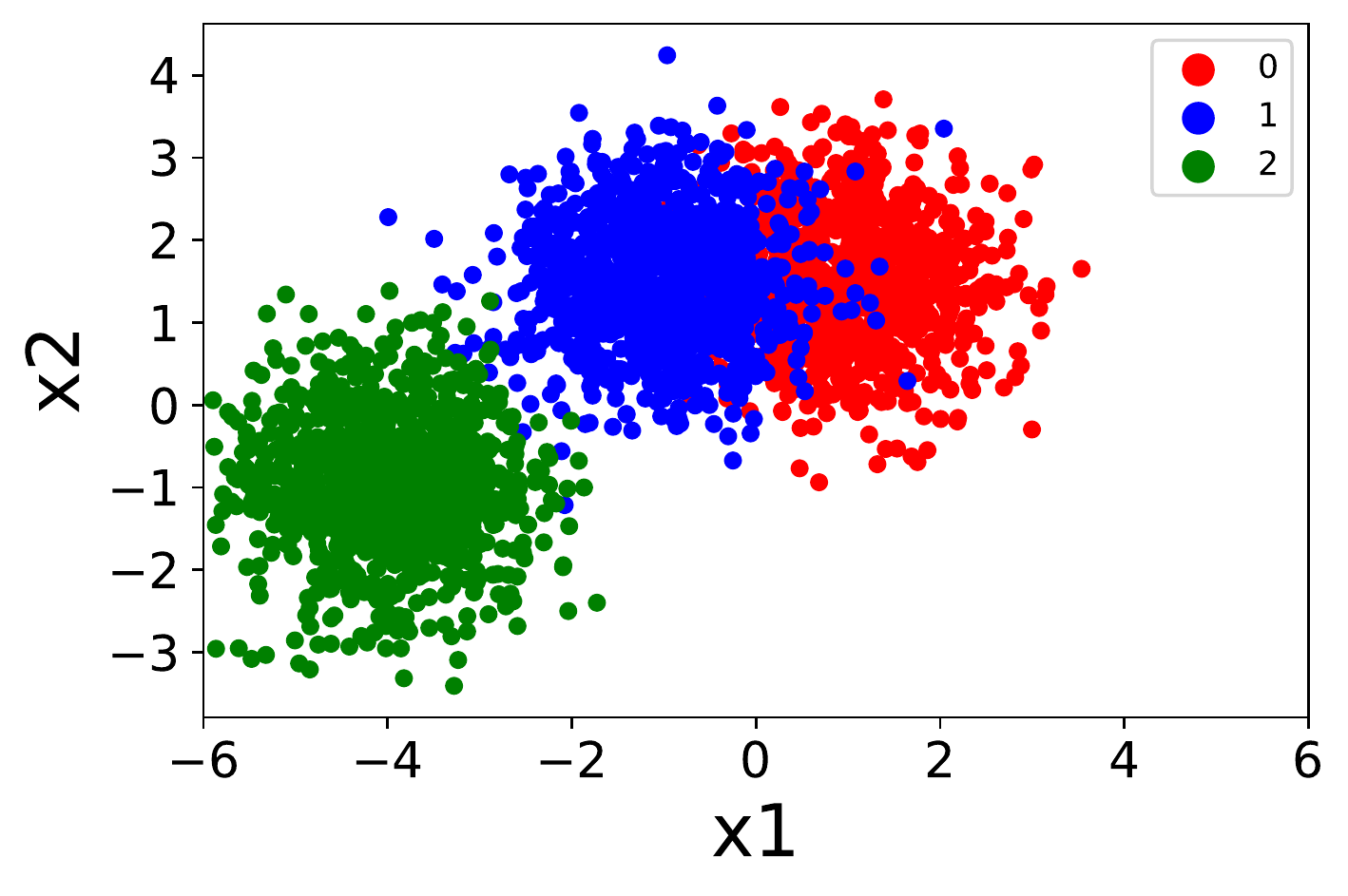}
        \caption{toy 2D dataset}
    \end{subfigure}
    \begin{subfigure}[b]{0.32\linewidth}
        \includegraphics[width=\linewidth]{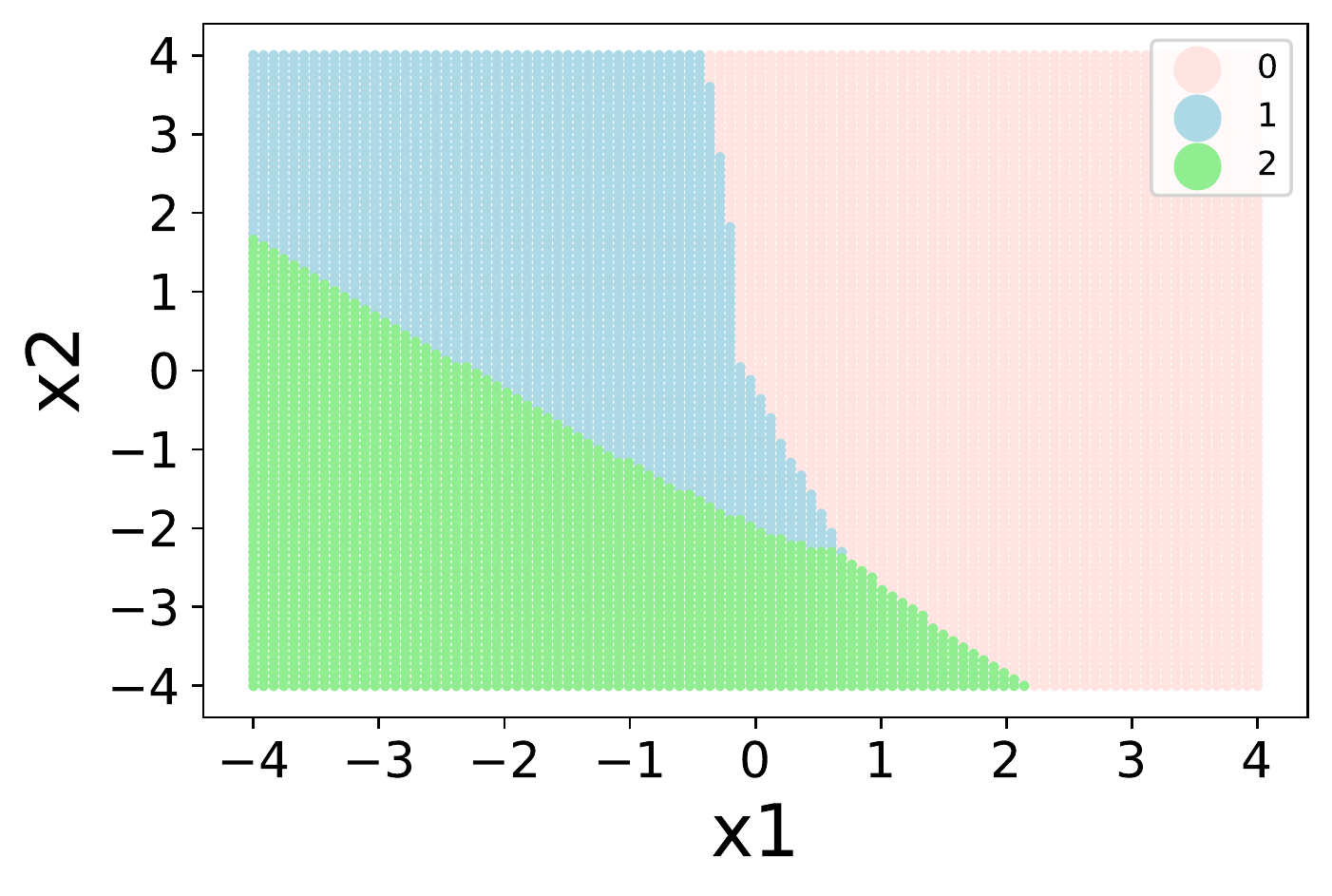}
        \caption{original model}
    \end{subfigure}
    \begin{subfigure}[b]{0.32\linewidth}
       \includegraphics[width=\linewidth]{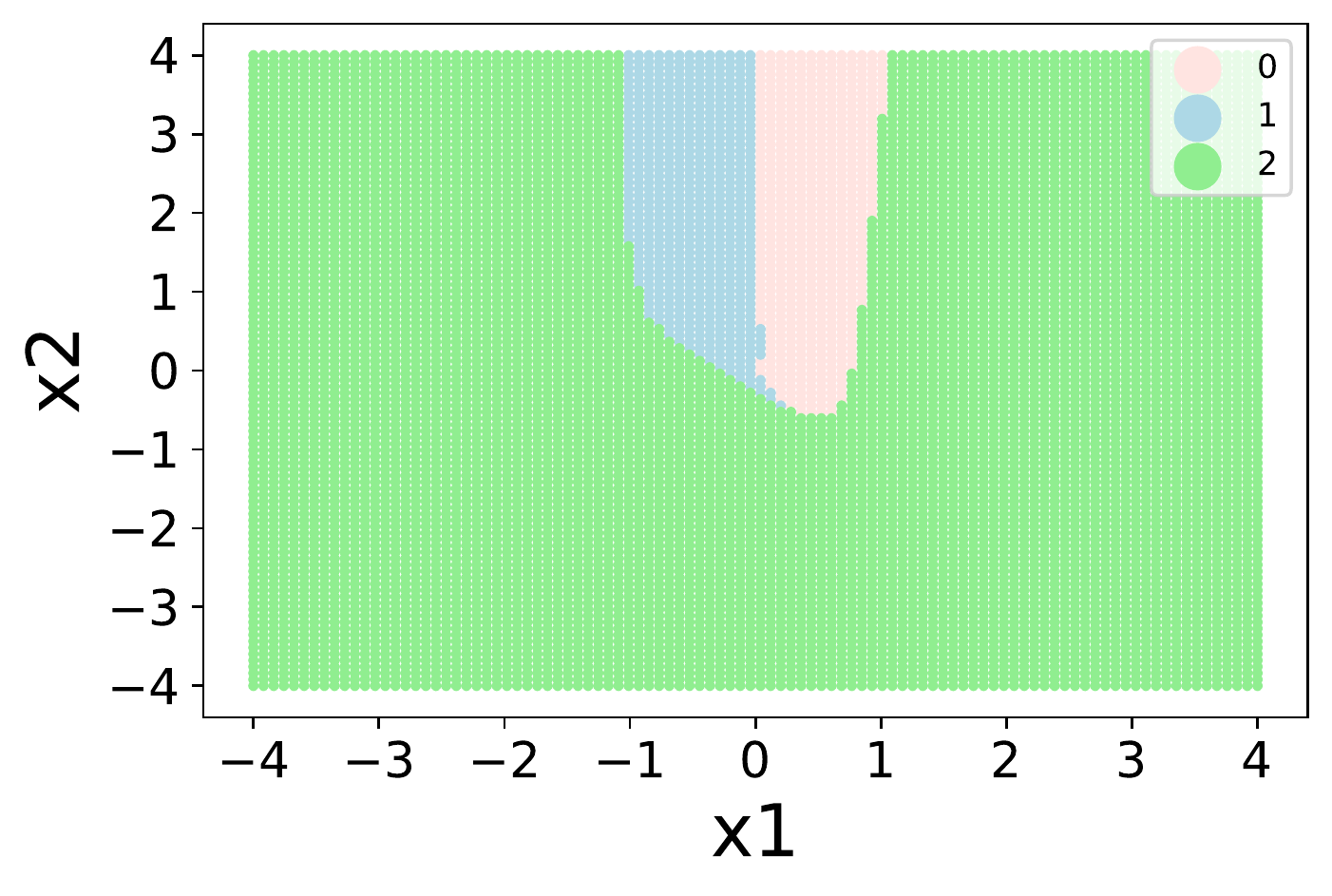}
       \caption{weighted BN}
    \end{subfigure}
\caption{\textbf{\small Shift of decision boundary using weighted BN}}
\label{fig:2d_bn}
\end{figure}

The weighted batch normalization method assumes the existence of batch norm layers of a given DNN model. 
Batch normalization is usually applied after convolutional layers for more stable training and faster convergence\cite{ioffe2015batch} by making the loss landscape smoother\cite{batchnorm18}. It re-centers and re-scales the input data using the estimated mean and variance during training into Gaussian distribution with mean $\beta$ and variance $\gamma$\cite{batchnorm}, where $\beta$ and $\gamma$ are learned during back propagation. 
w-bn redistributes each batch normalization layer by increasing the weights of the targeted classes when estimating the mean and variance of the input data for each batch norm layer. This method comes from a finding that when doing so, the decision boundaries between the target classes and non-target classes can be shifted towards the target classes. To demonstrate this phenomenon, \Cref{fig:2d_bn}(a) shows a toy 2D dataset composed of three classes. \Cref{fig:2d_bn}(b) shows the decision boundary of a well-trained simple ResNet model and \Cref{fig:2d_bn}(c) shows the decision boundary of the model retrained via w-bn. It is noticeable that the decision boundary of class 2 expand over class 0 and class 1. 

With the decision boundaries shifting towards the target classes, there are four consequences. First, the DNN's correct prediction on non-target classes will not be mispredicted to the non-target classes. Second, the DNN's wrong prediction on non-target classes to target classes can potentially be assigned correctly. Third, the DNN's correct prediction on the target classes might be mispredicted to non-target classes. Fourth, the DNN's wrong prediction on the target classes to other target classes might be mispredicted to non-target classes. 

Consequence four will reduce the confusion/bias error. Consequence two benefit while consequence three and four will hurt the overall accuracy. The overall consequence will reduce confusion/bias error and have an uncertain influence on the overall accuracy. Empirically, we find the later depends on specific problem and hyper-parameter tuning.

\begin{figure}[ht]
\centering
    {\includegraphics[width=0.8\textwidth]{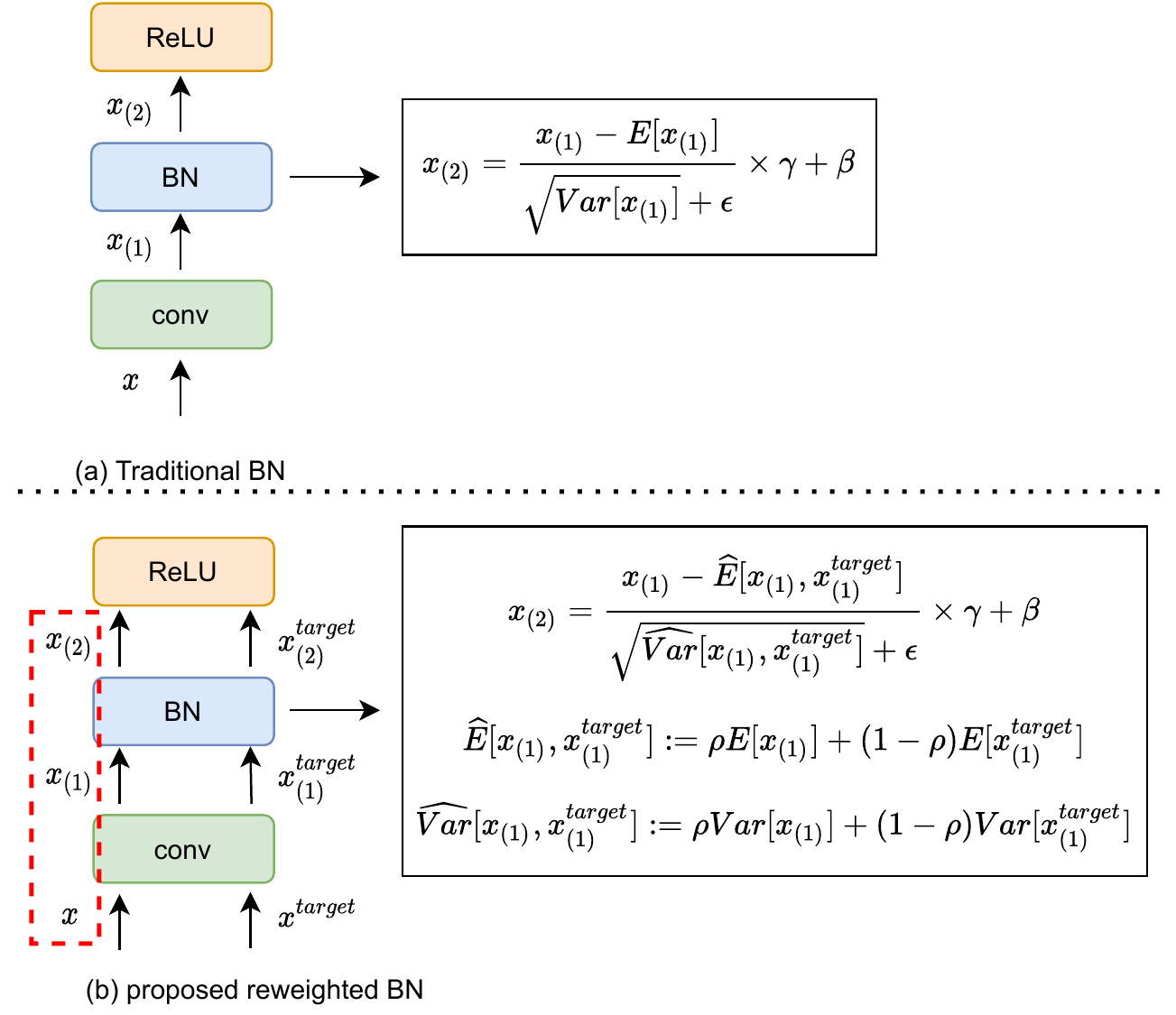}}
\caption{\small{\textbf{Illustration of the traditional BN layer and the proposed weighted BN layer.
}}}
\label{fig:batchnorm}
\vspace{-2mm}
\end{figure}

Next, we formally introduce w-bn in \Cref{fig:batchnorm}. We first denote $x$ to be a regular batch of images and $x^{target}$ to be a batch of images sampled only from the target classes. (a) shows a traditional BN layer and (b) shows a reweighted BN layer. The main differences are that the weighted BN layer passes an extra batch of the target classes and assigns more weights to those data (controlled using a hyper-parameter $\rho \new{~\in [0,1]}$) when estimating the BN statistics (i.e. batch mean $E$ and batch variance $Var$) in the DNN's forward pass. It should be noted that during the back-propagation, only the loss coming from the regular batch (highlighted in dashed red box) is considered. This is because if the extra sampled input data from the target classes are considered in the optimization, this part will be similar to w-aug but we want to evaluate the influence of w-aug and w-bn separately.

\subsubsection{Fixing confusion error} the pair of confused classes $A$ and $B$ are the target classes. As we discussed earlier, many instances confused between $A$ and $B$ will be predicted to non-target classes. Consequently, the confusion between class $A$ and $B$ drops.

\subsubsection{Fixing bias error} the biased triplet of classes $A$, $B$, and $C$ will be included in the target class. The decision boundaries of all the three classes will contract and a subset of images mispredicted from one target class to another is likely to be predicted to other non-target classes. As a result, both confusions between A and C, and B and C tend to drop. It follows that the bias will be reduced as long as the two pairs of confusion drop at relatively similar rate.

\subsubsection{Applicable scenario} Fine-tuning is allowed and the DNN model has batch norm layers.

\subsubsection{Extension to multiple pairs/triplets} \new{We consider all the involved classes in the pairs/triplets as the target classes.}

\subsection{Regularize Output: Weighted Output Smoothing (w-os)}
The weighted output smoothing method tries to reduce a DNN's confusion/bias error when fine-tuning the model is not possible. Since the model itself cannot be updated, only its output can be changed. The key observation is that misclassified images are more likely to have lower confidence in the last layer. Thus, to reduce the confusion/bias error of the target classes, we can make those low confident DNN's prediction on the target classes to be predicted to the most confident non-target classes instead. The consequences are essentially similar to w-bn which contracts the decision boundaries of the target classes. Based on similar reasoning, the confusion/bias errors are likely to be reduced while the impact on the overall accuracy depends on the specific case. 

\new{Note that similar prediction scaling methods have been used for other purposes like calibrating a DNN's prediction to be consistent with its true correctness likelihood\cite{temperaturescaling17} or post-processing a DNN's prediction to improve a DNN's group-level fairness\cite{fair_postprocess}. The main difference is that the concrete scaling strategies are developed differently for different purposes. For example, the post-processing method proposed in \cite{fair_postprocess} assigns different thresholds to each sensitive group for a binary classifier's to mitigate equalized odds/opportunity.} \sout{w-os is analogous to post-processing methods in fair ML literature\cite{fair_postprocess}, where different thresholds are assigned to each sensitive group for a binary classifier's to mitigate equalized odds/opportunity.} The thresholds are set to make trade-off between prediction accuracy and fairness criteria. In contrast, we apply different levels of smoothing for target classes to make trade-off between accuracy and confusion/bias errors.

We denote the last layer's output of a given input x to be $p(x)$, which is a $m$ (i.e. the number of classes) dimensional vector. Each field of $p(x)$ is positively correlated with the prediction probability of the class corresponding to that field.

\subsubsection{Fixing confusion error} For single-label classification, w-os multiplies the target class prediction probability $p(x)_{t}$ by a specified parameter $\new{\rho} \msout{\eta}\in [0,1]$ for images classified into any of the target classes $A$ and $B$; for multi-label classification, w-os multiplies the target class prediction probability $p(x)_{t}$ by $\new{\rho} \msout{\eta}$ for images predicted to have both target classes $A$ and $B$. 

\subsubsection{Fixing bias error} For single-label classification, w-os multiplies the target classes prediction probability $p(x)_{t}$ by $\new{\rho} \msout{\eta}$ for images predicted to be any of the target classes $A$, $B$, and $C$; for multi-label classification, w-os multiplies the target classes prediction probability $p(x)_{t}$ by $\new{\rho} \msout{\eta}$ for images predicted to have the classes $A$ and $C$ or the classes $B$ and $C$.

\subsubsection{Applicable Scenario} We only have access to a model's last layer output.

\subsubsection{Extension to multiple pairs/triplets} \new{We consider all the involved classes in the pairs/triplets as the target classes.}

\subsection{Regularize Loss: Weighted Loss (w-loss)}
The weighted loss method fine-tunes the given DNN model with more weights in the loss function assigned to images contributing to the confusion/bias error. Intuitively, since the model takes more loss when making mistakes on those contributing to the confusion/bias error, it is likely to correctly predict those even \sout{with} \new{at the} sacrifice of the overall accuracy to some extent. \new{This method is similar to the proposed class-balanced loss in \cite{classbalancedloss19} which increase the weight of the minority classes to address the class imbalance problem. The key difference is that the class-balanced loss increases the weights of the loss for those minority classes while w-loss increases the weights of confused samples / bias samples since the former aims to maximize the class-balanced accuracy while the latter aims to mitigate confusion/bias errors of the target classes.} We next provide the new loss functions for fixing confusion and bias error respectively.

\subsubsection{Fixing confusion error}
Denote $Y_{target}=\{A,B\}$. The loss function is defined as:
\begin{align*}
    & Loss_{rl} = \new{\rho}\msout{(1-\rho)} \mathbb{E}_{(x,y)\sim \mathbb{D}} \mathbf{L}(f(x), y)\\
    &+\new{(1-\rho)}\msout{\rho} \mathbb{E}_{(x,y)\sim \mathbb{D}(Y_{target})} \mathbf{L}(f(x), y)
\end{align*}
where the probability density function for the distribution $\mathbb{D}(Y_{target})$ is 
\[ 
pdf'(X, Y) = 
\begin{cases} 
\mbox{$pdf(X,Y)$,} & \mbox{if } (x,y)\sim \mathbb{D}~\textrm{s.t.}~y\in Y_{target} \\ 
\mbox{} &  \textrm{and } f(x)\neq y\textrm{ and }f(x)\in Y_{target}.\\
\mbox{$0$,} & \mbox{otherwise} 
\end{cases} 
\]
\new{ and $\rho\in [0, 1]$ is the hyper-parameter balacing the two objectives.}
Intuitively, the DNN model is encouraged to better differentiate between $A$ and $B$ compared with differentiating among other classes in general.

\subsubsection{Fixing bias error}
The loss function is defined as:
\begin{align*}
    Loss_{rl} &= \new{\rho}\msout{(1-\rho)} \mathbb{E}_{(x,y)\sim \mathbb{D}} \mathbf{L}(f(x), y)\\
    &+\new{(1-\rho)}\msout{\rho} \Big(\mathbb{E}_{(x,y)\sim \mathbb{D}'(Y_{target^+})} \mathbf{L}(f(x), y)\\
    &+\mathbb{E}_{(x,y)\sim \mathbb{D}(Y_{target^-})} \mathbf{L}(f(x), y)\Big)
\end{align*}
where $Y_{target^+}=\{A,C\}$ and $Y_{target^-}=\{B,C\}$. This loss function encourages the DNN model to better differentiate between $A$ and $C$ as well as $B$ and $C$ compared with differentiating among other classes in general.

\subsubsection{Applicable Scenario} Fine-tuning is allowed.

\subsubsection{Extension to multiple pairs/triplets} \new{For each pair(triplet), we add one(two) extra term(terms) into the second half of the loss function and divide this part by the number of pairs(triplets).}

\subsection{Regularize Loss: Weighted Distance-Based Regularization (w-dbr)}
The distance-based regularization method leverages the class-level representation and adds an extra regularization term in the loss function to balance the distance among the target classes in the representation space under a defined metric during the fine-tuning. The insight here is that the closer the two classes representations are, the more confused the model is between the two classes\cite{tian2019deepinspect}. 

Given a class A, we define its class-level representation
$$P_{new}(A) = \frac{[S(n_1), S(n_2), ..., S(n_t)]}{N}$$
where $S(n_i)$ is the sum of each output of neuron $n_i$, given $N$ input images. 
Then, we define the distance metric between two classes A and B as:
$$D_{new}(A, B) = \vert\vert(P_{new}(A),P_{new}(B))\vert\vert_2$$

\subsubsection{Fixing confusion error}
We define a new loss:
$$Loss_{dbr-conf} = \new{\rho}\msout{(1-\rho)} Loss_{orig} - \new{(1-\rho)}\msout{\rho} D_{new}(A, B)$$
where $\rho\new{\in [0,1]}$ trades off the original loss and the new distance-based regularization. In essence, the regularization term encourages a larger separation of the centroids of the two classes $A$ and $B$ in the representation space.

\subsubsection{Fixing bias error}
Similarly, we define a new loss for reducing bias:
$$Loss_{dbr-bias} = \new{\rho}\msout{(1-\rho)} Loss_{orig} + \new{(1-\rho)}\msout{\rho} ~abs(D_{new}(A, C) - D_{new}(B, C)).$$ The regularization term balances the difference between the centroid distance between class $A$ and $C$, and centroid distance between class $B$ and $C$ in the representation space such that the relative distances from $A$ to $B$ and $C$ are similar.

\subsubsection{Applicable Scenario} Fine-tuning is allowed.

\subsubsection{Extension to multiple pairs/triplets} \new{For each pair(triplet), we add one(two) extra term(terms) into the second half of the loss function and divide this part by the number of pairs(triplets).}

\section{Experimental Design}
\subsection{Study Subjects}
\label{sec:study_subject}
We evaluate the proposed method for single-label and multi-label DNN-based classifications including \new{six}\sout{five} combinations of \new{five}\sout{four} DNN architectures and four datasets. For each combination, we choose \new{highly}\sout{the most} confused pair/biased triplet \new{, those potentially have greater ethical implications, or randomly chosen pair/triplet having non-zero confusion/bias} as the target using DeepInspect\cite{tian2019deepinspect}. \sout{We observe the results also hold when selecting other pair/triplet as the target.}

\smallskip
\noindent
\textbf{Datasets:} We conduct our experiments on two single-label image classification datasets, CIFAR-10,  CIFAR-100 and two multi-label image classification datasets, MS-COCO\cite{lin2014microsoft} and MS-COCO gender\cite{zhao2017men}.

\begin{itemize}
    \item \textbf{CIFAR-10}: consists of 50,000 training and 10,000 testing 32x32 color images. It has 10 classes and 6,000 images per class.
    \item \textbf{CIFAR-100}: consists of 50,000 training and 10,000 testing 32x32 color images. It has 100 classes and 600 images per class.
    \item \textbf{MS-COCO}: consists of 80783 training images and 40504 validation images. Each image labeled a subset of 80 objects. 
    \item \textbf{MS-COCO gender}: same as MS-COCO but with the person class split into man and woman classes\cite{zhao2017men}. 
\end{itemize}

\smallskip
\noindent
\textbf{Architectures:} We evaluate our repairing performance on \new{five}\sout{four} different convolutional neural networks\cite{he2016deep, simonyan2014very}.

\begin{itemize}
    \item 
    \textbf{ResNet-18}: ResNet-18 is trained on CIFAR-10 dataset. The model is trained using the training scripts from CutMix\cite{yun2019cutmix}. The training takes 300 epochs and the repairing takes 60 epochs for methods requireing fine-tuning. The initial learning rate is 0.1 and is multiplied by 0.1 after 50\% and 75\% training epochs respectively.

    \item
    \textbf{VGG11\_BN}: VGG11\_BN is a variant of VGG11 model with batch normalization layers  \cite{simonyan2014very}. We train a VGG11\_BN model on CIFAR-10 dataset in \sout{a}\new{the} same way as above.
    
    \item
    \new{\textbf{MobileNetv2}: MobileNetv2 is a popular model tailored for mobile and resource constrained environments \cite{mobilenetv2}. We train it on CIFAR-10 dataset in the same way as above.}
    
    \item 
    \textbf{ResNet-34}: ResNet-34 is trained on CIFAR-100 dataset. The model is trained in the same way as above.
    
    \item 
    \textbf{ResNet-50}: Following Zhao et al\cite{zhao2017men}, we train ResNet-50 models for both MS-COCO and MS-COCO gender datasets. Both models are trained for 12 epochs and are repaired by retraining of another 6 epochs for methods requiring fine-tuning.
    
\end{itemize}
 
\Cref{tab:subj} summarizes our study subjects including the details of all the datasets and models used.

\begin{table}[h]
  \setlength\tabcolsep{1pt} 
  \centering
  \scriptsize
  \caption{\textbf{\small Study Subjects}
  \vspace{-2mm}}
   \label{tab:subj}
  \begin{tabular}{l|l|r||r|r|r|r}
    \toprule
    \multicolumn{3}{c||}{\textbf{Dataset}} & \multicolumn{4}{c}{\textbf{Model}}\\ 
    \midrule
    \textbf{Classification} &  &  & \textbf{Models} & &  & \textbf{Reported} \\
    \textbf{Task} & \textbf{Name} & \textbf{\#classes} &  & \textbf{\#Params} & \textbf{\#Layers} & \textbf{Accuracy} \\
    \midrule
    Multi-label        & \coco\cite{lin2014microsoft} & 80 & ResNet-50\cite{he2016deep} & 23,671,952 & 174 &  0.6603* \\
classification & \coco gender\cite{zhao2017men} & 81 & ResNet-50\cite{he2016deep} & 23,674,001 & 174 & 0.6691* \\
  \midrule
     Single-label  & CIFAR-100\cite{cifar} & 100  & ResNet-34 \cite{yun2019cutmix} & 336,244  & 101 & 0.6961\dag \\

 \cmidrule{2-7}

 classification  & CIFAR-10\cite{cifar}               &  10  & ResNet-18\cite{yun2019cutmix} & 127,642  & 41 & 0.8747\dag  \\
 & & & VGG11-BN\cite{simonyan2014very} &9,756,426   &36  & 0.9175\dag \\
 & & & \new{MobileNetv2\cite{mobilenetv2}} &\new{2,296,922}   &\new{115}  & \new{0.9420}\dag \\
    \bottomrule
    \end{tabular}%

{\small * reported in mean average precision, \dag reported in mean accuracy}

\end{table}

\smallskip
\noindent
\new{\textbf{Pairs/Triplets Selection:} In this work, we mainly focus on mitigating the user selected confusion pairs / biased triplets. We next discuss how we choose the pairs/triplets used for the experiments.}

\new{First, for the selection of confused pairs:}
\begin{itemize}
    \item \new{COCO, CIFAR-100, CIFAR-10: we choose the most confused pairs in terms of type1conf/type2conf (defined in \Cref{sec:group_wise_error}).}
    
    \item \new{COCO gender: since the major difference between COCO gender and COCO is that the person class is categorized into man and woman and the confusion related to gender are of high interest in the study of fairness, we choose pairs with respect to gender. In particular, we choose the “woman-handbag” pair which is a relatively (but not the most) confused pair.}
    
    \item \new{COCO two pairs and CIFAR-10 two pairs: we keep the original pair and choose another one randomly among the pairs with non-zero confusions with each other.}
\end{itemize}

\new{Second, for the selection of biased triplets:}
\begin{itemize}
    \item \new{COCO, CIFAR-100, CIFAR-10: we keep the original two classes in the confused pair and randomly choose a third class such that the bias in terms of cd (defined in \Cref{sec:group_wise_error_confusion}) is larger than 0.}
    
    \item \new{COCO gender: with the same motivation as in choosing the confused pairs, we choose the “man-woman-skis” triplet which involves both man and woman as well as being a highly biased triplet.}
\end{itemize}

\subsection{Baseline}
The baseline\new{s} we use \new{are}\sout{is} the original model\new{s} \new{and the fine-tuned original model} which \new{have}\sout{has} been properly trained as discussed in \Cref{sec:study_subject}. We did not compare with methods for instance-wise fixing or dataset-wise fixing because they target different problems. \new{We also did not compare with methods developed for class imbalance directly since our methods w-aug, w-os, w-loss can already be regarded as the adaptations of those methods for the current problem setting.}

\subsection{Evaluations Metrics}
For either fixing the confusion error or bias error, the goal is to reduce error while maintaining the model's overall accuracy. 
\subsubsection{Rank Sum} Since there are two goals i.e. high accuracy and low confusion/bias a model tries to achieve, for comparison purpose, we rank each fixed model (including the original model) by accuracy and confusion respectively. Next, we sum up the two ranks for each model and compare the rank sums. The model with the smallest rank sum is considered the one that achieves the best trade-off between accuracy and confusion/bias. Note that we choose this simple metric to give the most intuitive performance comparisons. A more complicated method like weighted combination of accuracy and confusion/bias can also be used.

\subsubsection{Statistical Tests} \new{To validate the statistical significance of the results, we additionally conduct Wilcoxon rank-sum test \cite{Wilcoxon} and Vargha-Delaney effect size test \cite{VDtest, guidetostatstest} between the proposed methods and the baseline orig-ft in terms of both accuracy difference and confusion difference. To highlight the results, we only apply the tests for confusion difference on those already having smaller average confusion than orig-ft. Next, we only further apply the tests for accuracy difference for those methods that have statistically significantly (at least at the 0.10 level) lower confusion than orig-ft. Ideally, when compared with orig-ft, a fixed model should have statistically significantly smaller confusion and no significant lower accuracy at the same time. If not stated otherwise, we use 0.05 as the default significance level.}

\subsection{Hyper-parameter Selection}
\label{sec:hyperparam_selection}
\new{To show the influence of hyper-parameter choice on each method's performance as well as selecting the best one, under each setting, we apply a grid search on $\{0.1, 0.3, 0.5, 0.7, 0.9\}$ by running each method with each hyper-parameter one times and compare the results. Since w-os is a very fast post-processing method but tends to have very different scales across datasets, for datasets (CIFAR-10, CIFAR-100) where w-os(0.1) does not reduce confusion/bias much, we further try 0.01, 0.001, and 0.0001 in order until we have found one that can reduce confusion/bias by at least 25\% when compared with that of orig-ft. We present the detailed results of hyper-parameter search in \Cref{sec:hyperparam_tuning}.
}
\new{Since there are two objectives to optimize: mitigating confusion for the target classes and keeping the overall accuracy of all classes, for each method, we apply the following procedures to select the best performing one:}
\begin{itemize}
    \item \new{If there are hyper-parameters that result in the models having confusion at least 25\% (one can change this number according to one’s need) smaller than the original model, among them, select the one with the highest overall accuracy.}
    
    \item \new{Otherwise, if there are hyper-parameters that result in confusion decreases or stays the same, among them, select the one with the highest overall accuracy.}
    
    \item \new{Otherwise, among all the hyper-parameters, select the one with the highest overall accuracy.}
\end{itemize}
\new{If the selected hyper-parameter does not result in higher accuracy and lower confusion when compared with the original model, and such a hyper-parameter exists for the current method, this initially not selected hyper-parameter is additionally selected.}

\subsection{Research Questions.}
We investigate two research questions to evaluate \tool for target bug fixing of DNNs: 

\begin{itemize}
    \item 
    \textbf{RQ1.}~Can \tool fix confusion errors of DNN models for both single-label classification and multi-label classification effectively? 
    \item
    \textbf{RQ2.}~Can \tool fix bias errors of DNN models for both single-label classification and multi-label classification effectively?
\end{itemize}

\section{Results}
\label{sec:res}
\RQ{1}{Fixing Confusion Error}
\label{sec:fix_conf}
We first explore if the proposed methods can reduce confusion errors effectively. In particular, we evaluate them on two settings for the task of multi-label classification and \new{four}\sout{three} settings for the task of single-label classification. 
\new{We run each method with the selected hyper-parameter(s) (based on the procedure in \Cref{sec:hyperparam_selection}) for five times to compare with each other as well as the baselines, and apply statistical tests to check significance.} 
\sout{To show the influence of hyper-parameter choice on each method's performance, we compare the results of two hyper-parameter choices for each method.In practice, because every method uses a hyper-parameter to balance the weights assigned to the target and non-target classes, one can start with a hyper-parameter that assigns equal weights, and then adjust the hyper-parameter based on results on a validation set. In total, we show the results of eleven methods (including the original model) for each setting.}

\Cref{tab:conf} shows the \new{main} results \new{for reducing confusion following the previously mentioned procedures}, where we highlight the \new{top2 (or top3 if tied)}\sout{top3 (or top4 if tied)} methods having the smallest rank sums. In summary, under every setting, \new{at least two}\sout{many} methods can achieve lower confusion while preserving decent overall accuracy (or mean average precision). For example, w-os can almost always decrease the confusion \new{more than 25\%}\sout{to close to 0} under every setting while maintaining accuracy at a reasonable level \new{(no more than 1\%)}.

\begin{table}[h]
  \setlength\tabcolsep{1pt} 
  \centering
  \scriptsize
  \caption{\textbf{\small Results on Confusion}
  \vspace{-2mm}}
   \label{tab:conf}
  \begin{tabular}{l|l|l|l|l|l|l|l|l|l|l|l|l}
    \toprule
    \textbf{Dataset} & \textbf{Model} & \textbf{Target} & \textbf{Method} & \textbf{Accuracy} & \textbf{Confusion} & \textbf{Acc} & \textbf{Conf} & \textbf{Rank} & \textbf{Acc} & \textbf{Acc} & \textbf{Conf} & \textbf{Conf} \\ 
     
     &  & \textbf{Classes} & & & & \textbf{Rank} & \textbf{Rank} & \textbf{Sum} & \textbf{W} & \textbf{VD} & \textbf{W} & \textbf{VD}\\ 
    \midrule

   \coco* &  ResNet-50 & person,   & orig & $0.6604$ & $0.2381$ & 5 & 7 & 12 & & & &\\
          &            & bus & orig-ft   & $0.6614\pm 0.0003$ & $0.2329\pm 0.0084$ & 2 & 6 & 8 & & & & \\
          &            & & w-aug(0.9)   & $0.6611\pm 0.0004$ & $0.2622\pm 0.0097$ & 4 & 8 & 12 & & & &\\
          &            & & \textcolor{red}{w-bn(0.9)}   & $0.6617\pm 0.0005$ & $0.1875\pm 0.0121$ & 1 & 4 & 5 & 0.347 & 0.680(m) & 0.009 & 0.0(l)\\
          &            & & \textcolor{red}{w-loss(0.9)}   & $0.6613\pm 0.0002$ & $0.1244\pm 0.0094$ & 3 & 3 & 6 & 0.676	& 0.420(s)	& 0.009	& 0.0(l)\\
          &            & & \textcolor{red}{w-dbr(0.9)}   & $0.6604\pm 0.0004$ & $0.0077\pm 0.0012$ & 5 & 1 & 6 & 0.009	& 0.0(l) & 0.009	& 0.0(l)\\
          &            & & w-os(0.7)   & $0.6602$ & $0.1159$ & 8 & 2 & 10 & 0.009 & 0.0(l) & 0.009 & 0.0(l)\\
          &            & & w-os(0.9)   & $0.6604$ & $0.203$ & 5 & 5 & 10 & & & &\\
    \midrule
    \coco gender*  &  ResNet-50 & handbag,  & orig & $0.6701$ & $0.0402$ & 5 & 5 & 10 & & & &\\
          &            & woman & \textcolor{red}{orig-ft}   & $0.6710\pm 0.0002$ & $0.0394\pm 0.0029$ & 1 & 4 & 5 & & & & \\
          &            &  & w-aug(0.9)   & $0.6707\pm 0.0004$ & $0.0442\pm 0.0069$ & 2 & 6 & 8 & & & &\\
          &            & & \textcolor{red}{w-bn(0.9)}   & $0.6707\pm 0.0003$ & $0.0225\pm 0.0059$ & 2 & 2 & 4 & 0.251	& 0.280(m)	& 0.009	& 0.0(l)\\
          &            & & \textcolor{red}{w-loss(0.7)}   & $0.6707\pm 0.0002$ & $0.0225\pm 0.0059$ & 2 & 3 & 5 & 0.095	& 0.180(l)	& 0.009	& 0.0(l)\\
          &            & & w-dbr(0.9)   & $0.6700\pm 0.0004$ & $0.0725\pm 0.0079$ & 6 & 7 & 13 & & & & \\
          &            & & w-os(0.7)   & $0.6698$ & $0.0044$ & 7 & 1 & 8 & 0.009	& 0.0(l)	& 0.009	& 0.0(l)\\
    \midrule
\cifar    &  ResNet-34 & girl, & orig & $0.6998$ &$0.155$ & 4 & 7 & 11 & & & &\\
          &            & woman & \textcolor{red}{orig-ft}   & $0.7051\pm 0.0025$ &$0.1430\pm 0.0179$ & 1 & 4 & 5 & & & &\\
          &            &  & w-aug(0.3)   &$0.7051\pm 0.0011$ &$0.1520\pm 0.0311$ & 1 & 6 & 7 & & & &\\
          &            & & \textcolor{red}{w-bn(0.9)}   &$0.7034\pm 0.0013$ & $0.1240\pm 0.0192$ & 3 & 3 & 6 & & & 0.175 & 0.240(l)\\
          &            & & w-loss(0.9)   &$0.6956\pm 0.0108$ & $0.1440\pm 0.0074$ & 6 & 5 & 11 & & & &\\
          &            & & w-dbr(0.9)   &$0.6818\pm 0.0046$ &$0.1210\pm 0.0082$ & 7 & 2 & 9 & 0.009 & 0.0(l) & 0.022 & 0.060(l)\\
          &            & & \textcolor{red}{w-os(0.1)}   &$0.6976$ &$0.105$ & 5 & 1 & 6 & 0.009 & 0.0(l) & 0.009 & 0.0(l)\\
    \midrule
    \cifars   &  ResNet-18 & cat,  & orig & $0.8747$  & $0.0960$ &5  &5 &10 & & & &\\
          &            & dog & \textcolor{red}{orig-ft}   & $0.8779\pm 0.0009$   & $0.0993\pm 0.0041$  &1  &7 &8 & & & &\\
          &            & & w-aug(0.7)       & $0.8778\pm 0.0016$ & $0.0966\pm 0.0017$ &3  &6 &9 & & & &\\
          &            & & w-bn(0.7)      & $0.8489\pm 0.0007$  & $0.0671\pm 0.0007$  &8  &1 &9 & 0.009 & 0.0(l) & 0.009 & 0.0(l)\\
          &            & & w-bn(0.9)      & $0.8746\pm 0.0010$   & $0.866\pm 0.0024$  &6  &3 &9 & 0.009 & 0.0(l) & 0.009 & 0.0(l)\\
          &            & & w-loss(0.9)      & $0.8761\pm 0.0005$  & $0.1003\pm 0.0013$ &4  &8 &12 & & & &\\
          &            & & \textcolor{red}{w-dbr(0.9)}    & $0.8779\pm0.0007$  & $0.0891\pm 0.0024$ &1   &4 &5 & 0.917 & 0.48(n) & 0.009 & 0.0(l)\\
          &            & & w-os(0.1)    & $0.8654$  & $0.071$ &7  &2 &9 & 0.009 & 0.0(l) & 0.009 & 0.0(l)\\
          
          \cmidrule{2-13}

              &  VGG-11  & cat,  & orig  & $0.9197$ & $0.081$ & 1 & 6 & 7& & & &\\
          &            with BN & dog & orig-ft   & $0.9180\pm 0.0020$ & $0.0757\pm 0.0070$ & 4 & 4 & 8& & & &\\
          &            & & \textcolor{red}{w-aug(0.9)}   & $0.9187\pm 0.0011$ & $0.0724\pm 0.0040$ & 3 & 3 & 6& 0.347 & 0.68(m) & 0.296 & 0.3(m)\\
          &            & & w-bn(0.7)   & $0.9070\pm 0.0011$ & $0.0539\pm 0.0032$ & 6 & 2 & 8& 0.009 & 0.0(l) & 0.009 & 0.0(l)\\
          &            & & w-loss(0.9)   & $0.9129\pm 0.0019$ & $0.0866\pm 0.0026$ & 5 & 7 & 12& & & &\\
          &            & & w-dbr(0.1)   & $0.7402\pm 0.0400$ & $0.0908\pm 0.0288$ & 8 & 8 & 16& & & &\\
          &            & & w-os(0.001)   & $0.9059$ & $0.0525$ & 7 & 1 & 8& 0.009 & 0.0(l) & 0.009 & 0.0(l)\\
          &            & & \textcolor{red}{w-os(0.9)}   & $0.9197$ & $0.0805$ & 1 & 5 & 6& & & &\\
          \cmidrule{2-13}

              &  MobileNetv2  & cat,  & \textcolor{red}{orig}  & $0.9420$ & $0.0565$ & 1 & 4 & 5& & & &\\
          &             & dog & orig-ft   & $0.9384\pm 0.0012$ & $0.0632\pm 0.0041$ & 5 & 7 & 12& & & &\\
          &            & & {w-aug(0.5)}   & $0.9388\pm 0.0015$ & $0.0591\pm 0.0033$ & 4 & 6 & 10& & & &\\
          &            & & w-bn(0.7)   & $0.9159\pm 0.0018$ & $0.0357\pm 0.0030$ & 8 & 1 & 9& 0.009 & 0.0(l) & 0.009 & 0.0(l)\\
          &            & & w-loss(0.9)   & $0.9341\pm 0.0012$ & $0.0667\pm 0.0064$ & 6 & 8 & 14& & & &\\
          &            & & w-dbr(0.9)   & $0.9397\pm 0.0014$ & $0.0571\pm 0.0033$ & 3 & 5 & 8& 0.009 & 0.0(l) & 0.009 & 0.0(l)\\
          &            & & w-os(0.001)   & $0.9226$ & $0.0360$ & 7 & 2 & 9& & & &\\
          &            & & \textcolor{red}{w-os(0.3)}   & $0.9420$ & $0.0415$ & 1 & 3 & 4& 0.009 & 1.0(l) & 0.009 & 0.0(l)\\

    \bottomrule
    \end{tabular}%
 
{\small * reported in mean average precision; \new{Acc: Accuracy; Conf: Confusion; VD: Vargha-Delaney effect size test; W: Wilcoxon rank-sum test; n: negligible; s: small; m:medium; l: large}}
\end{table}

\new{For the multi-label classification task, both w-loss and w-bn achieve good trade-off between mean average precision and confusion in terms of the rank sum. On both the \coco and \coco gender datasets, compared with orig-ft, at the 0.05 significance level, both w-loss and w-bn reduce confusion significantly with large effect size and do not significantly decrease the overall mean average precision at the same time. For example, on \coco, compared with orig-ft, w-loss(0.9) has much smaller average confusion between person and bus (0.1244 VS 0.2329) and its mean average precision is only smaller by 0.0001 which is not statistically significant.}

\sout{For the multi-label classification task, w-loss strikes the best trade-off between mean average precision and confusion in terms of the rank sum.  On both the \coco and \coco gender datasets, w-loss improves the mean average precision and reduces confusion compared with the original model. For example, on the \coco dataset, w-loss(0.4) improves the mean average precision from 0.6603 to 0.6611 and reduces confusion between person and bus from 0.2381 to 0.03457; on the \coco gender dataset, w-loss(0.4) improves the mean average precision from 0.6691 to 0.6697 and reduces confusion between handbag and woman from 0.0394 to 0.02063.}

\new{For the single-label classification task, w-os usually achieves good trade-offs between accuracy and confusion. It is among the top2 in terms of the rank sum in three out of the four settings. On \cifar, orig-ft is the best in terms of rank sum. In order to mitigate confusion, accuracy has to be sacrificed. This can be seen from w-os(0.1) where the average confusion decreases from 0.1430 to 0.105 but the average accuracy drops from 0.7051 to 0.6976. Both differences are statistically significant at the 0.01 level. In the setting of \cifars and ResNet-18, and \cifars and VGG-11 with BN, w-dbr(0.9) and w-aug(0.9) are ranked the top respectively. They both achieve significantly smaller confusion while maintaining the accuracy with no significant drop. On \cifars with MobileNetV2, w-os(0.3) is ranked the top and has both significant smaller confusion and higher accuracy when compared with orig-ft.}

\sout{For the single-label classification task, on \cifars, w-dbr(0.1) is ranked the top while on \cifar, w-aug(5) is ranked the top. They also achieve higher accuracy and lower confusion compared with the original models in the same setting, respectively. In particular, on the \cifars dataset and ResNet-18 model, w-dbr(0.1) improves the overall accuracy from 0.8747 to 0.8764 and reduces confusion between cat and dog from 0.096 to 0.09; on the \cifars dataset and VGG-11 with BN model, w-bn(0.4) improves the overall accuracy from 0.9175 to 0.919 and reduces confusion between cat and dog from 0.083 to 0.076; on the \cifar dataset, w-aug(5) improves the accuracy from 0.6961 to 0.6697 and reduces confusion between girl and woman from 0.15 to 0.12.}

\new{Under all the settings, both w-os and w-bn can always decrease confusion significantly, and are ranked among the top ones in many settings. Although w-loss works very well on \coco and \coco gender, it works poorly on \cifars and \cifar. A deeper exploration of the retraining process reveals that its retraining processes on \cifars and \cifar tend to be very unstable. For example, on \cifar, it tends to misclassify dog to cat much more frequently at one epoch and the reverse at another. w-aug does not perform well in general. This is potentially because the cause of confusion is not due to the lack of instances belonging to the target classes. This is particularly true for \cifar and \cifars where all the classes have the same number of instances for each class. w-dbr can reduce confusion for \coco , \cifar and \cifars but fails to do so on \coco gender. One possibility is that the confusion between handbag and woman is already relatively small and the class centroids between woman and handbag are far away from each other so the extra loss regularization term does not help much to reduce the confusion further.}

\sout{Under all the settings, w-os works very well and achieves decent trade-off between accuracy and confusion. In particular, w-os with different hyper-parameters is ranked at the top3 for every setting. On the flip side, it has to trade accuracy for reducing confusion.  w-bn also works reasonably well across all the settings but slightly worse than w-os in most settings. Although w-loss works very well on \coco and \coco gender, it works poorly on \cifars and \cifar. A deeper exploration of the retraining process reveals that its retraining processes on \cifars and \cifar tend to be very unstable. For example, on \cifar, it tends to misclassify dog to cat much more frequently at one epoch and the reverse at another. w-aug is ranked among the top on \cifar and \cifars but performs quite poorly on \coco and \coco gender. This is because the causes of confusions between single-label classification and multi-label classification are different. For \cifar and \cifars, the high confusions result from similar features between two classes when the target classes are oversampled during training, the difference of features are better learned by the models. However, for \coco and \coco gender the confusions mostly result from two objects appearing together frequently in the same image or having similar background. When the target classes are oversampled, the confusions may be increased. w-dbr can reduce confusion for \coco , \cifar and \cifars but fail to do so on \coco gender. One possibility is that the confusion between handbag and woman is already very low and the class centroids between woman and handbag are far away from each other so the extra loss regularization term does not help  reduce the confusion further.}

\begin{figure}[!htpb]
\centering
    \begin{subfigure}[b]{0.23\linewidth}
        \includegraphics[width=\linewidth]{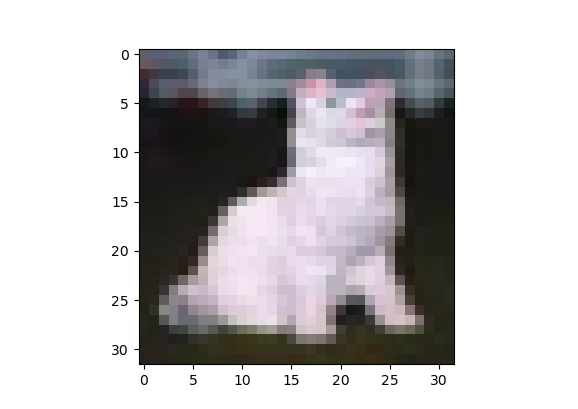}
        \caption{cat}
    \end{subfigure}
    \begin{subfigure}[b]{0.23\linewidth}
        \includegraphics[width=\linewidth]{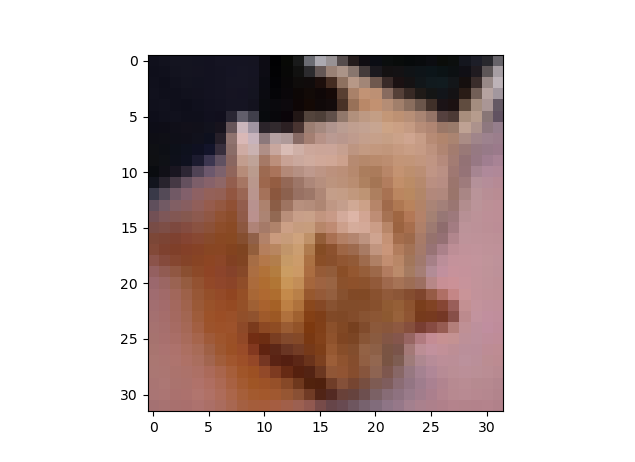}
        \caption{cat}
    \end{subfigure}
    \begin{subfigure}[b]{0.23\linewidth}
       \includegraphics[width=\linewidth]{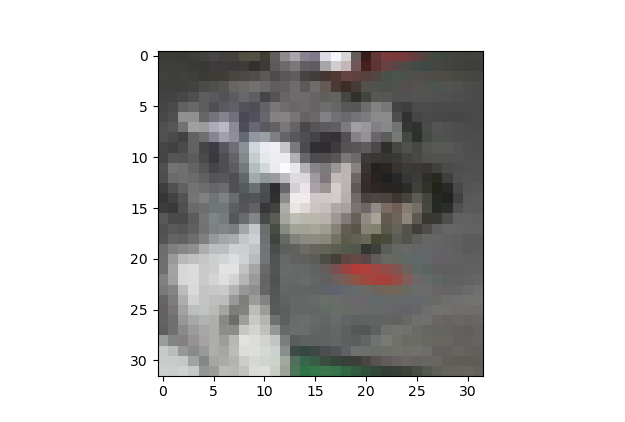}
       \caption{dog}
    \end{subfigure}
    \begin{subfigure}[b]{0.23\linewidth}
       \includegraphics[width=\linewidth]{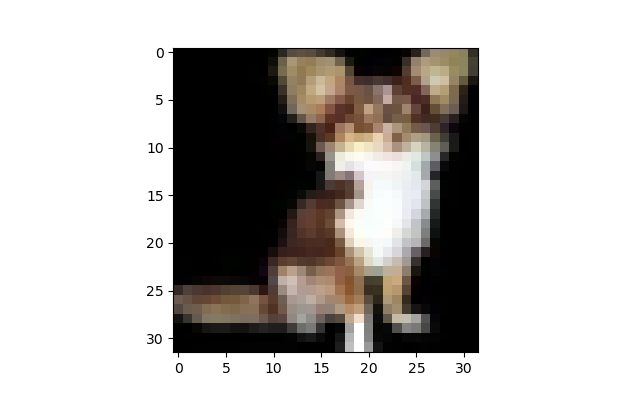}
       \caption{dog}
    \end{subfigure}
    
    \begin{subfigure}[b]{0.23\linewidth}
        \includegraphics[width=\linewidth]{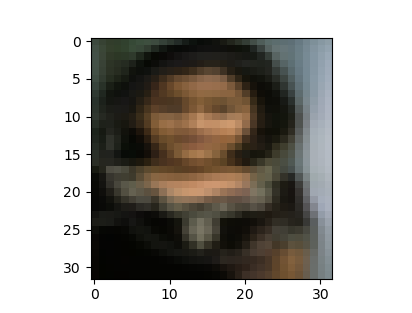}
        \caption{girl}
    \end{subfigure}
    \begin{subfigure}[b]{0.23\linewidth}
        \includegraphics[width=\linewidth]{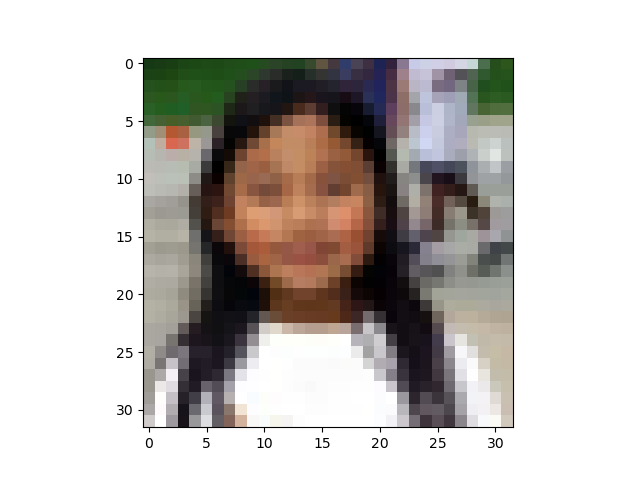}
        \caption{girl}
    \end{subfigure}
    \begin{subfigure}[b]{0.23\linewidth}
       \includegraphics[width=\linewidth]{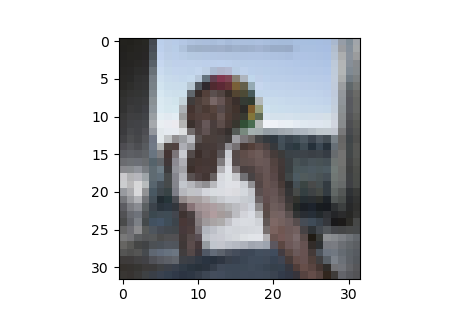}
       \caption{woman}
    \end{subfigure}
    \begin{subfigure}[b]{0.23\linewidth}
       \includegraphics[width=\linewidth]{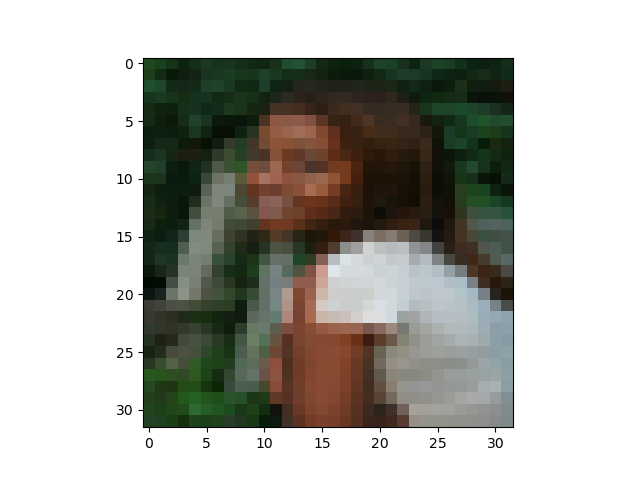}
       \caption{woman}
    \end{subfigure}
    
    \begin{subfigure}[b]{0.23\linewidth}
        \includegraphics[width=\linewidth]{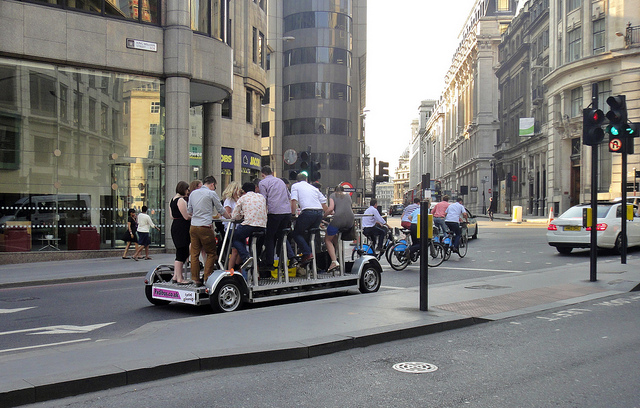}
        \caption{person}
    \end{subfigure}
    \begin{subfigure}[b]{0.23\linewidth}
        \includegraphics[width=\linewidth]{img/confusion_coco/person_to_bus2.jpg}
        \caption{person}
    \end{subfigure}
    \begin{subfigure}[b]{0.23\linewidth}
       \includegraphics[width=\linewidth]{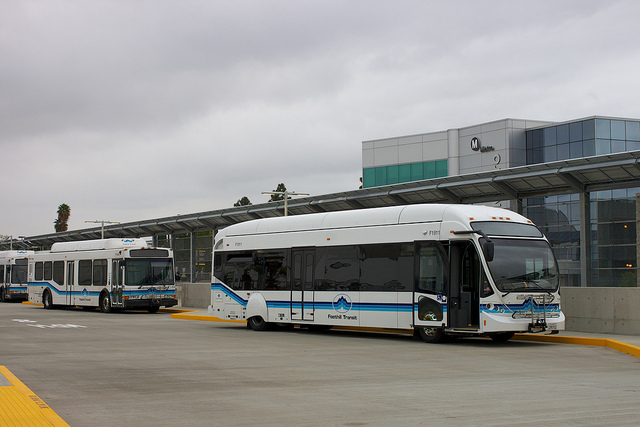}
       \caption{bus}
    \end{subfigure}
    \begin{subfigure}[b]{0.23\linewidth}
       \includegraphics[width=\linewidth]{img/confusion_coco/bus_to_person2.jpg}
       \caption{bus}
    \end{subfigure}
\caption{\textbf{\small Fixed confusion errors on \cifars ((a)-(d)), \cifar ((e)-(h)), and \coco ((i)-(l)) respectively.}}
\label{fig:fixed_examples}
\end{figure}

\Cref{fig:fixed_examples} shows some examples of the fixed confusion instances. On the \cifars dataset \new{and ResNet-18 combination}, \Cref{fig:fixed_examples}(a)-(b) show two cat images that were classified to dog by the original model. After applying w-dbr, they are correctly predicted to cat. \Cref{fig:fixed_examples}(c)-(d) show two dog images that were classified to cat by the original model. After applying w-dbr, they are correctly predicted to dog. Similarly, on the \cifar dataset \new{and ResNet-34 combination}, \Cref{fig:fixed_examples}(e)-(f) show two girl images that were classified to woman by the original model. After applying w-dbr, they are correctly predicted to girl.  \Cref{fig:fixed_examples}(g)-(h) show two woman images that were classified to girl by the original model. After applying w-dbr, they are correctly predicted to woman.  On the \coco dataset \new{and ResNet-50 combination}, \Cref{fig:fixed_examples}(i)-(j) show two images that contain only person but the original model mispredicts the presence of bus. After applying w-loss, the model correctly predicts the presence of person without false positively predicting the presence of bus. Similarly, \Cref{fig:fixed_examples}(k)-(l) show two images with only bus in them but the original model false positively predicts the presence of person as well. After applying w-loss, the model can correctly predict the presence of bus while not falsely predicting the presence of person.

Next, we evaluate if the proposed methods can be applied to fix confusion errors among multiple pairs at the same time by applying the proposed methods to fix confusions of two pairs (one top confused pair and one randomly picked confused pair) on \cifars and \coco. \Cref{tab:conf_multi} shows the results.  
\new{On \coco, w-loss(0.9) achieves the best trade-off between accuracy and confusion in terms of rank sum. Compared with orig-ft, it has smaller confusion at 0.01 significance level and no significant accuracy drop. On \cifars, w-bn(0.9) achieves the best trade-off. However, compared with orig-ft, it does not have significantly smaller confusion. In contrast, w-dbr(0.9) has smaller confusion than orig-ft at the 0.1 significance level and no significant accuracy drop.}
\sout{On \coco, w-loss(0.6) achieves the best trade-off between accuracy and confusion. It reduces the summed confusions of two pairs (person-bus and mouse-keyboard) from 0.4025 to 0.0723 while only sacrificing mean average precision by 0.0001 (from 0.6604 to 0.6603). On \cifars, w-dbr(0.1) achieves the best trade-off by increasing the overall accuracy from 0.8747 to 0.8778 and reducing the confusion from 0.134 to 0.128. We also want to highlight that w-os is ranked in top3 for both settings and significantly reduces the confusion in both settings while only slightly sacrificing the mean average precision and accuracy, respectively.}

\begin{table}[h]
  \setlength\tabcolsep{1pt} 
  \centering
  \scriptsize
  \caption{\textbf{\small Results on Confusion (Two Pairs)}
  \vspace{-2mm}}
   \label{tab:conf_multi}
  \begin{tabular}{l|l|l|l|l|l|l|l|l|l|l|l|l}
    \toprule
    \textbf{Dataset} & \textbf{Model} & \textbf{Target} & \textbf{Method} & \textbf{Accuracy} & \textbf{Confusion} & \textbf{Acc} & \textbf{Conf} & \textbf{Rank} & \textbf{Acc} & \textbf{Acc} & \textbf{Conf} & \textbf{Conf} \\ 
    & & \textbf{Classes} &  & & & \textbf{Rank} & \textbf{Rank} & \textbf{Sum} & \textbf{W} & \textbf{VD} & \textbf{W} & \textbf{VD}\\ 
    \midrule
    
    \coco* &  ResNet-50 & (person,  & orig & $0.6604$ & $0.2013$ & 5 & 5 & 10& & & &\\
          &            & bus), & orig-ft   & $0.6614\pm 0.0003$ & $0.2266\pm 0.0088$ & 1 & 6 & 7& & & &\\
          &            & (mouse, & w-aug(0.9)   & $0.6609\pm 0.0004$ & $0.2305\pm 0.0140$ & 3 & 7 & 10& & & &\\
          &            & keyboard) & \textcolor{red}{w-bn(0.7)}   & $0.6606\pm 0.0002$ & $0.1234\pm 0.0081$ & 4 & 2 & 6& 0.009 & 0.0(l) & 0.009 & 0.0(l)\\
          &            & & \textcolor{red}{w-loss(0.9)}   & $0.6611\pm 0.0003$ & $0.1307\pm 0.0158$ & 2 & 3 & 5& 0.251 & 0.280(m) & 0.009 & 0.0(l)\\
          &            & & {w-dbr(0.9)}   & $0.6589\pm 0.0002$ & $0.1629\pm 0.0068$ & 7 & 4 & 11& 0.009 & 0.0(l) & 0.009 & 0.0(l)\\
          &            & & w-os(0.7)  &  $0.6595$ & $0.0968$ & 6 & 1 & 7& 0.009 & 0.0(l) & 0.009 & 0.0(l)\\
    \midrule
    \cifars   &  ResNet-18 & (cat,  & orig &$0.8747$ & $0.0670$ & 7 & 5 & 12& & & &\\
          &            & dog), & orig-ft   &$0.8774\pm 0.0010$ & $0.0680\pm 0.0025$ & 1 & 8 & 9& & & &\\
          &            & (automobile, & w-aug(0.7) &$0.8764\pm 0.0012$ & $0.0678\pm 0.0011$ & 4 & 7 & 11& & & &\\
          &            & truck) & w-bn(0.5)   & $0.8471\pm 0.0004$ & $0.0495\pm 0.0007$ & 8 & 1 & 9& 0.009 & 0.0(l) & 0.009 & 0.0(l)\\
          &            & & \textcolor{red}{w-bn(0.9)}   & $0.8772\pm 0.0008$ & $0.0649\pm 0.0028$ & 2 & 2 & 4& & & 0.175 & 0.240(l)\\
          &            & & w-loss(0.9)   & $0.8755\pm 0.0024$ & $0.0670\pm 0.0034$ & 5 & 5 & 10& & & &\\
          &            & & \textcolor{red}{w-dbr(0.9)}  & $0.8770\pm 0.0018$ & $0.0650\pm 0.0034$ & 3 & 3 & 6& 0.465 & 0.360(s) & 0.076 & 0.160(l)\\
          &            & & w-os(0.9)  & $0.8749$ & $0.0660$ & 6 & 4 & 10& 0.009 & 0.0(l) & 0.602 & 0.400(s)\\
    \bottomrule
    \end{tabular}%
 
{\small * reported in mean average precision; \new{Acc: Accuracy; Conf: Confusion; VD: Vargha-Delaney effect size test; W: Wilcoxon rank-sum test; n: negligible; s: small; m:medium; l: large}}
\end{table}

\begin{figure}[ht]
\centering
\begin{subfigure}[b]{0.49\linewidth}
    \includegraphics[width=\linewidth]{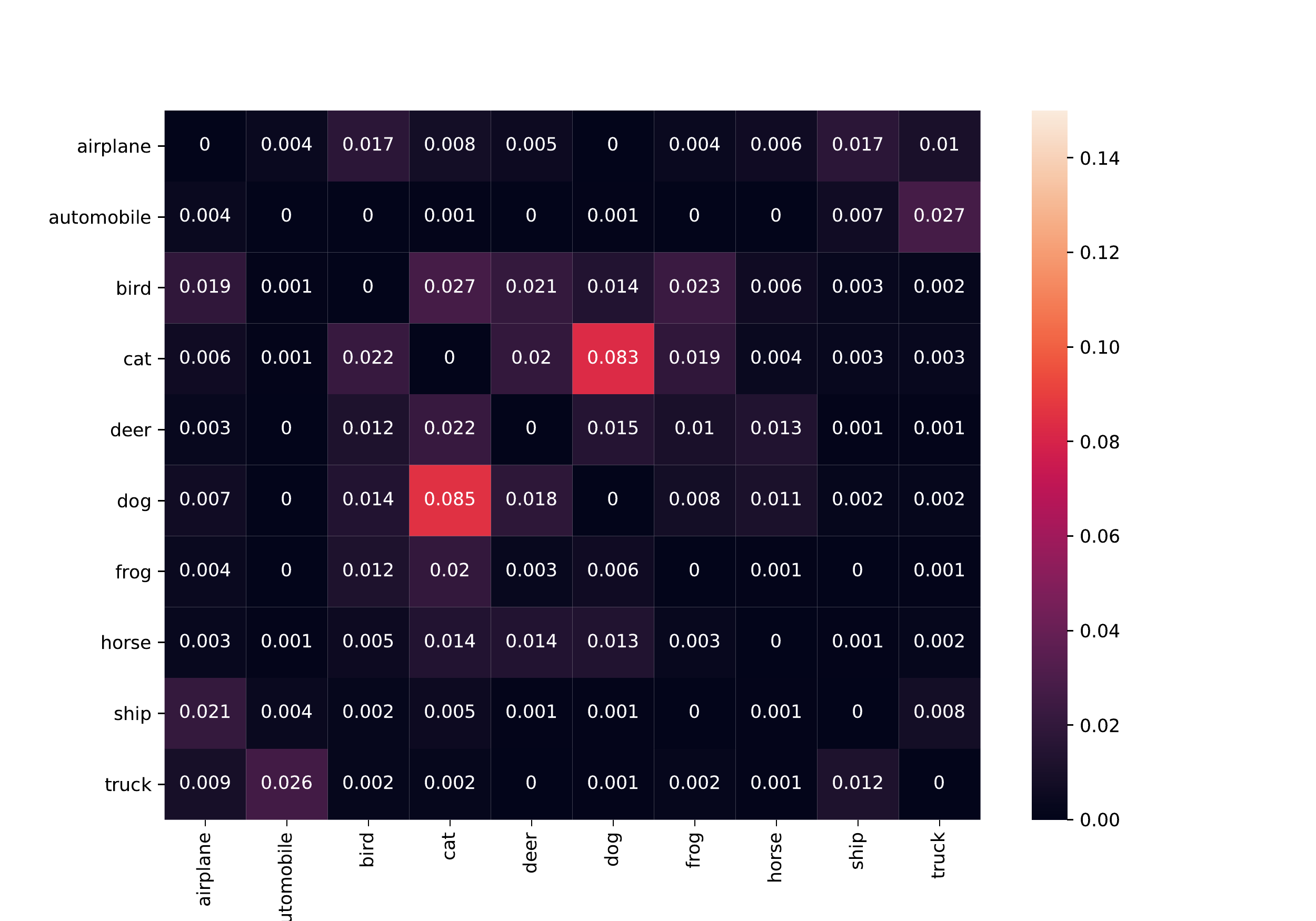}
    \caption{orig-ft(fine-tuned original model)\label{fig:confusion_matrix_origft}}
\end{subfigure}
\begin{subfigure}[b]{0.49\linewidth}
    \includegraphics[width=\linewidth]{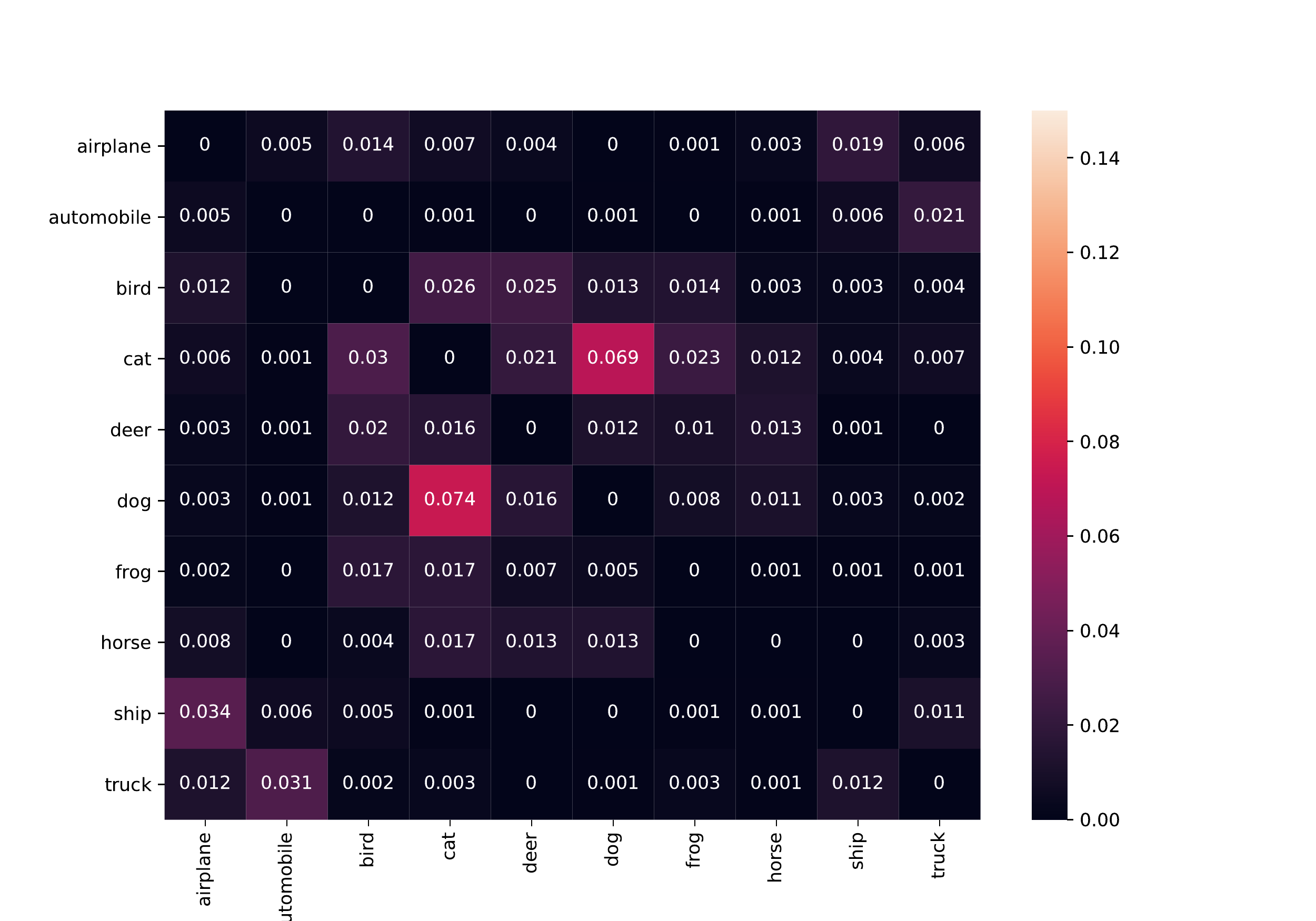}
    \caption{w-aug(0.9)\label{fig:confusion_matrix_w_aug}}
\end{subfigure}
\begin{subfigure}[b]{0.49\linewidth}
    \includegraphics[width=\linewidth]{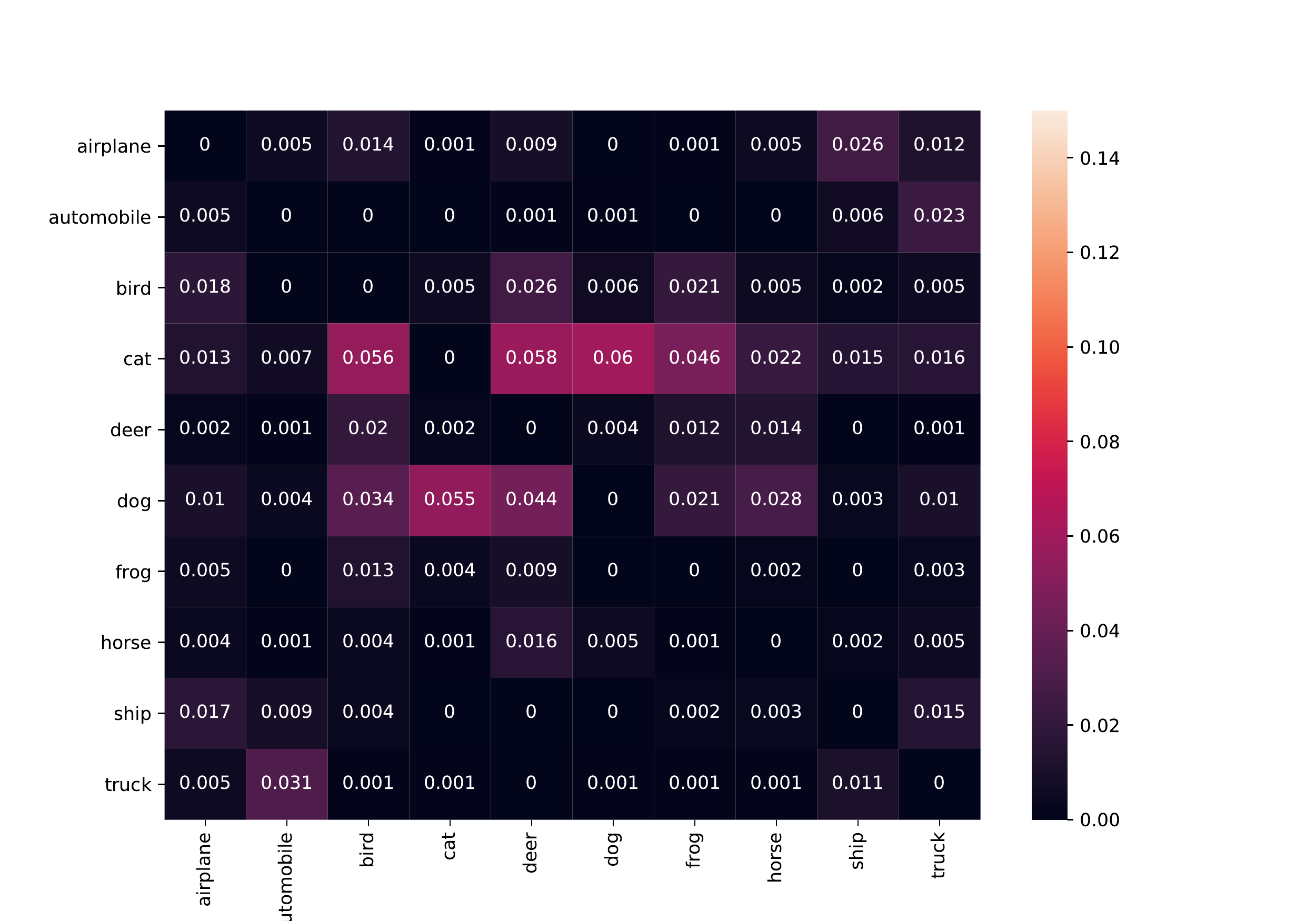}
    \caption{w-bn(0.7)\label{fig:confusion_matrix_w_bn}}
\end{subfigure}
\begin{subfigure}[b]{0.49\linewidth}
    \includegraphics[width=\linewidth]{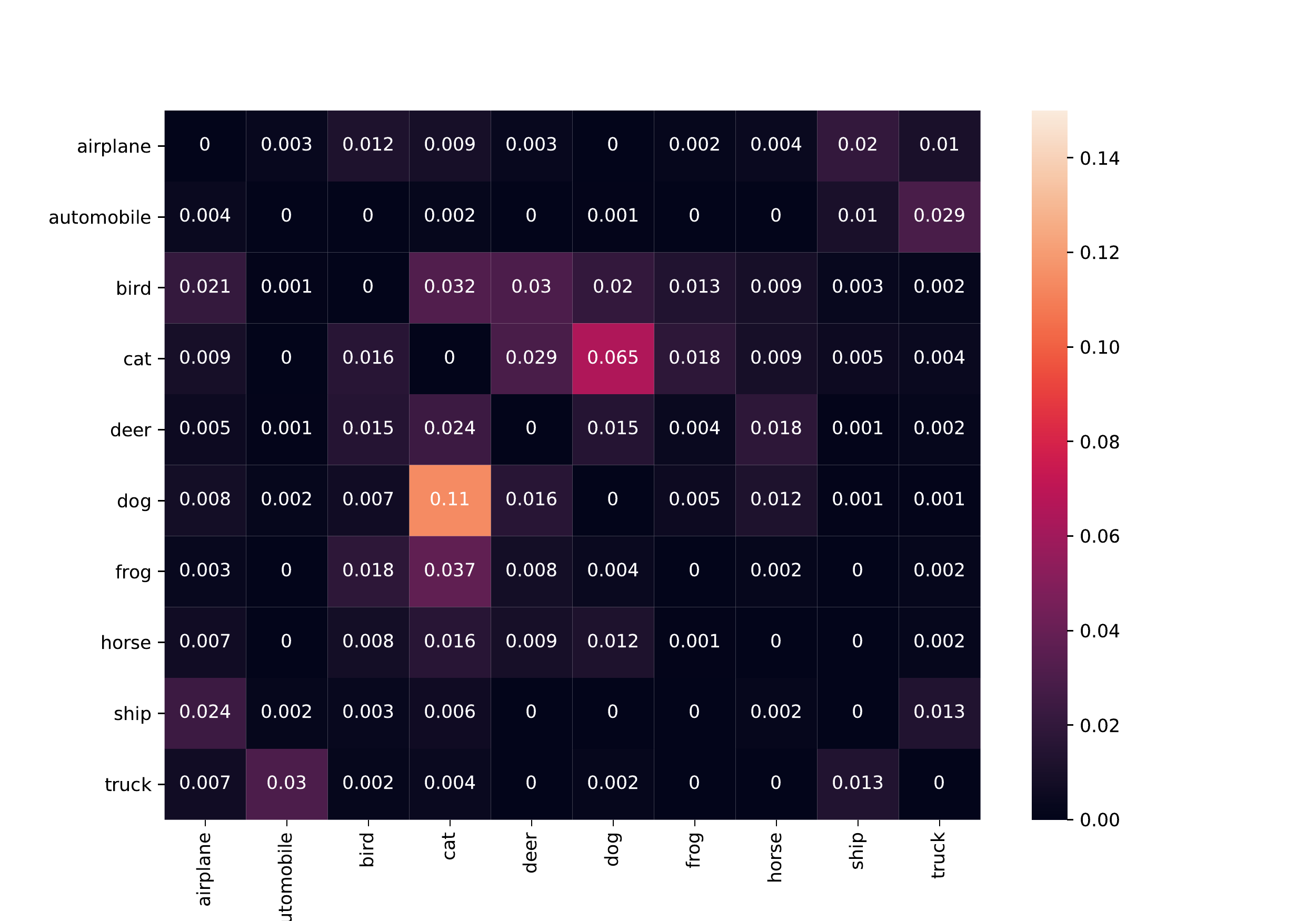}
    \caption{w-loss(0.9)\label{fig:confusion_matrix_w_loss}}
\end{subfigure}
\begin{subfigure}[b]{0.49\linewidth}
    \includegraphics[width=\linewidth]{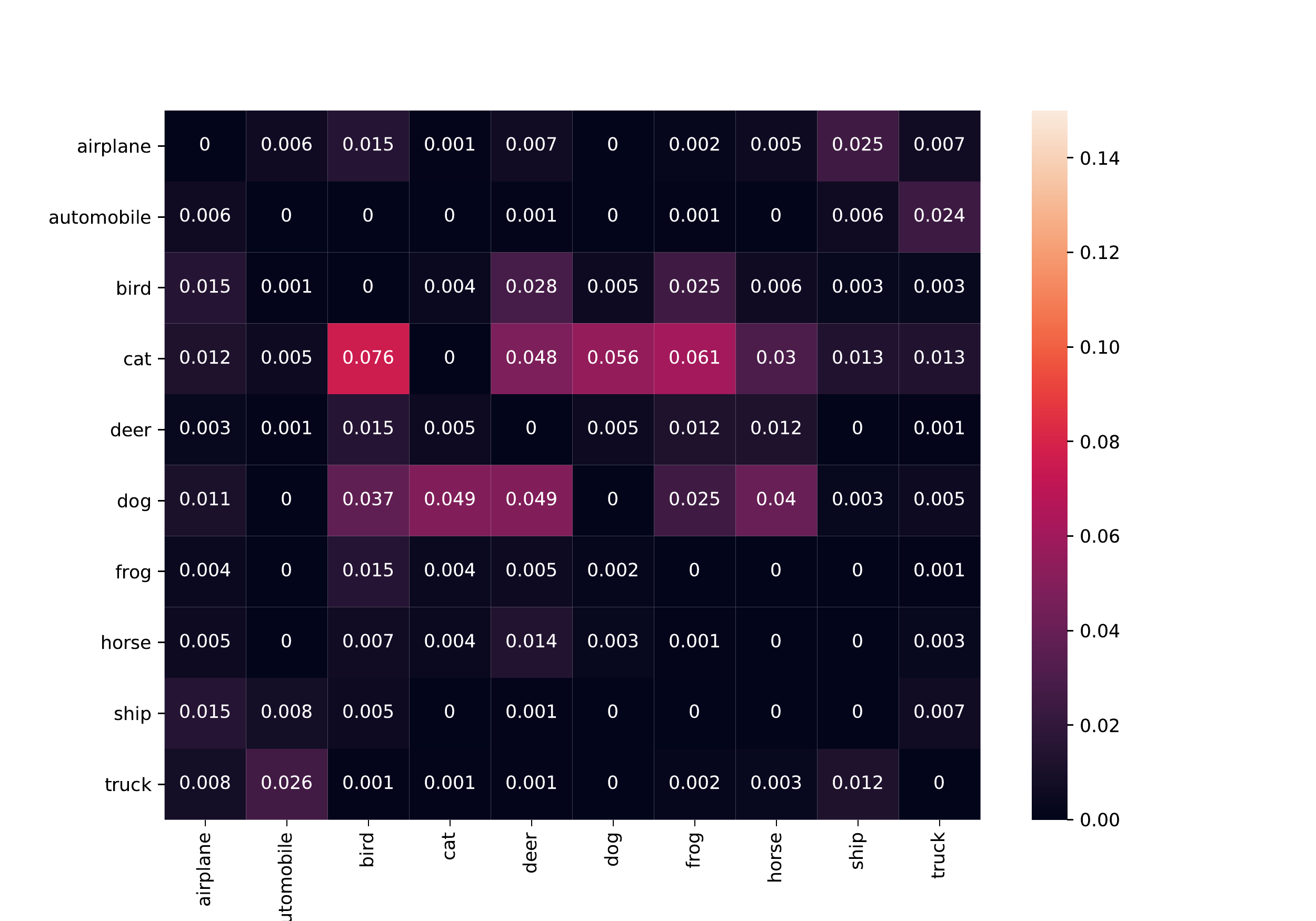}
    \caption{w-os(0.001)\label{fig:confusion_matrix_w_os}}
\end{subfigure}
\begin{subfigure}[b]{0.49\linewidth}
    \includegraphics[width=\linewidth]{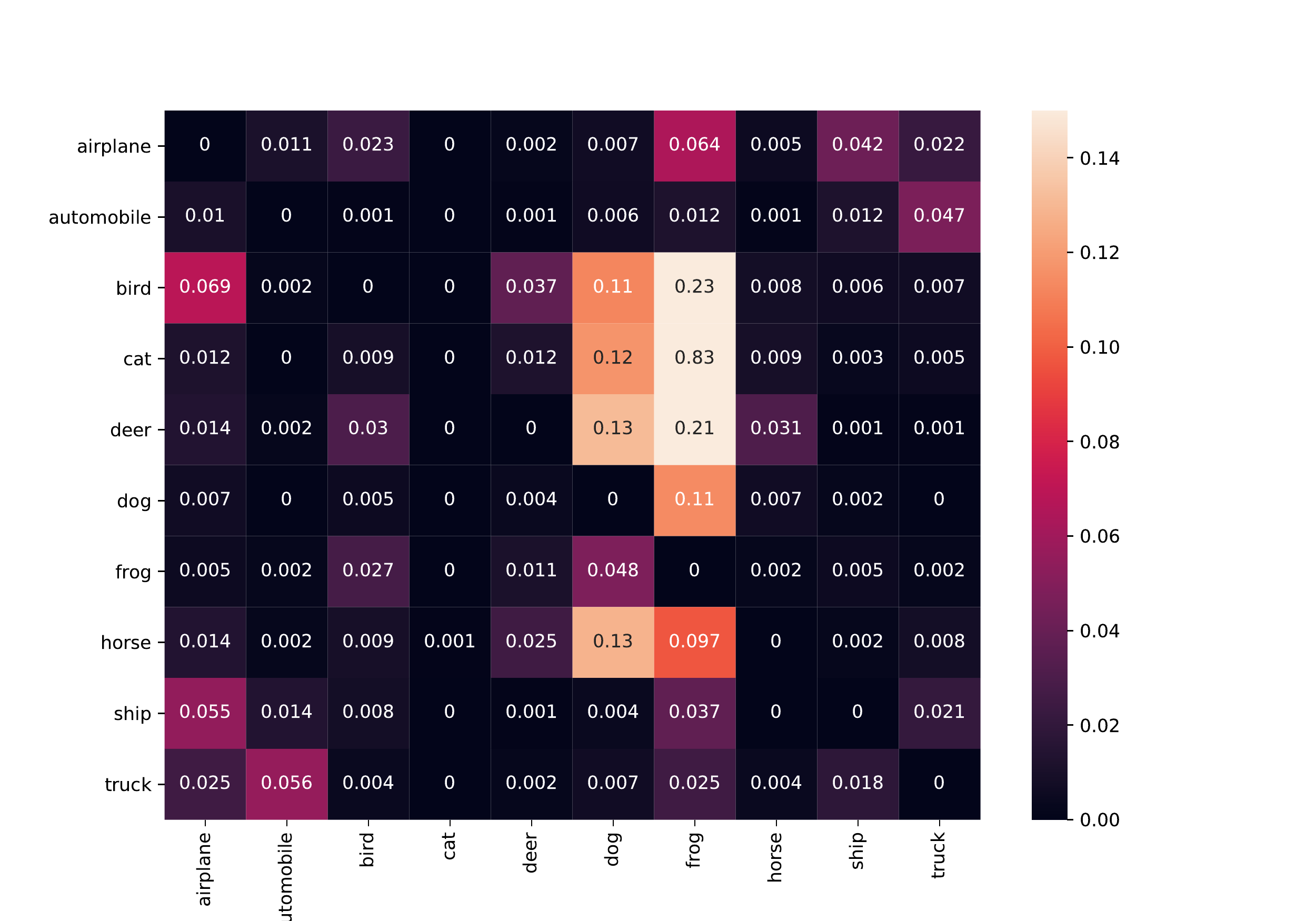}
    \caption{w-dbr(0.1)\label{fig:confusion_matrix_w_dbr}}
\end{subfigure}
\caption{\small{\textbf{Confusion matrices of the VGG11-BN after applying different methods on it.}}}
\label{fig:confusion_matrix_others}
\vspace{-2mm}
\end{figure}

\begin{figure}[ht]
\centering
\begin{subfigure}[b]{0.48\linewidth}
        \includegraphics[width=\linewidth]{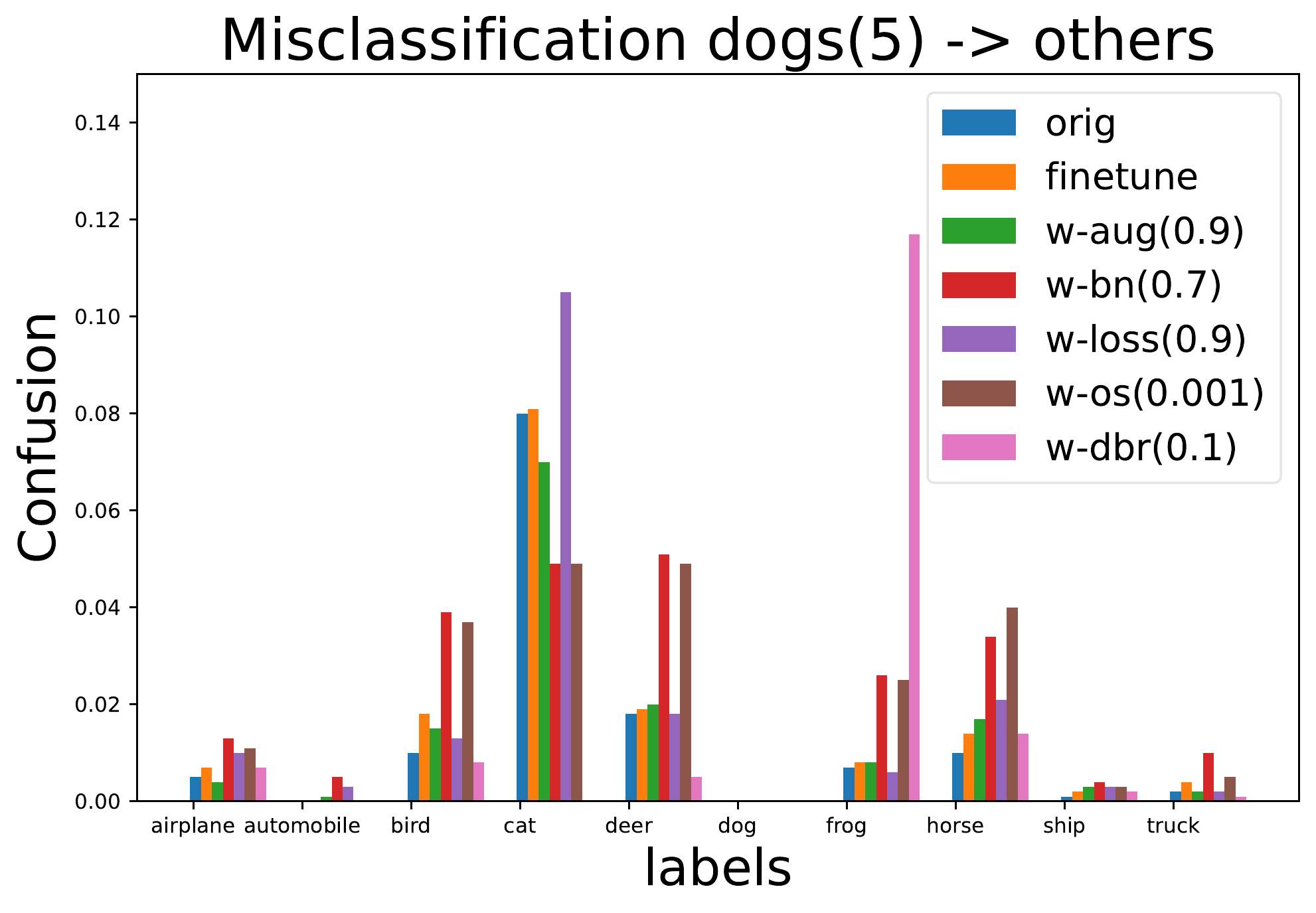}
        \caption{confusion between dog and other classes}
    \end{subfigure}
    \begin{subfigure}[b]{0.48\linewidth}
        \includegraphics[width=\linewidth]{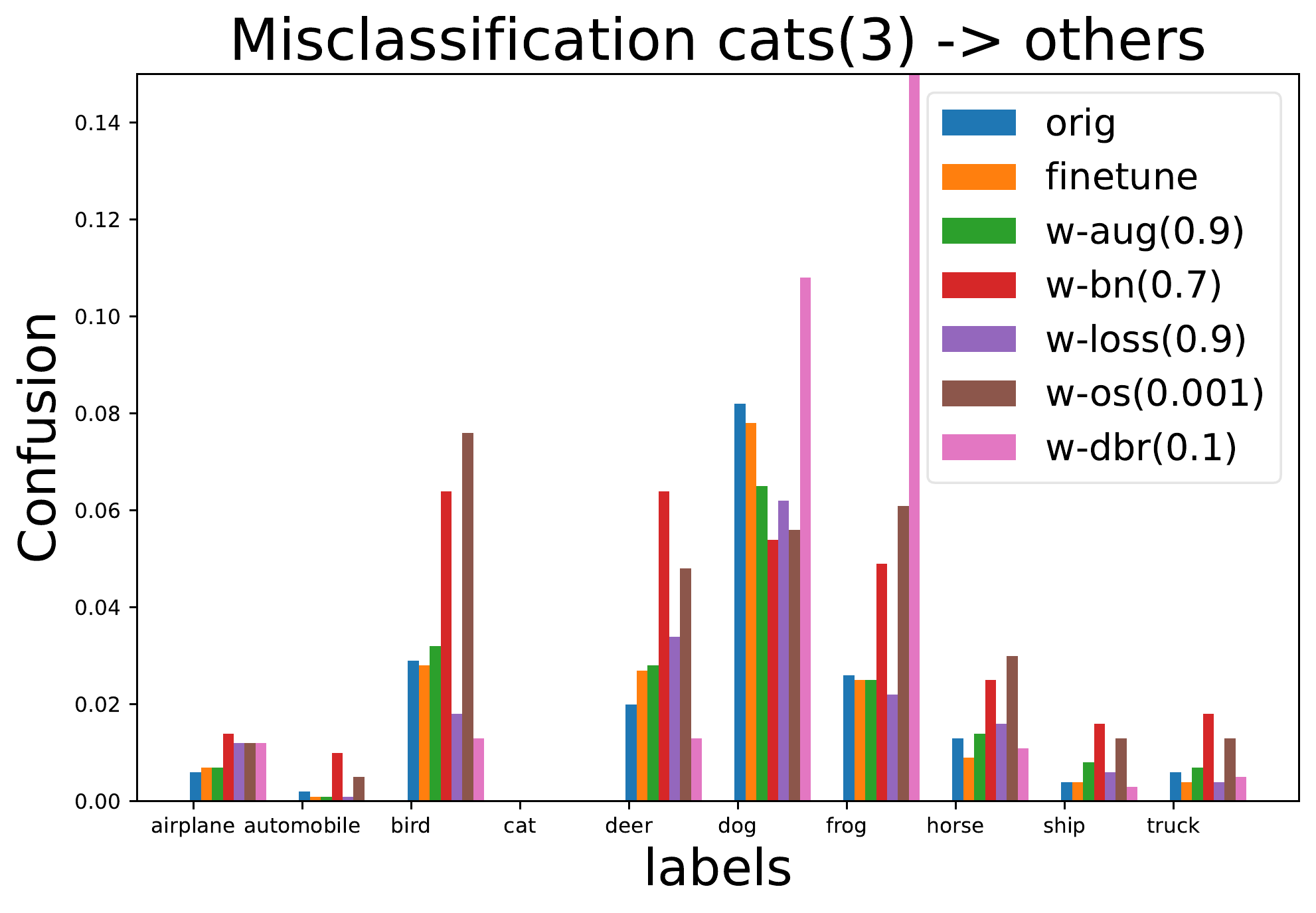}
        \caption{confusion between cat and other classes}
    \end{subfigure}
\caption{\small{\textbf{Confusion between target classes and non-target classes \new{for VGG-11 with BN} \sout{each model} on \cifars \new{after applying each method}.
}}}
\label{fig:cifar10_hist}
\vspace{-2mm}
\end{figure}

\new{Next, besides the overall accuracy and the confusion of the target pair as shown in \Cref{tab:conf} and \Cref{tab:conf_multi}, we explore more fine-grained impact of applying the proposed methods. We showed the confusion matrices for all the classes under different methods under the setting of \cifars and VGG-11 with BN in \Cref{fig:confusion_matrix_others}. In \Cref{fig:cifar10_hist}, we additionally visualize the confusion from the target classes to other classes for all the methods. In both the original model and the fine-tuned original model, dog(label 5) is highly confused with cat(label 3) than any other classes. 
Both w-os and w-bn reduce the confusion between dog and cat from both directions. As the trade-off, the confusion between dog/cat and other classes increase. This is consistent with the intuition that both methods contract the decision boundaries of the target classes. It also worth noting that w-bn provides a relatively uniform distribution of the confusion among all the pairs after the fixing. This might be a desirable property if a user does not want to overburden one particular non-target class in terms of the confusion distribution. The result also suggests a future exploration direction of our method: by adjusting hyper-parameters, one can optimize the model such that the maximum pair-wise confusion is the lowest.
w-aug similarly decreases the confusion between dog and cat but to a smaller extent. At the same time, it also increases less of the confusion for other pairs. 
Although w-dbr reduces the confusion between dog and cat, the main reduction comes from the zero confusion from dog to cat after w-dbr. The confusion from cat to dog actually increases. Besides, it increases the model's confusion from many classes to dog and frog (as shown in \Cref{fig:confusion_matrix_w_dbr}). w-loss reduces some confused instances from cat to dog, it fails to do so for instances from dog to cat.
The overall observation is that there is no free lunch for reducing the confusion for the target classes, the confusion for others has to be sacrificed to some extent.}

\sout{Next, we explore the impact of applying the proposed methods. In particular, besides their influence on the overall accuracy as shown in \Cref{tab:conf} and \Cref{tab:conf_multi}, we further explore how the confusion from the target classes to other classes change. \Cref{fig:cifar10_hist} shows the result. In the original model, dog(label 5) is highly confused with cat(label 3) than any other classes. After the fixing, the confusion between dog and cat drops slightly for w-aug and w-dbr and drops significantly for w-os and w-bn. However, the confusion between dog and other classes (also cat and other classes) increase at the same time as a trade-off for all the methods except w-aug. It also worths noting that w-bn(the purple bars) provides a uniform distribution of the confusion after fixing. This might be a desirable property if a user does not want to overburden one particular non-target class in terms of the confusion distribution. The result also suggests a future exploration direction of our method: by adjusting hyper-parameters, one can optimize the model such that the maximum pair-wise confusion is the lowest. In other words, no pair should be confused more than other pairs.}

\begin{figure}[ht]
\centering
    {\includegraphics[width=0.48\textwidth]{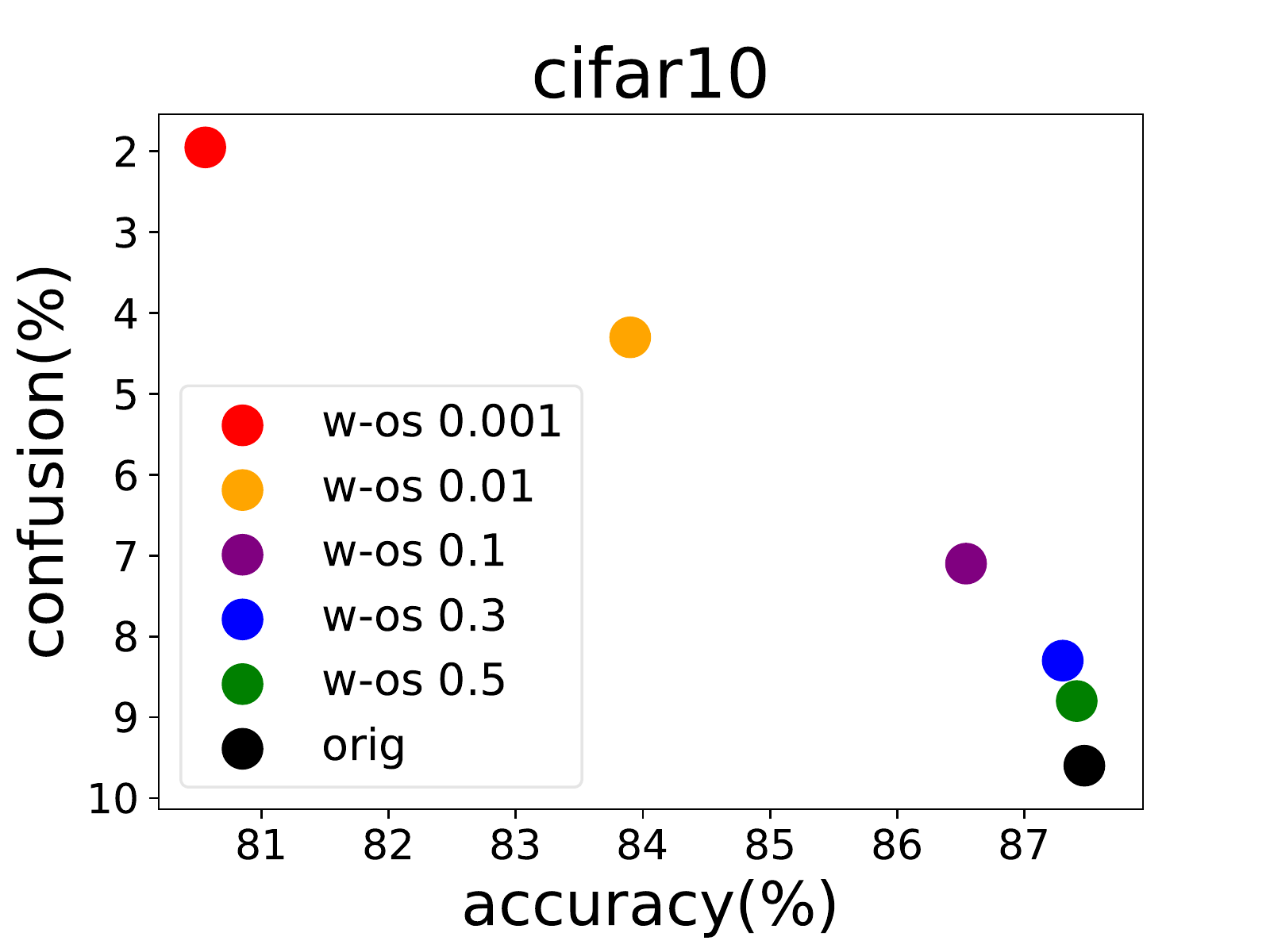}}
    {\includegraphics[width=0.48\textwidth]{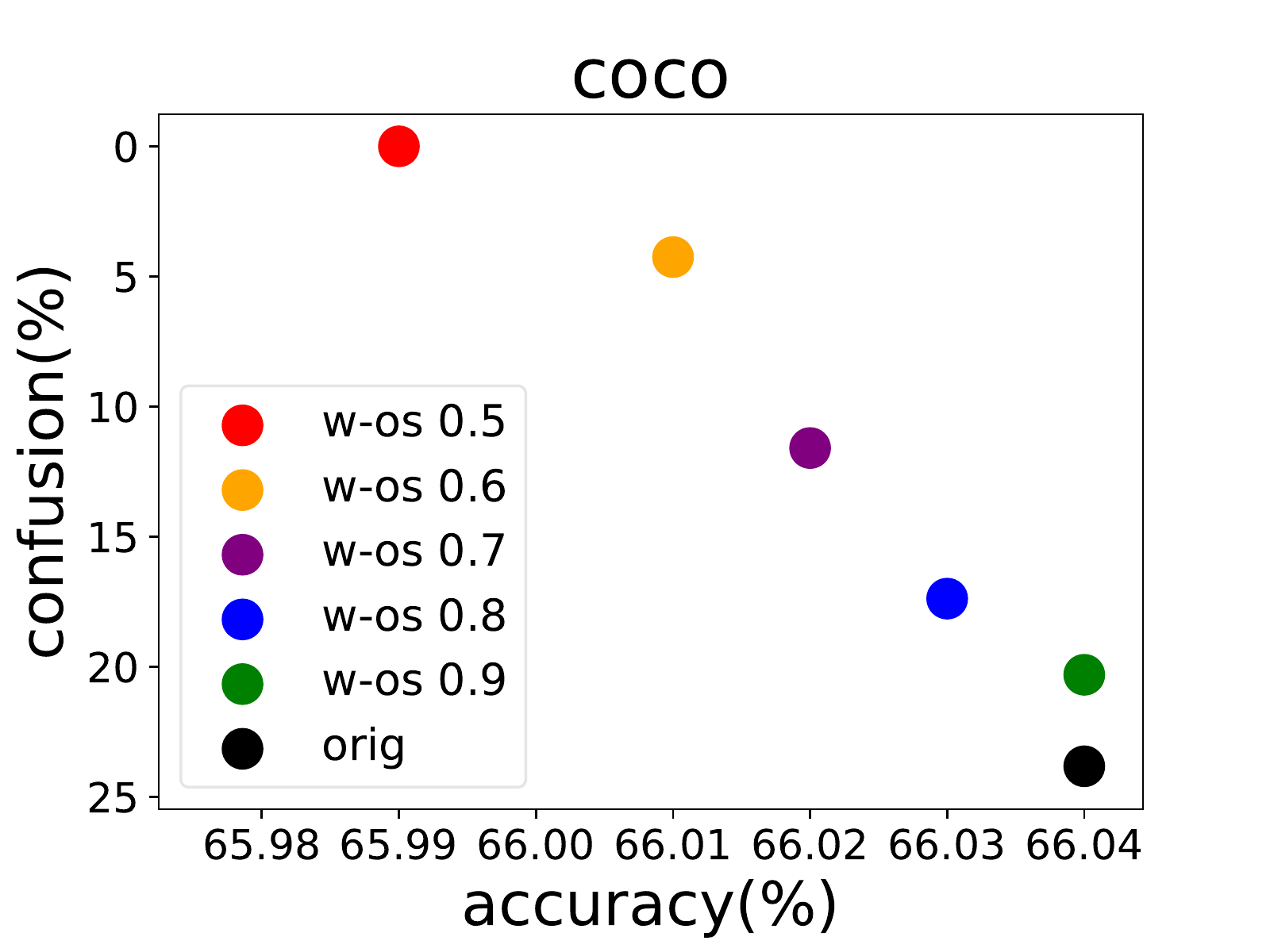}}
\caption{\small{\textbf{Accuracy and Confusion trade-off of different parameters for w-os.
}}}
\label{fig:acc_conf}
\vspace{-2mm}
\end{figure}

Since w-os can be applied when no training data is available or retraining is allowed and can reduce confusion by a significant amount \new{(>25\%)} while only slightly sacrificing the overall performance \new{(<1\%)} under every setting, we conduct a more comprehensive ablation study on its hyper-parameter $\new{\rho}\msout{\eta}$ to explore its trade-off between confusion and accuracy. \Cref{fig:acc_conf} shows the results. Note that by decreasing $\new{\rho}\msout{\eta}$, the confusion decreases and accuracy decrease at the same time. Thus, a user can decide what parameter to use depending on the significance of accuracy and confusion. \new{The influence of the hyper-parameter for other methods can also be found in \Cref{sec:hyperparam_tuning}.}

\RS{1}{The proposed method \tool can reduce confusion errors for both single-label and multi-label image classification. \new{Under every setting, at the 0.05 significance level, compared with the fine-tuned original model baseline, at least one fixing method can achieve significant lower confusion. Besides, under four out of six settings, the methods also do not have significant accuracy drop at the same time. Under the rest two settings, the sacrificed accuracy is less than 1\%.} \sout{Under every setting, one fixing method can achieve both higher accuracy and lower confusion than the original model.} The proposed method's also generalize to reduce confusion errors for multiple pairs.}

\RQ{2}{Fixing Bias Error}
\label{sec:fix_bias}
In this RQ, we explore if the proposed methods can fix bias errors. The settings \new{and procedures} are similar to those for evaluating confusion error fixing. \Cref{tab:bias} shows the results under different settings. In summary, the general trend is similar to fixing the confusion error. Under every setting, \new{at least two} \sout{many} of the proposed methods can achieve lower bias while preserving decent overall accuracy (or mean average precision).

\new{For the multi-label classification task, w-loss strikes the best trade-off between mean average precision and bias in terms of the rank sum. On both the \coco and \coco gender datasets, at the 0.05 significance level, compared with orig-ft, w-loss has lower bias with large effect size and does not have significant mean average precision drop. For example, on \coco, compared with orig-ft, w-loss(0.9) has much lower average bias between person and clock with respect to bus (0.1240 VS 0.2314) while only has 0.0001 smaller mean average precision on average.}

\sout{For the multi-label classification task, w-loss strikes the best trade-off between mean average precision and bias in terms of the rank sum. On both the \coco and \coco gender datasets, w-loss improves the mean average precision and reduce the bias compared with the original model. For example, on the COCO dataset, w-loss(0.6) improves the mean average precision from 0.6603 to 0.6611 and reduces bias between person and clock with respect to bus from 0.2366 to 0.0425; on the \coco gender dataset, w-loss(0.4) improves the mean average precision from 0.6691 to 0.6706 and reduces bias between woman and man with respect to skis from 0.2630 to 0.02472.}

\new{For the single-label classification task, w-os is ranked among the top2 under every setting. However, compared with orig-ft, at the 0.05 significance level, although w-os can always reduce bias, it has to sacrifice the accuracy as well. Under the setting of \cifar and ResNet-34, and \cifars and MobileNetv2, in order to reduce bias, any of the proposed method has to sacrifice accuracy. Under the setting of \cifars and ResNet-18, and \cifars and VGG-11 with BN, at the 0.05 significance level, w-bn(0.9) and w-aug(0.3) can reduce bias while still maintain the accuracy. For example, on \cifars and ResNet-18, compared with orig-ft, w-bn(0.9) has lower bias (0.0619 VS 0.0806) with large effect size and has only accuracy drop by 0.0002 which is not statistically significant.}

\sout{For the single-label classification task, on \cifars and ResNet-18 model, w-dbr(0.5) is ranked among the top3 while on \cifar, w-aug(5) is ranked among the top3. Both methods achieve higher accuracy and lower bias compared with the original models. In particular, on \cifars and ResNet-18 model, w-dbr(0.1) improves the overall accuracy from 0.8747 to 0.8763 and reduces the bias between dog and cat with respect to bird from 0.092 to 0.062; on \cifar, w-aug(5) improves the accuracy from 0.6961 to 0.7059 and reduces the bias between girl and boy with respect to woman from 0.09 to 0.075. On \cifars and VGG-11 with BN model, w-aug(3) has the top1 performance. It significantly decreases the bias from 0.065 to 0.04 with only a slight decrease of accuracy from 0.9175 to 0.914.} 

\new{Similar to fixing confusion errors, under all the settings, w-os and w-bn work reasonably well and give decent trade-off between accuracy and bias. w-bn can reduce bias significantly in most settings but tends to be slightly worse than w-os overall. w-loss works very well for the multi-label classification task but not for the single-label classification task. w-aug performs much better on \cifar and \cifars than on \coco and \coco gender. w-dbr can reduce bias for \coco , \coco gender and \cifar but fails to do so for \cifars.}

\sout{Similar to fixing confusion errors, under all the settings, w-os works reasonably well and gives decent trade-off between accuracy and bias. The observations for w-bn, w-loss, w-aug, and w-dbr are similar as in performance in fixing confusion errors. In particular, w-bn can reduce bias significantly in most settings but tends to be slightly worse than w-os overall. w-loss works very well for the multi-label classification task but not for the single-label classification task. w-aug performs much better on \cifar and \cifars than on \coco and \coco gender. w-dbr can reduce bias for \coco , \cifar and \cifars but fail to do so on \coco gender.}

\begin{table}[h]
  \setlength\tabcolsep{1pt} 
  \centering
  \scriptsize
  \caption{\textbf{\small Results on Bias}
  \vspace{-2mm}}
   \label{tab:bias}
  \begin{tabular}{l|l|l|l|l|l|l|l|l|l|l|l|l}
    \toprule
    \textbf{Dataset} & \textbf{Model} & \textbf{Target} & \textbf{Method} & \textbf{Accuracy} &  \textbf{Bias} & \textbf{Acc} & \textbf{Bias} & \textbf{Rank} & \textbf{Acc} & \textbf{Acc} & \textbf{Bias} & \textbf{Bias}\\ 
     &  & \textbf{Classes} & & & & \textbf{Rank} & \textbf{Rank} & \textbf{Sum} & \textbf{W} & \textbf{VD} & \textbf{W} & \textbf{VD}\\
    \midrule

    \coco* &  ResNet-50 & bus, & orig & $0.6604$ & $0.2366$ & 4 & 7 & 11 & & & &\\
          &            & person, & \textcolor{red}{orig-ft}   & $0.6614\pm 0.0003$ & $0.2314\pm 0.0084$ & 1 & 6 & 7 & & & &\\
          &            & clock & w-aug(0.9)   & $0.6607\pm 0.0005$ & $0.2483\pm 0.0209$ & 3 & 8 & 11 & & & &\\
          &            & & w-bn(0.3)   & $0.6575\pm 0.0005$ & $0.1722\pm 0.0176$ & 7 & 3 & 10 & 0.009 & 0.0(l) & 0.009 & 0.0(l)\\
          &            & & \textcolor{red}{w-loss(0.9)}   & $0.6613\pm 0.0002$ & $0.1240\pm 0.0098$ & 2 & 2 & 4 & 0.531 & 0.380(s) & 0.009 & 0.0(l)\\
          &            & & w-dbr(0.1)   & $0.6537\pm 0.0007$ & $0.1949\pm 0.0179$ & 8 & 4 & 12 & 0.009 & 0.0(l) & 0.009 & 0.0(l)\\
          &            & & \textcolor{red}{w-os(0.7)}   & $0.6602$ & $0.1151$ & 6 & 1 & 7 & 0.009 & 0.0(l) & 0.009 & 0.0(l)\\
          &            & & w-os(0.9)   & $0.6604$ & $0.2015$ & 4 & 5 & 9 & 0.009 & 0.0(l) & 0.009 & 0.0(l)\\
    \midrule
    \coco gender*  &  ResNet-50 & skis,  & orig & $0.6701$ & $0.2521$ & 4 & 7 & 11 & & & &\\
          &            & woman, & orig-ft   & $0.6709\pm 0.0002$ & $0.2501\pm 0.0153$ & 1 & 6 & 7 & & & & \\
          &            & man & w-aug(0.9)   & $0.6708\pm 0.0002$ & $0.2437\pm 0.0106$ & 2 & 5 & 7 & & & 0.465 &	0.360(s)\\
          &            & & w-bn(0.9)   & $0.6684\pm 0.0002$ & $0.1771\pm 0.0054$ & 6 & 4 & 10 & 0.009 &	0.0(l) & 0.009 & 0.0(l)\\
          &            & & \textcolor{red}{w-loss(0.9)}   & $0.6708\pm 0.0001$ & $0.1370\pm 0.0207$ & 2 & 3 & 5 & 0.296 & 0.3(m) & 0.009 & 0.0(l)\\
          &            & & w-dbr(0.3)   & $0.6677\pm 0.0003$ & $0.1311\pm 0.0564$ & 7 & 2 & 9 & 0.009 &	0.0(l) & 0.016 & 0.04(l) \\
          &            & & \textcolor{red}{w-os(0.5)}   & $0.6693$ & $0$ & 5 & 1 & 6 & 0.009 & 0.0(l) & 0.009 & 0.0(l)\\
    \midrule
    \cifar    &  ResNet-34 & woman, & orig & $0.6988$ &$0.07$ & 4 & 5 & 9 & & & &\\
          &            & girl, & \textcolor{red}{orig-ft}   & $0.7042\pm 0.0010$ &$0.0680\pm 0.0164$ & 2 & 4 & 6 & & & &\\
          &            & boy & w-aug(0.7)   &$0.7051\pm 0.0011$ &$0.0710\pm 0.0108$ & 1 & 7 & 8 & & & &\\
          &            & & w-bn(0.7)   &$0.6779\pm 0.0009$ & $0.0450\pm 0.0079$ & 7 & 1 & 8 & 0.009 & 0.0(l) & 0.028 & 0.080(l)\\
          &            & & w-bn(0.9)   &$0.7041\pm 0.0007$ & $0.0700\pm 0.0190$ & 3 & 5 & 8 & & & &\\
          &            & & w-loss(0.9)   &$0.6939\pm 0.0128$ & $0.0780\pm 0.0175$ & 5 & 8 & 13 & & & &\\
          &            & & w-dbr(0.1)   &$0.0472\pm 0.0111$ &$0.0590\pm 0.0277$ & 8 & 3 & 11 & & & 0.403 & 0.340(s)\\
          &            & & \textcolor{red}{w-os(0.01)}   &$0.6938$ &$0.045$ & 6 & 1 & 7 & 0.009 & 0.0(l) & 0.009 & 0.0(l)\\
    \midrule
    \cifars   &  ResNet-18 & dog,   & orig & $0.8747$ & $0.074$ & 5 & 7 & 12 & & & & \\
          &            & cat, & orig-ft   &$0.8776\pm 0.0011$ & $0.0806\pm 0.0044$ & 1 & 8 & 9 & & & &\\
          &            & bird & \textcolor{red}{w-aug(0.7)}   &$0.8760\pm 0.0008$ & $0.0651\pm 0.0031$ & 3 & 5 & 8 & 0.028 & 0.08(l) & 0.009 & 0.0(l)\\
          &            & & w-bn(0.7)   &$0.8617\pm 0.0015$ &$0.0553\pm 0.0023$ & 8 & 2 & 10 & 0.009 & 0.0(l) & 0.009 & 0.0(l)\\
          &            & & \textcolor{red}{w-bn(0.9)}   &$0.8774\pm 0.0020$ &$0.0619\pm 0.0030$ & 2 & 3 & 5 & 0.465 & 0.36(s) & 0.009 & 0.0(l)\\
          &            & & w-loss(0.9)   &$0.8734\pm 0.0011$ &$0.0654\pm 0.0058$ & 6 & 6 & 12 & 0.009 & 0.0(l) & 0.012 & 0.020(l)\\
          &            & & w-dbr(0.9)   &$0.8712\pm 0.0009$ &$0.0925\pm 0.0028$ & 7 & 9 & 16 & & & &\\
          &            & & w-os(0.01)   &$0.8181$ &$0.0425$ & 9 & 1 & 10 & 0.009 & 0.0(l) & 0.009 & 0.0(l)\\
          &            & & \textcolor{red}{w-os(0.9)}   &$0.8751$ &$0.063$ & 4 & 4 & 8 & 0.009 & 0.0(l) & 0.009 & 0.0(l)\\
          \cmidrule{2-13}

              &  VGG-11  & dog,  & orig  & $0.9197$ & $0.0675$ & 2 & 8 & 10 & & & &\\
          &     with BN  & cat, & orig-ft   & $0.9181\pm 0.0023$ & $0.0603\pm 0.0074$ & 4 & 6 & 10 & & & &\\
          &            & bird  & w-aug(0.3)   & $0.9171\pm 0.0015$ & $0.0497\pm 0.0055$ & 5 & 2 & 7& 0.251 & 0.28(m) & 0.047 & 0.120(l)\\
          &            & & \textcolor{red}{w-bn(0.9)}   & $0.9183\pm 0.0017$ & $0.0530\pm 0.0074$ & 3 & 3 & 6& & & 0.251 & 0.280(m)\\
          &            & & w-loss(0.9)   & $0.9122\pm 0.0009$ & $0.0602\pm 0.0039$ & 7 & 5 & 12& & & 0.917	& 0.480(n)\\
          &            & & w-dbr(0.9)   & $0.9170\pm 0.0015$ & $0.0634\pm 0.0072$ & 6 & 7 & 13 & & & &\\
          &            & & w-os(0.001)   & $0.9023$ & $0.041$ & 8 & 1 & 9& 0.009 & 0.0(l) & 0.009 & 0.0(l)\\
          &            & & \textcolor{red}{w-os(0.7)}   &$0.9198$ &$0.056$ & 1 & 4 & 5 & & & 0.117 & 0.2(l)\\
 
          \cmidrule{2-13}

              &  MobileNetv2  & dog,  & \textcolor{red}{orig}  & $0.942$ & $0.0415$ & 1 & 3 & 4 & & & &\\
          &      & cat, & \textcolor{red}{orig-ft}   & $0.9384\pm 0.0012$ & $0.0468\pm 0.0047$ & 2 & 5 & 7 & & & &\\
          &            & bird  & \textcolor{red}{w-aug(0.5)}   & $0.9383\pm 0.0010$ & $0.0418\pm 0.0018$ & 3 & 4 & 7 & & & &\\
          &            & & \textcolor{red}{w-bn(0.5)}   & $0.8981\pm 0.0017$ & $0.0272\pm 0.0025$ & 5 & 2 & 7 & 0.009 & 0.0(l) & 0.009 & 0.0(l)\\
          &            & & w-loss(0.1)   & $0.3494\pm 0.0420$ & $0.0916\pm 0.0482$ & 7 & 7 & 14 & & & &\\
          &            & & w-dbr(0.9)   & $0.9383\pm 0.0004$ & $0.0615\pm 0.0035$ & 3 & 6 & 9 & & & &\\
          &            & & \textcolor{red}{w-os(0.0001)}   & $0.8943$ & $0.0240$ & 6 & 1 & 7 & 0.009 & 0.0(l) & 0.009 & 0.0(l)\\
          
    \bottomrule
    \end{tabular}
 
{\small * reported in mean average precision; \new{Acc: Accuracy; VD: Vargha-Delaney effect size test; W: Wilcoxon rank-sum test; n: negligible; s: small; m:medium; l: large}}
\end{table}

\begin{figure}[!htpb]
\centering
    \begin{subfigure}[b]{0.36\linewidth}
        \includegraphics[width=\linewidth]{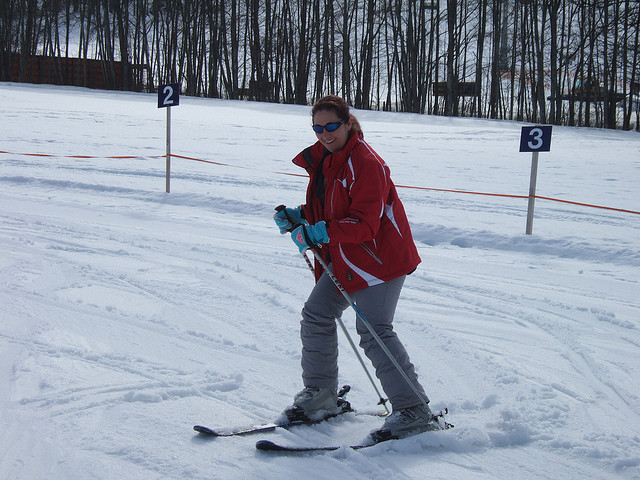}
        \caption{woman, skis}
    \end{subfigure}
    \begin{subfigure}[b]{0.36\linewidth}
        \includegraphics[width=\linewidth]{img/bias_coco_gender/woman_skis_man2.jpg}
        \caption{woman, man, skis}
    \end{subfigure}
\caption{\textbf{\small Fixed bias errors on \coco gender.}}
\label{fig:fixed_examples_bias}
\end{figure}

\Cref{fig:fixed_examples} shows two examples of the fixed bias instances \new{under the setting of \coco gender and ResNet-50}. \Cref{fig:fixed_examples_bias}(a) shows an image containing a woman and a skis but the original model classifies the woman to man. After applying w-loss, the model correctly predicts the presence of woman and skis. \Cref{fig:fixed_examples_bias}(b) shows an image containing several woman, man and skis while the original model only predicts the presence of only man and skis while missing woman. After fixing the model using w-loss, the model successfully predicts the presence woman, man and skis. 

\RS{2}{The proposed method \tool can effectively reduce bias errors for both single-label and multi-label image classification. \new{Under every setting, compared with the fine-tuned original model baseline, at least one fixing method can achieve significant lower bias. Besides, under four out of six settings, the method does not have significant accuracy drop at the same time. Under the rest two settings, significant lower bias can also be achieved but accuracy has to be sacrificed for 1\% and 5\% respectively.}\sout{Under every setting, one fixing method can achieve both higher accuracy and lower bias than the original model.}}


\section{Related Work}
\subsection{Software Repairing}
Automatic software repairing is very challenging and most of existing work focus on traditional software.\cite{monperrus2018automatic}. Traditional automatic repairing techniques include random or guided mutation of AST(Abstract Syntax Tree)\cite{forrest2009genetic, le2012systematic, weimer2010automatic, weimer2009automatically, le2011genprog, schulte2010automated, debroy2010using, nica2013use}, static program analysis or symbolic execution/concrete execution\cite{tian2017automatically, nguyen2013semfix, logozzo2012modular, logozzo2013automatic, gao2015safe, muntean2014automated}. The most recent techniques involve language models training and program synthesis\cite{gupta2017deepfix, yang2020applying}. All these techniques proposed to repair traditional programs such as C, C++, Java or Python, cannot work on DNN based software because there is no program logic or AST in DNN models. In this paper, we propose a generic method for automatic target repairing group-wise errors in DNN based software. 

\subsection{DNN Testing and Repairing}
An increasing number of works in SE for AI area focus on DNN testing and repairing. The testing techniques usually leverage metamorphic relation as oracle and coverage guided image transformation or perturbation for generating test cases\cite{pei2017deepxplore, tian2018deeptest, ma2018deepgauge, xie2019deephunter, kim2019guiding, ma2019deepct, ma2018deepmutation, deeprobust}.
Data augmentation and retraining techniques are usually proposed for repairing DNN models in improving overall accuracy\cite{ren2020few, sohn2019search, mode2018}. There are also works in improving robustness of models against adversarial instances\cite{gao2020fuzz, zhong2021understanding,wang2018mixtrain,katz2017reluplex,mirman2018differentiable,wong2018scaling,wang2018formal,rotation2019,madry2018,tramer2017ensemble,ma2018characterizing, metric_adv}. All of these papers focus on repairing instance-wise or dataset-wise errors. In contrast, our paper focuses on fixing group-wise errors.

\subsection{Fairness}
Fairness is an important problem from both a theoretical and a practical perspective~\cite{luong2011k, ZemelICML, zafar2017fairness, Brun:2018:SF:3236024.3264838}. Related works in fairness usually define a fairness criteria and optimize the original objective while satisfying the fairness criteria ~\cite{Dwork:2011, Hardt2016, barocas-hardt-narayanan, DBLP:conf/fat/MenonW18, DBLP:conf/nips/DoniniOBSP18, DBLP:journals/corr/abs-1901-10837}. These properties are defined at individual~\cite{Dwork:2011,Kusner,Kim} or group levels~\cite{Calders,Hardt2016,ZafarWWW}. Our paper focuses on fixing errors based on a group-level fairness definition called bias error proposed in \cite{tian2019deepinspect}. To the best of our knowledge, methods on repairing this bias error have not been proposed before.  

\section{Threats to Validity and Discussion}
There are many potential ways to fix group-level errors of DNNs. To mitigate this threat, we propose and compare the performance of five different methods \new{along with two baselines}. For each method, we set a parameter to make trade-off between accuracy and confusion/bias and show results of each method when using \new{at least five}\sout{two} different hyper-parameters.

There are many datasets and models, which can be used for the evaluation purpose. We choose \new{six}\sout{five} combinations of four widely used datasets and \new{five}\sout{models} for image classification. Besides, DNN training is stochastic so the results may have some fluctuations. \new{We repeat each method with top performing hyper-parameter(s) for five times and apply statistical tests to check the significance of the results.}\sout{The general patterns of the results we observe hold across different runs of the same methods.} Lastly, our method can be \new{potentially} applied to DNN models used in applications beyond image classifications such as object detection and recommendation systems. We leave that for future work.

\section{Conclusion}
In this work, we propose a generic method called Weighted Regularization(\tool) consists of five concrete methods that can fix group-level errors for DNNs with different trade-offs. To the best of our knowledge, this is the first work proposing, exploring and comparing target fixing methods, which can be applied in different stages of DNN retraining or inference, on repairing group-level DNN model errors. Our experimental results show that \tool can effectively fix confusion and bias errors and these methods all have their pros, cons and applicable scenarios.


\bibliographystyle{ACM-Reference-Format}
\bibliography{main}


\begin{thebibliography}{86}


\ifx \showCODEN    \undefined \def \showCODEN     #1{\unskip}     \fi
\ifx \showDOI      \undefined \def \showDOI       #1{#1}\fi
\ifx \showISBNx    \undefined \def \showISBNx     #1{\unskip}     \fi
\ifx \showISBNxiii \undefined \def \showISBNxiii  #1{\unskip}     \fi
\ifx \showISSN     \undefined \def \showISSN      #1{\unskip}     \fi
\ifx \showLCCN     \undefined \def \showLCCN      #1{\unskip}     \fi
\ifx \shownote     \undefined \def \shownote      #1{#1}          \fi
\ifx \showarticletitle \undefined \def \showarticletitle #1{#1}   \fi
\ifx \showURL      \undefined \def \showURL       {\relax}        \fi
\providecommand\bibfield[2]{#2}
\providecommand\bibinfo[2]{#2}
\providecommand\natexlab[1]{#1}
\providecommand\showeprint[2][]{arXiv:#2}

\bibitem[\protect\citeauthoryear{??}{bat}{2018}]%
        {batchnorm}
 \bibinfo{year}{2018}\natexlab{}.
\newblock \bibinfo{title}{Pytorch BATCHNORM2D}.
\newblock
  \bibinfo{howpublished}{\url{https://pytorch.org/docs/stable/generated/torch.nn.BatchNorm2d.html}}.
\newblock


\bibitem[\protect\citeauthoryear{Agarwal, Beygelzimer, Dudík, Langford, and
  Wallach}{Agarwal et~al\mbox{.}}{2018}]%
        {reduction18}
\bibfield{author}{\bibinfo{person}{Alekh Agarwal}, \bibinfo{person}{Alina
  Beygelzimer}, \bibinfo{person}{Miroslav Dudík}, \bibinfo{person}{John
  Langford}, {and} \bibinfo{person}{Hanna Wallach}.}
  \bibinfo{year}{2018}\natexlab{}.
\newblock \showarticletitle{A Reductions Approach to Fair Classification}. In
  \bibinfo{booktitle}{\emph{Proceedings of the 35th International Conference on
  Machine Learning}}, \bibfield{editor}{\bibinfo{person}{Jennifer~G. Dy} {and}
  \bibinfo{person}{Andreas Krause}} (Eds.), Vol.~\bibinfo{volume}{80}.
  \bibinfo{publisher}{JMLR}, \bibinfo{address}{Stanford, CA},
  \bibinfo{pages}{60--69}.
\newblock


\bibitem[\protect\citeauthoryear{Arcuri and Briand}{Arcuri and Briand}{2014}]%
        {guidetostatstest}
\bibfield{author}{\bibinfo{person}{Andrea Arcuri} {and} \bibinfo{person}{Lionel
  Briand}.} \bibinfo{year}{2014}\natexlab{}.
\newblock \showarticletitle{A Hitchhiker's guide to statistical tests for
  assessing randomized algorithms in software engineering}.
\newblock \bibinfo{journal}{\emph{Software Testing, Verification and
  Reliability}} \bibinfo{volume}{24}, \bibinfo{number}{3}
  (\bibinfo{year}{2014}), \bibinfo{pages}{219--250}.
\newblock
\urldef\tempurl%
\url{https://doi.org/10.1002/stvr.1486}
\showDOI{\tempurl}


\bibitem[\protect\citeauthoryear{Barocas, Hardt, and Narayanan}{Barocas
  et~al\mbox{.}}{2018}]%
        {barocas-hardt-narayanan}
\bibfield{author}{\bibinfo{person}{Solon Barocas}, \bibinfo{person}{Moritz
  Hardt}, {and} \bibinfo{person}{Arvind Narayanan}.}
  \bibinfo{year}{2018}\natexlab{}.
\newblock \bibinfo{booktitle}{\emph{Fairness and Machine Learning}}.
\newblock \bibinfo{publisher}{fairmlbook.org}.
\newblock
\newblock
\shownote{\url{http://www.fairmlbook.org}}.


\bibitem[\protect\citeauthoryear{Bengio, Courville, and Vincent}{Bengio
  et~al\mbox{.}}{2013}]%
        {bengio2013representation}
\bibfield{author}{\bibinfo{person}{Yoshua Bengio}, \bibinfo{person}{Aaron
  Courville}, {and} \bibinfo{person}{Pascal Vincent}.}
  \bibinfo{year}{2013}\natexlab{}.
\newblock \showarticletitle{Representation learning: A review and new
  perspectives}.
\newblock \bibinfo{journal}{\emph{IEEE transactions on pattern analysis and
  machine intelligence}} \bibinfo{volume}{35}, \bibinfo{number}{8}
  (\bibinfo{year}{2013}), \bibinfo{pages}{1798--1828}.
\newblock


\bibitem[\protect\citeauthoryear{Brun and Meliou}{Brun and Meliou}{2018}]%
        {Brun:2018:SF:3236024.3264838}
\bibfield{author}{\bibinfo{person}{Yuriy Brun} {and} \bibinfo{person}{Alexandra
  Meliou}.} \bibinfo{year}{2018}\natexlab{}.
\newblock \showarticletitle{Software Fairness}. In
  \bibinfo{booktitle}{\emph{Proceedings of the 2018 26th ACM Joint Meeting on
  European Software Engineering Conference and Symposium on the Foundations of
  Software Engineering}} (Lake Buena Vista, FL, USA)
  \emph{(\bibinfo{series}{ESEC/FSE 2018})}. \bibinfo{publisher}{ACM},
  \bibinfo{address}{New York, NY, USA}, \bibinfo{pages}{754--759}.
\newblock
\showISBNx{978-1-4503-5573-5}
\urldef\tempurl%
\url{https://doi.org/10.1145/3236024.3264838}
\showDOI{\tempurl}


\bibitem[\protect\citeauthoryear{Buda, Maki, and Mazurowski}{Buda
  et~al\mbox{.}}{2018}]%
        {oversampling18}
\bibfield{author}{\bibinfo{person}{Mateusz Buda}, \bibinfo{person}{Atsuto
  Maki}, {and} \bibinfo{person}{Maciej~A. Mazurowski}.}
  \bibinfo{year}{2018}\natexlab{}.
\newblock \showarticletitle{A systematic study of the class imbalance problem
  in convolutional neural networks}.
\newblock \bibinfo{journal}{\emph{Neural Networks}}  \bibinfo{volume}{106}
  (\bibinfo{year}{2018}), \bibinfo{pages}{249--259}.
\newblock
\showISSN{0893-6080}
\urldef\tempurl%
\url{https://doi.org/10.1016/j.neunet.2018.07.011}
\showDOI{\tempurl}


\bibitem[\protect\citeauthoryear{Calders, Kamiran, and Pechenizkiy}{Calders
  et~al\mbox{.}}{2009}]%
        {Calders}
\bibfield{author}{\bibinfo{person}{T. Calders}, \bibinfo{person}{F. Kamiran},
  {and} \bibinfo{person}{M. Pechenizkiy}.} \bibinfo{year}{2009}\natexlab{}.
\newblock \showarticletitle{Building Classifiers with Independency
  Constraints}. In \bibinfo{booktitle}{\emph{2009 IEEE International Conference
  on Data Mining Workshops}}. \bibinfo{pages}{13--18}.
\newblock


\bibitem[\protect\citeauthoryear{Capon.}{Capon.}{1991}]%
        {Wilcoxon}
\bibfield{author}{\bibinfo{person}{J.~Anthony Capon.}}
  \bibinfo{year}{1991}\natexlab{}.
\newblock \showarticletitle{Elementary Statistics for the Social Sciences:
  Study Guide}.
\newblock  (\bibinfo{year}{1991}).
\newblock


\bibitem[\protect\citeauthoryear{Cui, Jia, Lin, Song, and Belongie}{Cui
  et~al\mbox{.}}{2019}]%
        {classbalancedloss19}
\bibfield{author}{\bibinfo{person}{Yin Cui}, \bibinfo{person}{Menglin Jia},
  \bibinfo{person}{Tsung-Yi Lin}, \bibinfo{person}{Yang Song}, {and}
  \bibinfo{person}{Serge Belongie}.} \bibinfo{year}{2019}\natexlab{}.
\newblock \showarticletitle{Class-Balanced Loss Based on Effective Number of
  Samples}. In \bibinfo{booktitle}{\emph{CVPR}}.
\newblock


\bibitem[\protect\citeauthoryear{Debroy and Wong}{Debroy and Wong}{2010}]%
        {debroy2010using}
\bibfield{author}{\bibinfo{person}{Vidroha Debroy} {and}
  \bibinfo{person}{W~Eric Wong}.} \bibinfo{year}{2010}\natexlab{}.
\newblock \showarticletitle{Using mutation to automatically suggest fixes for
  faulty programs}. In \bibinfo{booktitle}{\emph{2010 Third International
  Conference on Software Testing, Verification and Validation}}. IEEE,
  \bibinfo{pages}{65--74}.
\newblock


\bibitem[\protect\citeauthoryear{Donini, Oneto, Ben{-}David, Shawe{-}Taylor,
  and Pontil}{Donini et~al\mbox{.}}{2018}]%
        {DBLP:conf/nips/DoniniOBSP18}
\bibfield{author}{\bibinfo{person}{Michele Donini}, \bibinfo{person}{Luca
  Oneto}, \bibinfo{person}{Shai Ben{-}David}, \bibinfo{person}{John
  Shawe{-}Taylor}, {and} \bibinfo{person}{Massimiliano Pontil}.}
  \bibinfo{year}{2018}\natexlab{}.
\newblock \showarticletitle{Empirical Risk Minimization Under Fairness
  Constraints}. In \bibinfo{booktitle}{\emph{Advances in Neural Information
  Processing Systems 31: Annual Conference on Neural Information Processing
  Systems 2018, NeurIPS 2018, 3-8 December 2018, Montr{\'{e}}al, Canada.}}
  \bibinfo{pages}{2796--2806}.
\newblock
\urldef\tempurl%
\url{http://papers.nips.cc/paper/7544-empirical-risk-minimization-under-fairness-constraints}
\showURL{%
\tempurl}


\bibitem[\protect\citeauthoryear{Dwork, Hardt, Pitassi, Reingold, and
  Zemel}{Dwork et~al\mbox{.}}{2012}]%
        {Dwork:2011}
\bibfield{author}{\bibinfo{person}{Cynthia Dwork}, \bibinfo{person}{Moritz
  Hardt}, \bibinfo{person}{Toniann Pitassi}, \bibinfo{person}{Omer Reingold},
  {and} \bibinfo{person}{Richard~S. Zemel}.} \bibinfo{year}{2012}\natexlab{}.
\newblock \showarticletitle{Fairness Through Awareness}.
\newblock \bibinfo{journal}{\emph{In Proceedings of the Innovations in
  Theoretical Computer Science Conference}}  \bibinfo{volume}{abs/1104.3913}
  (\bibinfo{year}{2012}), \bibinfo{pages}{214--226}.
\newblock


\bibitem[\protect\citeauthoryear{Engstrom, Tran, Tsipras, Schmidt, and
  Madry}{Engstrom et~al\mbox{.}}{2019a}]%
        {engstrom2019exploring}
\bibfield{author}{\bibinfo{person}{Logan Engstrom}, \bibinfo{person}{Brandon
  Tran}, \bibinfo{person}{Dimitris Tsipras}, \bibinfo{person}{Ludwig Schmidt},
  {and} \bibinfo{person}{Aleksander Madry}.} \bibinfo{year}{2019}\natexlab{a}.
\newblock \showarticletitle{Exploring the landscape of spatial robustness}. In
  \bibinfo{booktitle}{\emph{International Conference on Machine Learning}}.
  \bibinfo{pages}{1802--1811}.
\newblock


\bibitem[\protect\citeauthoryear{Engstrom, Tran, Tsipras, Schmidt, and
  Madry}{Engstrom et~al\mbox{.}}{2019b}]%
        {rotation2019}
\bibfield{author}{\bibinfo{person}{Logan Engstrom}, \bibinfo{person}{Brandon
  Tran}, \bibinfo{person}{Dimitris Tsipras}, \bibinfo{person}{Ludwig Schmidt},
  {and} \bibinfo{person}{Aleksander Madry}.} \bibinfo{year}{2019}\natexlab{b}.
\newblock \showarticletitle{A Rotation and a Translation Suffice: Fooling CNNs
  with Simple Transformations}. In \bibinfo{booktitle}{\emph{Proceedings of the
  36th international conference on machine learning (ICML)}}.
\newblock


\bibitem[\protect\citeauthoryear{Evtimov, Eykholt, Fernandes, Kohno, Li,
  Prakash, Rahmati, and Song}{Evtimov et~al\mbox{.}}{2017}]%
        {evtimov2017robust}
\bibfield{author}{\bibinfo{person}{Ivan Evtimov}, \bibinfo{person}{Kevin
  Eykholt}, \bibinfo{person}{Earlence Fernandes}, \bibinfo{person}{Tadayoshi
  Kohno}, \bibinfo{person}{Bo Li}, \bibinfo{person}{Atul Prakash},
  \bibinfo{person}{Amir Rahmati}, {and} \bibinfo{person}{Dawn Song}.}
  \bibinfo{year}{2017}\natexlab{}.
\newblock \showarticletitle{Robust Physical-World Attacks on Machine Learning
  Models}.
\newblock \bibinfo{journal}{\emph{arXiv preprint arXiv:1707.08945}}
  (\bibinfo{year}{2017}).
\newblock


\bibitem[\protect\citeauthoryear{Forrest, Nguyen, Weimer, and Le~Goues}{Forrest
  et~al\mbox{.}}{2009}]%
        {forrest2009genetic}
\bibfield{author}{\bibinfo{person}{Stephanie Forrest}, \bibinfo{person}{ThanhVu
  Nguyen}, \bibinfo{person}{Westley Weimer}, {and} \bibinfo{person}{Claire
  Le~Goues}.} \bibinfo{year}{2009}\natexlab{}.
\newblock \showarticletitle{A genetic programming approach to automated
  software repair}. In \bibinfo{booktitle}{\emph{Proceedings of the 11th Annual
  conference on Genetic and evolutionary computation}}.
  \bibinfo{pages}{947--954}.
\newblock


\bibitem[\protect\citeauthoryear{Gao, Xiong, Mi, Zhang, Yang, Zhou, Xie, and
  Mei}{Gao et~al\mbox{.}}{2015}]%
        {gao2015safe}
\bibfield{author}{\bibinfo{person}{Qing Gao}, \bibinfo{person}{Yingfei Xiong},
  \bibinfo{person}{Yaqing Mi}, \bibinfo{person}{Lu Zhang},
  \bibinfo{person}{Weikun Yang}, \bibinfo{person}{Zhaoping Zhou},
  \bibinfo{person}{Bing Xie}, {and} \bibinfo{person}{Hong Mei}.}
  \bibinfo{year}{2015}\natexlab{}.
\newblock \showarticletitle{Safe memory-leak fixing for c programs}. In
  \bibinfo{booktitle}{\emph{2015 IEEE/ACM 37th IEEE International Conference on
  Software Engineering}}, Vol.~\bibinfo{volume}{1}. IEEE,
  \bibinfo{pages}{459--470}.
\newblock


\bibitem[\protect\citeauthoryear{Gao, Saha, Prasad, and Roychoudhury}{Gao
  et~al\mbox{.}}{2020}]%
        {gao2020fuzz}
\bibfield{author}{\bibinfo{person}{Xiang Gao}, \bibinfo{person}{Ripon~K Saha},
  \bibinfo{person}{Mukul~R Prasad}, {and} \bibinfo{person}{Abhik
  Roychoudhury}.} \bibinfo{year}{2020}\natexlab{}.
\newblock \showarticletitle{Fuzz testing based data augmentation to improve
  robustness of deep neural networks}. In \bibinfo{booktitle}{\emph{2020
  IEEE/ACM 42nd International Conference on Software Engineering (ICSE)}}.
  IEEE, \bibinfo{pages}{1147--1158}.
\newblock


\bibitem[\protect\citeauthoryear{Goodfellow, Bengio, Courville, and
  Bengio}{Goodfellow et~al\mbox{.}}{2016}]%
        {goodfellow2016deep}
\bibfield{author}{\bibinfo{person}{Ian Goodfellow}, \bibinfo{person}{Yoshua
  Bengio}, \bibinfo{person}{Aaron Courville}, {and} \bibinfo{person}{Yoshua
  Bengio}.} \bibinfo{year}{2016}\natexlab{}.
\newblock \bibinfo{booktitle}{\emph{Deep learning}}. Vol.~\bibinfo{volume}{1}.
\newblock \bibinfo{publisher}{MIT press Cambridge}.
\newblock


\bibitem[\protect\citeauthoryear{Grush}{Grush}{2015}]%
        {google_tag}
\bibfield{author}{\bibinfo{person}{Loren Grush}.}
  \bibinfo{year}{2015}\natexlab{}.
\newblock \showarticletitle{Google engineer apologizes after Photos app tags
  two black people as gorillas}.
\newblock  (\bibinfo{year}{2015}).
\newblock
\urldef\tempurl%
\url{https://www.theverge.com/2015/7/1/8880363/google-apologizes-photos-app-tags-two-black-people-gorillas}
\showURL{%
\tempurl}


\bibitem[\protect\citeauthoryear{Guo, Pleiss, Sun, and Weinberger}{Guo
  et~al\mbox{.}}{2017}]%
        {temperaturescaling17}
\bibfield{author}{\bibinfo{person}{Chuan Guo}, \bibinfo{person}{Geoff Pleiss},
  \bibinfo{person}{Yu Sun}, {and} \bibinfo{person}{Kilian~Q. Weinberger}.}
  \bibinfo{year}{2017}\natexlab{}.
\newblock \showarticletitle{On Calibration of Modern Neural Networks}. In
  \bibinfo{booktitle}{\emph{Proceedings of the 34th International Conference on
  Machine Learning - Volume 70}} (Sydney, NSW, Australia)
  \emph{(\bibinfo{series}{ICML'17})}. \bibinfo{publisher}{JMLR.org},
  \bibinfo{pages}{1321–1330}.
\newblock


\bibitem[\protect\citeauthoryear{Gupta, Pal, Kanade, and Shevade}{Gupta
  et~al\mbox{.}}{2017}]%
        {gupta2017deepfix}
\bibfield{author}{\bibinfo{person}{Rahul Gupta}, \bibinfo{person}{Soham Pal},
  \bibinfo{person}{Aditya Kanade}, {and} \bibinfo{person}{Shirish Shevade}.}
  \bibinfo{year}{2017}\natexlab{}.
\newblock \showarticletitle{Deepfix: Fixing common c language errors by deep
  learning}. In \bibinfo{booktitle}{\emph{Proceedings of the aaai conference on
  artificial intelligence}}, Vol.~\bibinfo{volume}{31}.
\newblock


\bibitem[\protect\citeauthoryear{Hardt, Price, and Srebro}{Hardt
  et~al\mbox{.}}{2016a}]%
        {fair_postprocess}
\bibfield{author}{\bibinfo{person}{Moritz Hardt}, \bibinfo{person}{Eric Price},
  {and} \bibinfo{person}{Nathan Srebro}.} \bibinfo{year}{2016}\natexlab{a}.
\newblock \showarticletitle{Equality of Opportunity in Supervised Learning}. In
  \bibinfo{booktitle}{\emph{Proceedings of the 30th International Conference on
  Neural Information Processing Systems}} (Barcelona, Spain)
  \emph{(\bibinfo{series}{NIPS'16})}. \bibinfo{publisher}{Curran Associates
  Inc.}, \bibinfo{address}{Red Hook, NY, USA}, \bibinfo{pages}{3323–3331}.
\newblock
\showISBNx{9781510838819}


\bibitem[\protect\citeauthoryear{Hardt, Price, and Srebro}{Hardt
  et~al\mbox{.}}{2016b}]%
        {Hardt2016}
\bibfield{author}{\bibinfo{person}{Moritz Hardt}, \bibinfo{person}{Eric Price},
  {and} \bibinfo{person}{Nathan Srebro}.} \bibinfo{year}{2016}\natexlab{b}.
\newblock \showarticletitle{Equality of Opportunity in Supervised Learning}. In
  \bibinfo{booktitle}{\emph{Proceedings of the 30th International Conference on
  Neural Information Processing Systems}} (Barcelona, Spain)
  \emph{(\bibinfo{series}{NIPS'16})}. \bibinfo{address}{USA},
  \bibinfo{pages}{3323--3331}.
\newblock
\showISBNx{978-1-5108-3881-9}


\bibitem[\protect\citeauthoryear{He, Zhang, Ren, and Sun}{He
  et~al\mbox{.}}{2016}]%
        {he2016deep}
\bibfield{author}{\bibinfo{person}{Kaiming He}, \bibinfo{person}{Xiangyu
  Zhang}, \bibinfo{person}{Shaoqing Ren}, {and} \bibinfo{person}{Jian Sun}.}
  \bibinfo{year}{2016}\natexlab{}.
\newblock \showarticletitle{Deep residual learning for image recognition}. In
  \bibinfo{booktitle}{\emph{Proceedings of the IEEE conference on computer
  vision and pattern recognition}}. \bibinfo{pages}{770--778}.
\newblock


\bibitem[\protect\citeauthoryear{Ioffe and Szegedy}{Ioffe and Szegedy}{2015}]%
        {ioffe2015batch}
\bibfield{author}{\bibinfo{person}{Sergey Ioffe} {and}
  \bibinfo{person}{Christian Szegedy}.} \bibinfo{year}{2015}\natexlab{}.
\newblock \showarticletitle{Batch normalization: Accelerating deep network
  training by reducing internal covariate shift}. In
  \bibinfo{booktitle}{\emph{International conference on machine learning}}.
  PMLR, \bibinfo{pages}{448--456}.
\newblock


\bibitem[\protect\citeauthoryear{Islam, Pan, Nguyen, and Rajan}{Islam
  et~al\mbox{.}}{2020}]%
        {repairing_DNNoverview}
\bibfield{author}{\bibinfo{person}{Md~Johirul Islam}, \bibinfo{person}{Rangeet
  Pan}, \bibinfo{person}{Giang Nguyen}, {and} \bibinfo{person}{Hridesh Rajan}.}
  \bibinfo{year}{2020}\natexlab{}.
\newblock \showarticletitle{Repairing Deep Neural Networks: Fix Patterns and
  Challenges}. In \bibinfo{booktitle}{\emph{2020 IEEE/ACM 42nd International
  Conference on Software Engineering (ICSE)}}. \bibinfo{pages}{1135--1146}.
\newblock


\bibitem[\protect\citeauthoryear{Johnson and Khoshgoftaar}{Johnson and
  Khoshgoftaar}{2019}]%
        {oversampling19}
\bibfield{author}{\bibinfo{person}{J.M. Johnson} {and} \bibinfo{person}{T.M.
  Khoshgoftaar}.} \bibinfo{year}{2019}\natexlab{}.
\newblock \showarticletitle{Survey on deep learning with class imbalance}.
\newblock \bibinfo{journal}{\emph{J Big Data}}  \bibinfo{volume}{6}
  (\bibinfo{year}{2019}).
\newblock
\urldef\tempurl%
\url{https://doi.org/10.1186/s40537-019-0192-5}
\showDOI{\tempurl}


\bibitem[\protect\citeauthoryear{Katz, Barrett, Dill, Julian, and
  Kochenderfer}{Katz et~al\mbox{.}}{2017}]%
        {katz2017reluplex}
\bibfield{author}{\bibinfo{person}{Guy Katz}, \bibinfo{person}{Clark Barrett},
  \bibinfo{person}{David~L. Dill}, \bibinfo{person}{Kyle Julian}, {and}
  \bibinfo{person}{Mykel~J. Kochenderfer}.} \bibinfo{year}{2017}\natexlab{}.
\newblock \bibinfo{booktitle}{\emph{Reluplex: An Efficient SMT Solver for
  Verifying Deep Neural Networks}}.
\newblock \bibinfo{publisher}{Springer International Publishing},
  \bibinfo{address}{Cham}, \bibinfo{pages}{97--117}.
\newblock


\bibitem[\protect\citeauthoryear{Kim, Feldt, and Yoo}{Kim
  et~al\mbox{.}}{2019}]%
        {kim2019guiding}
\bibfield{author}{\bibinfo{person}{Jinhan Kim}, \bibinfo{person}{Robert Feldt},
  {and} \bibinfo{person}{Shin Yoo}.} \bibinfo{year}{2019}\natexlab{}.
\newblock \showarticletitle{Guiding deep learning system testing using surprise
  adequacy}. In \bibinfo{booktitle}{\emph{Proceedings of the 41st International
  Conference on Software Engineering}}. IEEE Press,
  \bibinfo{pages}{1039--1049}.
\newblock


\bibitem[\protect\citeauthoryear{Kim, Reingold, and Rothblum}{Kim
  et~al\mbox{.}}{2018}]%
        {Kim}
\bibfield{author}{\bibinfo{person}{Michael~P. Kim}, \bibinfo{person}{Omer
  Reingold}, {and} \bibinfo{person}{Guy~N. Rothblum}.}
  \bibinfo{year}{2018}\natexlab{}.
\newblock \showarticletitle{Fairness Through Computationally-Bounded
  Awareness}.
\newblock \bibinfo{journal}{\emph{32nd Conference on Neural Information
  Processing Systems (NeurIPS 2018)}} (\bibinfo{year}{2018}).
\newblock


\bibitem[\protect\citeauthoryear{Krizhevsky}{Krizhevsky}{2012}]%
        {cifar}
\bibfield{author}{\bibinfo{person}{Alex Krizhevsky}.}
  \bibinfo{year}{2012}\natexlab{}.
\newblock \showarticletitle{Learning Multiple Layers of Features from Tiny
  Images}.
\newblock \bibinfo{journal}{\emph{University of Toronto}} (\bibinfo{date}{05}
  \bibinfo{year}{2012}).
\newblock


\bibitem[\protect\citeauthoryear{Kusner, Loftus, Russell, and Silva}{Kusner
  et~al\mbox{.}}{2017}]%
        {Kusner}
\bibfield{author}{\bibinfo{person}{Matt~J Kusner}, \bibinfo{person}{Joshua
  Loftus}, \bibinfo{person}{Chris Russell}, {and} \bibinfo{person}{Ricardo
  Silva}.} \bibinfo{year}{2017}\natexlab{}.
\newblock \showarticletitle{Counterfactual Fairness}. In
  \bibinfo{booktitle}{\emph{Advances in Neural Information Processing Systems
  30}}. \bibinfo{pages}{4066--4076}.
\newblock


\bibitem[\protect\citeauthoryear{Lamy, Zhong, Menon, and Verma}{Lamy
  et~al\mbox{.}}{2019}]%
        {DBLP:journals/corr/abs-1901-10837}
\bibfield{author}{\bibinfo{person}{Alexandre~Louis Lamy},
  \bibinfo{person}{Ziyuan Zhong}, \bibinfo{person}{Aditya~Krishna Menon}, {and}
  \bibinfo{person}{Nakul Verma}.} \bibinfo{year}{2019}\natexlab{}.
\newblock \showarticletitle{Noise-tolerant fair classification}.
\newblock \bibinfo{journal}{\emph{CoRR}}  \bibinfo{volume}{abs/1901.10837}
  (\bibinfo{year}{2019}).
\newblock
\showeprint[arxiv]{1901.10837}
\urldef\tempurl%
\url{http://arxiv.org/abs/1901.10837}
\showURL{%
\tempurl}


\bibitem[\protect\citeauthoryear{Le~Goues, Dewey-Vogt, Forrest, and
  Weimer}{Le~Goues et~al\mbox{.}}{2012}]%
        {le2012systematic}
\bibfield{author}{\bibinfo{person}{Claire Le~Goues}, \bibinfo{person}{Michael
  Dewey-Vogt}, \bibinfo{person}{Stephanie Forrest}, {and}
  \bibinfo{person}{Westley Weimer}.} \bibinfo{year}{2012}\natexlab{}.
\newblock \showarticletitle{A systematic study of automated program repair:
  Fixing 55 out of 105 bugs for \$8 each}. In \bibinfo{booktitle}{\emph{2012
  34th International Conference on Software Engineering (ICSE)}}. IEEE,
  \bibinfo{pages}{3--13}.
\newblock


\bibitem[\protect\citeauthoryear{Le~Goues, Nguyen, Forrest, and
  Weimer}{Le~Goues et~al\mbox{.}}{2011}]%
        {le2011genprog}
\bibfield{author}{\bibinfo{person}{Claire Le~Goues}, \bibinfo{person}{ThanhVu
  Nguyen}, \bibinfo{person}{Stephanie Forrest}, {and} \bibinfo{person}{Westley
  Weimer}.} \bibinfo{year}{2011}\natexlab{}.
\newblock \showarticletitle{Genprog: A generic method for automatic software
  repair}.
\newblock \bibinfo{journal}{\emph{Ieee transactions on software engineering}}
  \bibinfo{volume}{38}, \bibinfo{number}{1} (\bibinfo{year}{2011}),
  \bibinfo{pages}{54--72}.
\newblock


\bibitem[\protect\citeauthoryear{Lin, Maire, Belongie, Hays, Perona, Ramanan,
  Doll{\'a}r, and Zitnick}{Lin et~al\mbox{.}}{2014}]%
        {lin2014microsoft}
\bibfield{author}{\bibinfo{person}{Tsung-Yi Lin}, \bibinfo{person}{Michael
  Maire}, \bibinfo{person}{Serge Belongie}, \bibinfo{person}{James Hays},
  \bibinfo{person}{Pietro Perona}, \bibinfo{person}{Deva Ramanan},
  \bibinfo{person}{Piotr Doll{\'a}r}, {and} \bibinfo{person}{C~Lawrence
  Zitnick}.} \bibinfo{year}{2014}\natexlab{}.
\newblock \showarticletitle{Microsoft coco: Common objects in context}. In
  \bibinfo{booktitle}{\emph{European conference on computer vision}}. Springer,
  \bibinfo{pages}{740--755}.
\newblock


\bibitem[\protect\citeauthoryear{Logozzo and Ball}{Logozzo and Ball}{2012}]%
        {logozzo2012modular}
\bibfield{author}{\bibinfo{person}{Francesco Logozzo} {and}
  \bibinfo{person}{Thomas Ball}.} \bibinfo{year}{2012}\natexlab{}.
\newblock \showarticletitle{{Modular and Verified Automatic Program Repair}}.
  In \bibinfo{booktitle}{\emph{Proceedings of the ACM International Conference
  on Object Oriented Programming Systems Languages and Applications}} (Tucson,
  Arizona, USA) \emph{(\bibinfo{series}{OOPSLA '12})}.
  \bibinfo{publisher}{ACM}, \bibinfo{address}{New York, NY, USA},
  \bibinfo{pages}{133--146}.
\newblock
\showISBNx{978-1-4503-1561-6}
\urldef\tempurl%
\url{https://doi.org/10.1145/2384616.2384626}
\showDOI{\tempurl}


\bibitem[\protect\citeauthoryear{Logozzo and Martel}{Logozzo and
  Martel}{2013}]%
        {logozzo2013automatic}
\bibfield{author}{\bibinfo{person}{Francesco Logozzo} {and}
  \bibinfo{person}{Matthieu Martel}.} \bibinfo{year}{2013}\natexlab{}.
\newblock \showarticletitle{Automatic repair of overflowing expressions with
  abstract interpretation}.
\newblock \bibinfo{journal}{\emph{arXiv preprint arXiv:1309.5148}}
  (\bibinfo{year}{2013}).
\newblock


\bibitem[\protect\citeauthoryear{Luong, Ruggieri, and Turini}{Luong
  et~al\mbox{.}}{2011}]%
        {luong2011k}
\bibfield{author}{\bibinfo{person}{Binh~Thanh Luong},
  \bibinfo{person}{Salvatore Ruggieri}, {and} \bibinfo{person}{Franco Turini}.}
  \bibinfo{year}{2011}\natexlab{}.
\newblock \showarticletitle{k-NN as an implementation of situation testing for
  discrimination discovery and prevention}. In
  \bibinfo{booktitle}{\emph{Proceedings of the 17th ACM SIGKDD international
  conference on Knowledge discovery and data mining}}. ACM,
  \bibinfo{pages}{502--510}.
\newblock


\bibitem[\protect\citeauthoryear{Ma, Juefei-Xu, Xue, Li, Li, Liu, and Zhao}{Ma
  et~al\mbox{.}}{2019}]%
        {ma2019deepct}
\bibfield{author}{\bibinfo{person}{Lei Ma}, \bibinfo{person}{Felix Juefei-Xu},
  \bibinfo{person}{Minhui Xue}, \bibinfo{person}{Bo Li}, \bibinfo{person}{Li
  Li}, \bibinfo{person}{Yang Liu}, {and} \bibinfo{person}{Jianjun Zhao}.}
  \bibinfo{year}{2019}\natexlab{}.
\newblock \showarticletitle{Deepct: Tomographic combinatorial testing for deep
  learning systems}. In \bibinfo{booktitle}{\emph{2019 IEEE 26th International
  Conference on Software Analysis, Evolution and Reengineering (SANER)}}. IEEE,
  \bibinfo{pages}{614--618}.
\newblock


\bibitem[\protect\citeauthoryear{Ma, Juefei-Xu, Zhang, Sun, Xue, Li, Chen, Su,
  Li, Liu, Zhao, and Wang}{Ma et~al\mbox{.}}{2018a}]%
        {ma2018deepgauge}
\bibfield{author}{\bibinfo{person}{Lei Ma}, \bibinfo{person}{Felix Juefei-Xu},
  \bibinfo{person}{Fuyuan Zhang}, \bibinfo{person}{Jiyuan Sun},
  \bibinfo{person}{Minhui Xue}, \bibinfo{person}{Bo Li},
  \bibinfo{person}{Chunyang Chen}, \bibinfo{person}{Ting Su},
  \bibinfo{person}{Li Li}, \bibinfo{person}{Yang Liu}, \bibinfo{person}{Jianjun
  Zhao}, {and} \bibinfo{person}{Yadong Wang}.}
  \bibinfo{year}{2018}\natexlab{a}.
\newblock \showarticletitle{DeepGauge: Multi-granularity Testing Criteria for
  Deep Learning Systems}.
\newblock  (\bibinfo{year}{2018}), \bibinfo{pages}{120--131}.
\newblock
\showISBNx{978-1-4503-5937-5}
\urldef\tempurl%
\url{https://doi.org/10.1145/3238147.3238202}
\showDOI{\tempurl}


\bibitem[\protect\citeauthoryear{Ma, Zhang, Sun, Xue, Li, Juefei-Xu, Xie, Li,
  Liu, Zhao, et~al\mbox{.}}{Ma et~al\mbox{.}}{2018d}]%
        {ma2018deepmutation}
\bibfield{author}{\bibinfo{person}{Lei Ma}, \bibinfo{person}{Fuyuan Zhang},
  \bibinfo{person}{Jiyuan Sun}, \bibinfo{person}{Minhui Xue},
  \bibinfo{person}{Bo Li}, \bibinfo{person}{Felix Juefei-Xu},
  \bibinfo{person}{Chao Xie}, \bibinfo{person}{Li Li}, \bibinfo{person}{Yang
  Liu}, \bibinfo{person}{Jianjun Zhao}, {et~al\mbox{.}}}
  \bibinfo{year}{2018}\natexlab{d}.
\newblock \showarticletitle{Deepmutation: Mutation testing of deep learning
  systems}. In \bibinfo{booktitle}{\emph{2018 IEEE 29th International Symposium
  on Software Reliability Engineering (ISSRE)}}. IEEE,
  \bibinfo{pages}{100--111}.
\newblock


\bibitem[\protect\citeauthoryear{Ma, Liu, Lee, Zhang, and Grama}{Ma
  et~al\mbox{.}}{2018c}]%
        {mode2018}
\bibfield{author}{\bibinfo{person}{Shiqing Ma}, \bibinfo{person}{Yingqi Liu},
  \bibinfo{person}{Wen-Chuan Lee}, \bibinfo{person}{Xiangyu Zhang}, {and}
  \bibinfo{person}{Ananth Grama}.} \bibinfo{year}{2018}\natexlab{c}.
\newblock \showarticletitle{MODE: Automated Neural Network Model Debugging via
  State Differential Analysis and Input Selection}. In
  \bibinfo{booktitle}{\emph{Proceedings of the 2018 26th ACM Joint Meeting on
  European Software Engineering Conference and Symposium on the Foundations of
  Software Engineering}} (Lake Buena Vista, FL, USA)
  \emph{(\bibinfo{series}{ESEC/FSE 2018})}. \bibinfo{publisher}{ACM},
  \bibinfo{address}{New York, NY, USA}, \bibinfo{pages}{175--186}.
\newblock
\showISBNx{978-1-4503-5573-5}
\urldef\tempurl%
\url{https://doi.org/10.1145/3236024.3236082}
\showDOI{\tempurl}


\bibitem[\protect\citeauthoryear{Ma, Li, Wang, Erfani, Wijewickrema,
  Schoenebeck, Song, Houle, and Bailey}{Ma et~al\mbox{.}}{2018b}]%
        {ma2018characterizing}
\bibfield{author}{\bibinfo{person}{Xingjun Ma}, \bibinfo{person}{Bo Li},
  \bibinfo{person}{Yisen Wang}, \bibinfo{person}{Sarah~M Erfani},
  \bibinfo{person}{Sudanthi Wijewickrema}, \bibinfo{person}{Grant Schoenebeck},
  \bibinfo{person}{Dawn Song}, \bibinfo{person}{Michael~E Houle}, {and}
  \bibinfo{person}{James Bailey}.} \bibinfo{year}{2018}\natexlab{b}.
\newblock \showarticletitle{Characterizing adversarial subspaces using local
  intrinsic dimensionality}. In \bibinfo{booktitle}{\emph{International
  Conference on Learning Representations (ICLR)}}.
\newblock


\bibitem[\protect\citeauthoryear{Madry, Makelov, Schmidt, Tsipras, and
  Vladu}{Madry et~al\mbox{.}}{2018}]%
        {madry2018}
\bibfield{author}{\bibinfo{person}{A. Madry}, \bibinfo{person}{A. Makelov},
  \bibinfo{person}{L. Schmidt}, \bibinfo{person}{D. Tsipras}, {and}
  \bibinfo{person}{A. Vladu}.} \bibinfo{year}{2018}\natexlab{}.
\newblock \showarticletitle{Towards Deep Learning Models Resistant to
  Adversarial Attacks}. In \bibinfo{booktitle}{\emph{International Conference
  on Learning Representations (ICLR)}}.
\newblock


\bibitem[\protect\citeauthoryear{MalletsDarker}{MalletsDarker}{2018}]%
        {google_photo}
\bibfield{author}{\bibinfo{person}{MalletsDarker}.}
  \bibinfo{year}{2018}\natexlab{}.
\newblock \showarticletitle{I took a few shots at Lake Louise today and Google
  offered me this panorama}.
\newblock  (\bibinfo{year}{2018}).
\newblock
\urldef\tempurl%
\url{https://www.reddit.com/r/funny/comments/7r9ptc/i_took_a_few_shots_at_lake_louise_today_and/dsvv1nw/}
\showURL{%
\tempurl}


\bibitem[\protect\citeauthoryear{Mao, Zhong, Yang, Vondrick, and Ray}{Mao
  et~al\mbox{.}}{2019}]%
        {metric_adv}
\bibfield{author}{\bibinfo{person}{Chengzhi Mao}, \bibinfo{person}{Ziyuan
  Zhong}, \bibinfo{person}{Junfeng Yang}, \bibinfo{person}{Carl Vondrick},
  {and} \bibinfo{person}{Baishakhi Ray}.} \bibinfo{year}{2019}\natexlab{}.
\newblock \showarticletitle{Metric Learning for Adversarial Robustness}.
\newblock In \bibinfo{booktitle}{\emph{Advances in Neural Information
  Processing Systems 32}}, \bibfield{editor}{\bibinfo{person}{H.~Wallach},
  \bibinfo{person}{H.~Larochelle}, \bibinfo{person}{A.~Beygelzimer},
  \bibinfo{person}{F.~d\textquotesingle Alch\'{e}-Buc},
  \bibinfo{person}{E.~Fox}, {and} \bibinfo{person}{R.~Garnett}} (Eds.).
  \bibinfo{publisher}{Curran Associates, Inc.}, \bibinfo{pages}{478--489}.
\newblock
\urldef\tempurl%
\url{http://papers.nips.cc/paper/8339-metric-learning-for-adversarial-robustness.pdf}
\showURL{%
\tempurl}


\bibitem[\protect\citeauthoryear{Mehrabi, Morstatter, Saxena, Lerman, and
  Galstyan}{Mehrabi et~al\mbox{.}}{2021}]%
        {fairsurvey21}
\bibfield{author}{\bibinfo{person}{Ninareh Mehrabi}, \bibinfo{person}{Fred
  Morstatter}, \bibinfo{person}{Nripsuta Saxena}, \bibinfo{person}{Kristina
  Lerman}, {and} \bibinfo{person}{Aram Galstyan}.}
  \bibinfo{year}{2021}\natexlab{}.
\newblock \showarticletitle{A Survey on Bias and Fairness in Machine Learning}.
\newblock \bibinfo{journal}{\emph{ACM Comput. Surv.}} \bibinfo{volume}{54},
  \bibinfo{number}{6}, Article \bibinfo{articleno}{115} (\bibinfo{date}{jul}
  \bibinfo{year}{2021}), \bibinfo{numpages}{35}~pages.
\newblock
\showISSN{0360-0300}
\urldef\tempurl%
\url{https://doi.org/10.1145/3457607}
\showDOI{\tempurl}


\bibitem[\protect\citeauthoryear{Menon and Williamson}{Menon and
  Williamson}{2018}]%
        {DBLP:conf/fat/MenonW18}
\bibfield{author}{\bibinfo{person}{Aditya~Krishna Menon} {and}
  \bibinfo{person}{Robert~C. Williamson}.} \bibinfo{year}{2018}\natexlab{}.
\newblock \showarticletitle{The cost of fairness in binary classification}. In
  \bibinfo{booktitle}{\emph{Conference on Fairness, Accountability and
  Transparency, {FAT} 2018, 23-24 February 2018, New York, NY, {USA}}}.
  \bibinfo{pages}{107--118}.
\newblock
\urldef\tempurl%
\url{http://proceedings.mlr.press/v81/menon18a.html}
\showURL{%
\tempurl}


\bibitem[\protect\citeauthoryear{Mirman, Gehr, and Vechev}{Mirman
  et~al\mbox{.}}{2018}]%
        {mirman2018differentiable}
\bibfield{author}{\bibinfo{person}{Matthew Mirman}, \bibinfo{person}{Timon
  Gehr}, {and} \bibinfo{person}{Martin Vechev}.}
  \bibinfo{year}{2018}\natexlab{}.
\newblock \showarticletitle{Differentiable abstract interpretation for provably
  robust neural networks}. In \bibinfo{booktitle}{\emph{International
  Conference on Machine Learning}}. \bibinfo{pages}{3575--3583}.
\newblock


\bibitem[\protect\citeauthoryear{Monperrus}{Monperrus}{2018}]%
        {monperrus2018automatic}
\bibfield{author}{\bibinfo{person}{Martin Monperrus}.}
  \bibinfo{year}{2018}\natexlab{}.
\newblock \showarticletitle{Automatic software repair: a bibliography}.
\newblock \bibinfo{journal}{\emph{ACM Computing Surveys (CSUR)}}
  \bibinfo{volume}{51}, \bibinfo{number}{1} (\bibinfo{year}{2018}),
  \bibinfo{pages}{1--24}.
\newblock


\bibitem[\protect\citeauthoryear{Muntean, Kommanapalli, Ibing, and
  Eckert}{Muntean et~al\mbox{.}}{2014}]%
        {muntean2014automated}
\bibfield{author}{\bibinfo{person}{Paul Muntean}, \bibinfo{person}{Vasantha
  Kommanapalli}, \bibinfo{person}{Andreas Ibing}, {and}
  \bibinfo{person}{Claudia Eckert}.} \bibinfo{year}{2014}\natexlab{}.
\newblock \showarticletitle{Automated generation of buffer overflow quick fixes
  using symbolic execution and SMT}. In \bibinfo{booktitle}{\emph{International
  Conference on Computer Safety, Reliability, and Security}}. Springer,
  \bibinfo{pages}{441--456}.
\newblock


\bibitem[\protect\citeauthoryear{Nguyen, Qi, Roychoudhury, and Chandra}{Nguyen
  et~al\mbox{.}}{2013}]%
        {nguyen2013semfix}
\bibfield{author}{\bibinfo{person}{Hoang Duong~Thien Nguyen},
  \bibinfo{person}{Dawei Qi}, \bibinfo{person}{Abhik Roychoudhury}, {and}
  \bibinfo{person}{Satish Chandra}.} \bibinfo{year}{2013}\natexlab{}.
\newblock \showarticletitle{Semfix: Program repair via semantic analysis}. In
  \bibinfo{booktitle}{\emph{2013 35th International Conference on Software
  Engineering (ICSE)}}. IEEE, \bibinfo{pages}{772--781}.
\newblock


\bibitem[\protect\citeauthoryear{Nica, Nica, and Wotawa}{Nica
  et~al\mbox{.}}{2013}]%
        {nica2013use}
\bibfield{author}{\bibinfo{person}{Mihai Nica}, \bibinfo{person}{Simona Nica},
  {and} \bibinfo{person}{Franz Wotawa}.} \bibinfo{year}{2013}\natexlab{}.
\newblock \showarticletitle{On the use of mutations and testing for debugging}.
\newblock \bibinfo{journal}{\emph{Software: practice and experience}}
  \bibinfo{volume}{43}, \bibinfo{number}{9} (\bibinfo{year}{2013}),
  \bibinfo{pages}{1121--1142}.
\newblock


\bibitem[\protect\citeauthoryear{Pei, Cao, Yang, and Jana}{Pei
  et~al\mbox{.}}{2017}]%
        {pei2017deepxplore}
\bibfield{author}{\bibinfo{person}{Kexin Pei}, \bibinfo{person}{Yinzhi Cao},
  \bibinfo{person}{Junfeng Yang}, {and} \bibinfo{person}{Suman Jana}.}
  \bibinfo{year}{2017}\natexlab{}.
\newblock \showarticletitle{Deepxplore: Automated whitebox testing of deep
  learning systems}. In \bibinfo{booktitle}{\emph{proceedings of the 26th
  Symposium on Operating Systems Principles}}. \bibinfo{pages}{1--18}.
\newblock


\bibitem[\protect\citeauthoryear{Ren, Yu, Qi, Juefei-Xu, Li, Xue, Ma, and
  Zhao}{Ren et~al\mbox{.}}{2020a}]%
        {ren20}
\bibfield{author}{\bibinfo{person}{Xuhong Ren}, \bibinfo{person}{Bing Yu},
  \bibinfo{person}{Hua Qi}, \bibinfo{person}{Felix Juefei-Xu},
  \bibinfo{person}{Zhuo Li}, \bibinfo{person}{Wanli Xue}, \bibinfo{person}{Lei
  Ma}, {and} \bibinfo{person}{Jianjun Zhao}.} \bibinfo{year}{2020}\natexlab{a}.
\newblock \showarticletitle{Few-Shot Guided Mix for DNN Repairing}. In
  \bibinfo{booktitle}{\emph{2020 IEEE International Conference on Software
  Maintenance and Evolution (ICSME)}}. \bibinfo{pages}{717--721}.
\newblock
\urldef\tempurl%
\url{https://doi.org/10.1109/ICSME46990.2020.00079}
\showDOI{\tempurl}


\bibitem[\protect\citeauthoryear{Ren, Yu, Qi, Juefei-Xu, Li, Xue, Ma, and
  Zhao}{Ren et~al\mbox{.}}{2020b}]%
        {ren2020few}
\bibfield{author}{\bibinfo{person}{Xuhong Ren}, \bibinfo{person}{Bing Yu},
  \bibinfo{person}{Hua Qi}, \bibinfo{person}{Felix Juefei-Xu},
  \bibinfo{person}{Zhuo Li}, \bibinfo{person}{Wanli Xue}, \bibinfo{person}{Lei
  Ma}, {and} \bibinfo{person}{Jianjun Zhao}.} \bibinfo{year}{2020}\natexlab{b}.
\newblock \showarticletitle{Few-Shot Guided Mix for DNN Repairing}. In
  \bibinfo{booktitle}{\emph{2020 IEEE International Conference on Software
  Maintenance and Evolution (ICSME)}}. IEEE, \bibinfo{pages}{717--721}.
\newblock


\bibitem[\protect\citeauthoryear{Sandler, Howard, Zhu, Zhmoginov, and
  Chen}{Sandler et~al\mbox{.}}{2018}]%
        {mobilenetv2}
\bibfield{author}{\bibinfo{person}{Mark Sandler}, \bibinfo{person}{Andrew
  Howard}, \bibinfo{person}{Menglong Zhu}, \bibinfo{person}{Andrey Zhmoginov},
  {and} \bibinfo{person}{Liang-Chieh Chen}.} \bibinfo{year}{2018}\natexlab{}.
\newblock \showarticletitle{MobileNetV2: Inverted Residuals and Linear
  Bottlenecks}. In \bibinfo{booktitle}{\emph{2018 IEEE/CVF Conference on
  Computer Vision and Pattern Recognition}}. \bibinfo{pages}{4510--4520}.
\newblock
\urldef\tempurl%
\url{https://doi.org/10.1109/CVPR.2018.00474}
\showDOI{\tempurl}


\bibitem[\protect\citeauthoryear{Santurkar, Tsipras, Ilyas, and
  Madry}{Santurkar et~al\mbox{.}}{2018}]%
        {batchnorm18}
\bibfield{author}{\bibinfo{person}{Shibani Santurkar},
  \bibinfo{person}{Dimitris Tsipras}, \bibinfo{person}{Andrew Ilyas}, {and}
  \bibinfo{person}{Aleksander Madry}.} \bibinfo{year}{2018}\natexlab{}.
\newblock \showarticletitle{How Does Batch Normalization Help Optimization?}.
  In \bibinfo{booktitle}{\emph{Advances in Neural Information Processing
  Systems}}, \bibfield{editor}{\bibinfo{person}{S.~Bengio},
  \bibinfo{person}{H.~Wallach}, \bibinfo{person}{H.~Larochelle},
  \bibinfo{person}{K.~Grauman}, \bibinfo{person}{N.~Cesa-Bianchi}, {and}
  \bibinfo{person}{R.~Garnett}} (Eds.), Vol.~\bibinfo{volume}{31}.
  \bibinfo{publisher}{Curran Associates, Inc.}
\newblock
\urldef\tempurl%
\url{https://proceedings.neurips.cc/paper/2018/file/905056c1ac1dad141560467e0a99e1cf-Paper.pdf}
\showURL{%
\tempurl}


\bibitem[\protect\citeauthoryear{Schulte, Forrest, and Weimer}{Schulte
  et~al\mbox{.}}{2010}]%
        {schulte2010automated}
\bibfield{author}{\bibinfo{person}{Eric Schulte}, \bibinfo{person}{Stephanie
  Forrest}, {and} \bibinfo{person}{Westley Weimer}.}
  \bibinfo{year}{2010}\natexlab{}.
\newblock \showarticletitle{Automated program repair through the evolution of
  assembly code}. In \bibinfo{booktitle}{\emph{Proceedings of the IEEE/ACM
  international conference on Automated software engineering}}.
  \bibinfo{pages}{313--316}.
\newblock


\bibitem[\protect\citeauthoryear{Simonyan and Zisserman}{Simonyan and
  Zisserman}{2015}]%
        {simonyan2014very}
\bibfield{author}{\bibinfo{person}{Karen Simonyan} {and}
  \bibinfo{person}{Andrew Zisserman}.} \bibinfo{year}{2015}\natexlab{}.
\newblock \showarticletitle{Very deep convolutional networks for large-scale
  image recognition}. In \bibinfo{booktitle}{\emph{International Conference on
  Learning Representations (ICLR)}}.
\newblock


\bibitem[\protect\citeauthoryear{Sohn, Kang, and Yoo}{Sohn
  et~al\mbox{.}}{2019}]%
        {sohn2019search}
\bibfield{author}{\bibinfo{person}{Jeongju Sohn}, \bibinfo{person}{Sungmin
  Kang}, {and} \bibinfo{person}{Shin Yoo}.} \bibinfo{year}{2019}\natexlab{}.
\newblock \showarticletitle{Search based repair of deep neural networks}.
\newblock \bibinfo{journal}{\emph{arXiv preprint arXiv:1912.12463}}
  (\bibinfo{year}{2019}).
\newblock


\bibitem[\protect\citeauthoryear{Tian, Pei, Jana, and Ray}{Tian
  et~al\mbox{.}}{2018a}]%
        {tian2017deeptest}
\bibfield{author}{\bibinfo{person}{Yuchi Tian}, \bibinfo{person}{Kexin Pei},
  \bibinfo{person}{Suman Jana}, {and} \bibinfo{person}{Baishakhi Ray}.}
  \bibinfo{year}{2018}\natexlab{a}.
\newblock \showarticletitle{DeepTest: Automated testing of
  deep-neural-network-driven autonomous cars}. In
  \bibinfo{booktitle}{\emph{International Conference of Software Engineering
  (ICSE), 2018 IEEE conference on}}. IEEE.
\newblock


\bibitem[\protect\citeauthoryear{Tian, Pei, Jana, and Ray}{Tian
  et~al\mbox{.}}{2018b}]%
        {tian2018deeptest}
\bibfield{author}{\bibinfo{person}{Yuchi Tian}, \bibinfo{person}{Kexin Pei},
  \bibinfo{person}{Suman Jana}, {and} \bibinfo{person}{Baishakhi Ray}.}
  \bibinfo{year}{2018}\natexlab{b}.
\newblock \showarticletitle{Deeptest: Automated testing of
  deep-neural-network-driven autonomous cars}. In
  \bibinfo{booktitle}{\emph{Proceedings of the 40th international conference on
  software engineering}}. \bibinfo{pages}{303--314}.
\newblock


\bibitem[\protect\citeauthoryear{Tian and Ray}{Tian and Ray}{2017}]%
        {tian2017automatically}
\bibfield{author}{\bibinfo{person}{Yuchi Tian} {and} \bibinfo{person}{Baishakhi
  Ray}.} \bibinfo{year}{2017}\natexlab{}.
\newblock \showarticletitle{Automatically diagnosing and repairing error
  handling bugs in c}. In \bibinfo{booktitle}{\emph{Proceedings of the 2017
  11th Joint Meeting on Foundations of Software Engineering}}.
  \bibinfo{pages}{752--762}.
\newblock


\bibitem[\protect\citeauthoryear{Tian, Zhong, Ordonez, Kaiser, and Ray}{Tian
  et~al\mbox{.}}{2020}]%
        {tian2019deepinspect}
\bibfield{author}{\bibinfo{person}{Yuchi Tian}, \bibinfo{person}{Ziyuan Zhong},
  \bibinfo{person}{Vicente Ordonez}, \bibinfo{person}{Gail Kaiser}, {and}
  \bibinfo{person}{Baishakhi Ray}.} \bibinfo{year}{2020}\natexlab{}.
\newblock \showarticletitle{Testing DNN Image Classifier for Confusion \& Bias
  Errors}. In \bibinfo{booktitle}{\emph{International Conference of Software
  Engineering (ICSE)}}.
\newblock


\bibitem[\protect\citeauthoryear{Tram{\`e}r, Kurakin, Papernot, Goodfellow,
  Boneh, and McDaniel}{Tram{\`e}r et~al\mbox{.}}{2017}]%
        {tramer2017ensemble}
\bibfield{author}{\bibinfo{person}{Florian Tram{\`e}r}, \bibinfo{person}{Alexey
  Kurakin}, \bibinfo{person}{Nicolas Papernot}, \bibinfo{person}{Ian
  Goodfellow}, \bibinfo{person}{Dan Boneh}, {and} \bibinfo{person}{Patrick
  McDaniel}.} \bibinfo{year}{2017}\natexlab{}.
\newblock \showarticletitle{Ensemble adversarial training: Attacks and
  defenses}.
\newblock \bibinfo{journal}{\emph{arXiv preprint arXiv:1705.07204}}
  (\bibinfo{year}{2017}).
\newblock


\bibitem[\protect\citeauthoryear{Vargha and Delaney}{Vargha and
  Delaney}{2000}]%
        {VDtest}
\bibfield{author}{\bibinfo{person}{András Vargha} {and}
  \bibinfo{person}{Harold~D. Delaney}.} \bibinfo{year}{2000}\natexlab{}.
\newblock \showarticletitle{A Critique and Improvement of the CL Common
  Language Effect Size Statistics of McGraw and Wong}.
\newblock \bibinfo{journal}{\emph{Journal of Educational and Behavioral
  Statistics}} \bibinfo{volume}{25}, \bibinfo{number}{2}
  (\bibinfo{year}{2000}), \bibinfo{pages}{101--132}.
\newblock
\urldef\tempurl%
\url{https://doi.org/10.3102/10769986025002101}
\showDOI{\tempurl}
\showeprint{https://doi.org/10.3102/10769986025002101}


\bibitem[\protect\citeauthoryear{Wang, Chen, Abdou, and Jana}{Wang
  et~al\mbox{.}}{2018a}]%
        {wang2018mixtrain}
\bibfield{author}{\bibinfo{person}{Shiqi Wang}, \bibinfo{person}{Yizheng Chen},
  \bibinfo{person}{Ahmed Abdou}, {and} \bibinfo{person}{Suman Jana}.}
  \bibinfo{year}{2018}\natexlab{a}.
\newblock \showarticletitle{Mixtrain: Scalable training of formally robust
  neural networks}.
\newblock \bibinfo{journal}{\emph{arXiv preprint arXiv:1811.02625}}
  (\bibinfo{year}{2018}).
\newblock


\bibitem[\protect\citeauthoryear{Wang, Pei, Whitehouse, Yang, and Jana}{Wang
  et~al\mbox{.}}{2018b}]%
        {wang2018formal}
\bibfield{author}{\bibinfo{person}{Shiqi Wang}, \bibinfo{person}{Kexin Pei},
  \bibinfo{person}{Justin Whitehouse}, \bibinfo{person}{Junfeng Yang}, {and}
  \bibinfo{person}{Suman Jana}.} \bibinfo{year}{2018}\natexlab{b}.
\newblock \showarticletitle{Formal Security Analysis of Neural Networks using
  Symbolic Intervals}.
\newblock  (\bibinfo{year}{2018}).
\newblock


\bibitem[\protect\citeauthoryear{Weimer, Forrest, Le~Goues, and Nguyen}{Weimer
  et~al\mbox{.}}{2010}]%
        {weimer2010automatic}
\bibfield{author}{\bibinfo{person}{Westley Weimer}, \bibinfo{person}{Stephanie
  Forrest}, \bibinfo{person}{Claire Le~Goues}, {and} \bibinfo{person}{ThanhVu
  Nguyen}.} \bibinfo{year}{2010}\natexlab{}.
\newblock \showarticletitle{Automatic program repair with evolutionary
  computation}.
\newblock \bibinfo{journal}{\emph{Commun. ACM}} \bibinfo{volume}{53},
  \bibinfo{number}{5} (\bibinfo{year}{2010}), \bibinfo{pages}{109--116}.
\newblock


\bibitem[\protect\citeauthoryear{Weimer, Nguyen, Goues, and Forrest}{Weimer
  et~al\mbox{.}}{2009}]%
        {weimer2009automatically}
\bibfield{author}{\bibinfo{person}{W. Weimer}, \bibinfo{person}{T. Nguyen},
  \bibinfo{person}{C.~Le Goues}, {and} \bibinfo{person}{S. Forrest}.}
  \bibinfo{year}{2009}\natexlab{}.
\newblock \showarticletitle{Automatically finding patches using genetic
  programming}. In \bibinfo{booktitle}{\emph{ICSE}}. \bibinfo{pages}{364--374}.
\newblock


\bibitem[\protect\citeauthoryear{Wolpert and Macready}{Wolpert and
  Macready}{1997}]%
        {nofreelunch}
\bibfield{author}{\bibinfo{person}{D.H. Wolpert} {and} \bibinfo{person}{W.G.
  Macready}.} \bibinfo{year}{1997}\natexlab{}.
\newblock \showarticletitle{No free lunch theorems for optimization}.
\newblock \bibinfo{journal}{\emph{IEEE Transactions on Evolutionary
  Computation}} \bibinfo{volume}{1}, \bibinfo{number}{1}
  (\bibinfo{year}{1997}), \bibinfo{pages}{67--82}.
\newblock
\urldef\tempurl%
\url{https://doi.org/10.1109/4235.585893}
\showDOI{\tempurl}


\bibitem[\protect\citeauthoryear{Wong, Schmidt, Metzen, and Kolter}{Wong
  et~al\mbox{.}}{2018}]%
        {wong2018scaling}
\bibfield{author}{\bibinfo{person}{Eric Wong}, \bibinfo{person}{Frank Schmidt},
  \bibinfo{person}{Jan~Hendrik Metzen}, {and} \bibinfo{person}{J~Zico Kolter}.}
  \bibinfo{year}{2018}\natexlab{}.
\newblock \showarticletitle{Scaling provable adversarial defenses}. In
  \bibinfo{booktitle}{\emph{Advances in Neural Information Processing
  Systems}}. \bibinfo{pages}{8400--8409}.
\newblock


\bibitem[\protect\citeauthoryear{Xie, Ma, Juefei-Xu, Xue, Chen, Liu, Zhao, Li,
  Yin, and See}{Xie et~al\mbox{.}}{2019}]%
        {xie2019deephunter}
\bibfield{author}{\bibinfo{person}{Xiaofei Xie}, \bibinfo{person}{Lei Ma},
  \bibinfo{person}{Felix Juefei-Xu}, \bibinfo{person}{Minhui Xue},
  \bibinfo{person}{Hongxu Chen}, \bibinfo{person}{Yang Liu},
  \bibinfo{person}{Jianjun Zhao}, \bibinfo{person}{Bo Li},
  \bibinfo{person}{Jianxiong Yin}, {and} \bibinfo{person}{Simon See}.}
  \bibinfo{year}{2019}\natexlab{}.
\newblock \showarticletitle{DeepHunter: a coverage-guided fuzz testing
  framework for deep neural networks}. In \bibinfo{booktitle}{\emph{Proceedings
  of the 28th ACM SIGSOFT International Symposium on Software Testing and
  Analysis}}. \bibinfo{pages}{146--157}.
\newblock


\bibitem[\protect\citeauthoryear{Yang, Min, and Lee}{Yang
  et~al\mbox{.}}{2020}]%
        {yang2020applying}
\bibfield{author}{\bibinfo{person}{Geunseok Yang}, \bibinfo{person}{Kyeongsic
  Min}, {and} \bibinfo{person}{Byungjeong Lee}.}
  \bibinfo{year}{2020}\natexlab{}.
\newblock \showarticletitle{Applying deep learning algorithm to automatic bug
  localization and repair}. In \bibinfo{booktitle}{\emph{Proceedings of the
  35th Annual ACM Symposium on Applied Computing}}.
  \bibinfo{pages}{1634--1641}.
\newblock


\bibitem[\protect\citeauthoryear{Yun, Han, Oh, Chun, Choe, and Yoo}{Yun
  et~al\mbox{.}}{2019}]%
        {yun2019cutmix}
\bibfield{author}{\bibinfo{person}{Sangdoo Yun}, \bibinfo{person}{Dongyoon
  Han}, \bibinfo{person}{Seong~Joon Oh}, \bibinfo{person}{Sanghyuk Chun},
  \bibinfo{person}{Junsuk Choe}, {and} \bibinfo{person}{Youngjoon Yoo}.}
  \bibinfo{year}{2019}\natexlab{}.
\newblock \showarticletitle{CutMix: Regularization Strategy to Train Strong
  Classifiers with Localizable Features}. In
  \bibinfo{booktitle}{\emph{International Conference on Computer Vision
  (ICCV)}}.
\newblock


\bibitem[\protect\citeauthoryear{Zafar, Valera, Gomez~Rodriguez, and
  Gummadi}{Zafar et~al\mbox{.}}{2017a}]%
        {ZafarWWW}
\bibfield{author}{\bibinfo{person}{Muhammad~Bilal Zafar},
  \bibinfo{person}{Isabel Valera}, \bibinfo{person}{Manuel Gomez~Rodriguez},
  {and} \bibinfo{person}{Krishna~P. Gummadi}.}
  \bibinfo{year}{2017}\natexlab{a}.
\newblock \showarticletitle{Fairness Beyond Disparate Treatment \& Disparate
  Impact: Learning Classification Without Disparate Mistreatment}. In
  \bibinfo{booktitle}{\emph{Proceedings of the 26th International Conference on
  World Wide Web}} (Perth, Australia). \bibinfo{pages}{1171--1180}.
\newblock


\bibitem[\protect\citeauthoryear{Zafar, Valera, Gomez~Rodriguez, and
  Gummadi}{Zafar et~al\mbox{.}}{2017b}]%
        {zafar2017fairness}
\bibfield{author}{\bibinfo{person}{Muhammad~Bilal Zafar},
  \bibinfo{person}{Isabel Valera}, \bibinfo{person}{Manuel Gomez~Rodriguez},
  {and} \bibinfo{person}{Krishna~P Gummadi}.} \bibinfo{year}{2017}\natexlab{b}.
\newblock \showarticletitle{Fairness constraints: Mechanisms for fair
  classification}. In \bibinfo{booktitle}{\emph{Proceedings of the 20th
  International Conference on Artificial Intelligence and Statistics}}
  \emph{(\bibinfo{series}{(AISTATS) 2017}, Vol.~\bibinfo{volume}{54})}.
  \bibinfo{publisher}{JMLR}.
\newblock


\bibitem[\protect\citeauthoryear{Zemel, Wu, Swersky, Pitassi, and Dwork}{Zemel
  et~al\mbox{.}}{2013}]%
        {ZemelICML}
\bibfield{author}{\bibinfo{person}{Rich Zemel}, \bibinfo{person}{Yu Wu},
  \bibinfo{person}{Kevin Swersky}, \bibinfo{person}{Toni Pitassi}, {and}
  \bibinfo{person}{Cynthia Dwork}.} \bibinfo{year}{2013}\natexlab{}.
\newblock \showarticletitle{Learning Fair Representations}. In
  \bibinfo{booktitle}{\emph{Proceedings of the 30th International Conference on
  Machine Learning}}. \bibinfo{pages}{325--333}.
\newblock


\bibitem[\protect\citeauthoryear{Zhang and Chan}{Zhang and Chan}{2019}]%
        {zhang19}
\bibfield{author}{\bibinfo{person}{Hao Zhang} {and} \bibinfo{person}{W.K.
  Chan}.} \bibinfo{year}{2019}\natexlab{}.
\newblock \showarticletitle{Apricot: A Weight-Adaptation Approach to Fixing
  Deep Learning Models}. In \bibinfo{booktitle}{\emph{2019 34th IEEE/ACM
  International Conference on Automated Software Engineering (ASE)}}.
  \bibinfo{pages}{376--387}.
\newblock
\urldef\tempurl%
\url{https://doi.org/10.1109/ASE.2019.00043}
\showDOI{\tempurl}


\bibitem[\protect\citeauthoryear{Zhao, Wang, Yatskar, Ordonez, and Chang}{Zhao
  et~al\mbox{.}}{2017}]%
        {zhao2017men}
\bibfield{author}{\bibinfo{person}{Jieyu Zhao}, \bibinfo{person}{Tianlu Wang},
  \bibinfo{person}{Mark Yatskar}, \bibinfo{person}{Vicente Ordonez}, {and}
  \bibinfo{person}{Kai-Wei Chang}.} \bibinfo{year}{2017}\natexlab{}.
\newblock \showarticletitle{Men Also Like Shopping: Reducing Gender Bias
  Amplification using Corpus-level Constraints}. In
  \bibinfo{booktitle}{\emph{Proceedings of the 2017 Conference on Empirical
  Methods in Natural Language Processing}}. \bibinfo{pages}{2941--2951}.
\newblock


\bibitem[\protect\citeauthoryear{Zhong, Tian, and Ray}{Zhong
  et~al\mbox{.}}{2021}]%
        {zhong2021understanding}
\bibfield{author}{\bibinfo{person}{Ziyuan Zhong}, \bibinfo{person}{Yuchi Tian},
  {and} \bibinfo{person}{Baishakhi Ray}.} \bibinfo{year}{2021}\natexlab{}.
\newblock \showarticletitle{Understanding Local Robustness of Deep Neural
  Networks under Natural Variations}.
\newblock \bibinfo{journal}{\emph{Fundamental Approaches to Software
  Engineering}}  \bibinfo{volume}{12649} (\bibinfo{year}{2021}),
  \bibinfo{pages}{313}.
\newblock


\bibitem[\protect\citeauthoryear{Zhong~Z}{Zhong~Z}{2021}]%
        {deeprobust}
\bibfield{author}{\bibinfo{person}{Ray~B. Zhong~Z, Tian~Y}.}
  \bibinfo{year}{2021}\natexlab{}.
\newblock \showarticletitle{Understanding Local Robustness of Deep Neural
  Networks under Natural Variations}. In \bibinfo{booktitle}{\emph{Fundamental
  Approaches to Software Engineering}}.
\newblock


\end{thebibliography}

\appendix
\newpage
\section{Additional Results on Hyper-parameter selection}
\label{sec:hyperparam_tuning}
In this section, we provide the detailed results of our hyper-parameter search process. The numbers colored in blue represent being selected based on the first criteria and those colored in yellow are those additionally selected based on our second criteria. The details of the selection have been discussed in \Cref{{sec:hyperparam_selection}}. We then run all the selected settings for five runs with different random seeds and report the results in RQ1 and RQ2. The results presented here also reflect the influence of the choice of the hyper-parameters on each method. In particular, the smaller $\eta$ is, w-os and w-bn have lower confusion/bias and accuracy (or mean average precision). The influence of the hyper-parameters for other methods are also usually monotonic for accuracy. However, the influence on confusion/bias usually varies case by case.  

\begin{table}[h]
\begin{minipage}{.48\linewidth}
  \setlength\tabcolsep{1pt} 
  \centering
  \scriptsize
  \caption{\textbf{\small Results on \cifars and ResNet-18}\label{tab:cifar_resnet}}
  \begin{tabular}{l|l|l|l|l}
 \toprule
   &                          \multicolumn{2}{c|}{\textbf{confusion}}                                   & \multicolumn{2}{c|}{\textbf{bias}}                               \\
      &                          \multicolumn{2}{c|}{\textbf{model}}                                   & \multicolumn{2}{c|}{\textbf{model}}                               \\
\midrule
model              & acc                           & conf                          & acc                           & bias                          \\
\midrule
orig               & 0.8747                        & 0.096                         & 0.8747                        & 0.074                         \\
orig-ft            & 0.8775                        & 0.0978                        & 0.8763                        & 0.084                         \\
w-aug 0.1          & 0.8456                        & 0.096                         & 0.8335                        & 0.0655                        \\
w-aug 0.3          & 0.8731                        & 0.0985                        & 0.8683                        & 0.0675                        \\
w-aug 0.5          & 0.8757                        & 0.093                         & 0.8731                        & 0.0655                        \\
w-aug 0.7          & {\color[HTML]{4285F4} 0.8777} & {\color[HTML]{4285F4} 0.095}  & {\color[HTML]{4285F4} 0.8765} & {\color[HTML]{4285F4} 0.063}  \\
w-aug 0.9          & 0.877                         & 0.0975                        & 0.8765                        & 0.0675                        \\
\midrule
w-bn 0.1           & 0.7361                        & 0.041                         & 0.7472                        & 0.032                         \\
w-bn 0.3           & 0.7916                        & 0.1035                        & 0.7942                        & 0.1005                        \\
w-bn 0.5           & 0.8099                        & 0.097                         & 0.8317                        & 0.0505                        \\
w-bn 0.7           & {\color[HTML]{4285F4} 0.8485} & {\color[HTML]{4285F4} 0.068}  & {\color[HTML]{4285F4} 0.864}  & {\color[HTML]{4285F4} 0.0545} \\
w-bn 0.9           & {\color[HTML]{FBBC04} 0.8761} & {\color[HTML]{FBBC04} 0.083}  & {\color[HTML]{FBBC04} 0.8806} & {\color[HTML]{FBBC04} 0.066}  \\
\midrule
w-loss 0.1         & 0.4078                        & 0.3325                        & 0.51                          & 0.1135                        \\
w-loss 0.3         & 0.6848                        & 0.2165                        & 0.6775                        & 0.116                         \\
w-loss 0.5         & 0.7921                        & 0.158                         & 0.7619                        & 0.102                         \\
w-loss 0.7         & 0.8663                        & 0.1035                        & 0.8214                        & 0.072                         \\
w-loss 0.9         & {\color[HTML]{4285F4} 0.8767} & {\color[HTML]{4285F4} 0.1005} & {\color[HTML]{4285F4} 0.8716} & {\color[HTML]{4285F4} 0.06}   \\
\midrule
w-os 0.001         & 0.8056                        & 0.0195                        & 0.7675                        & 0.022                         \\
w-os 0.01          & 0.839                         & 0.043                         & {\color[HTML]{4285F4} 0.8181} & {\color[HTML]{4285F4} 0.0425} \\
w-os 0.1           & {\color[HTML]{4285F4} 0.8654} & {\color[HTML]{4285F4} 0.071}  & 0.8591                        & 0.0565                        \\
w-os 0.3           & 0.873                         & 0.083                         & 0.8711                        & 0.057                         \\
w-os 0.5           & 0.8741                        & 0.088                         & 0.8731                        & 0.058                         \\
w-os 0.7           & 0.8745                        & 0.093                         & 0.8741                        & 0.0615                        \\
w-os 0.9           & 0.8745                        & 0.093                         & {\color[HTML]{FBBC04} 0.8751} & {\color[HTML]{FBBC04} 0.063}  \\
\midrule
w-dbr 0.1          & 0.7036                        & 0.12                          & 0.5327                        & 0.463                         \\
w-dbr 0.3          & 0.8238                        & 0.0915                        & 0.7518                        & 0.427                         \\
w-dbr 0.5          & 0.8588                        & 0.084                         & 0.7845                        & 0.4285                        \\
w-dbr 0.7          & 0.8696                        & 0.0935                        & 0.8288                        & 0.2665                        \\
w-dbr 0.9          & {\color[HTML]{4285F4} 0.8781} & {\color[HTML]{4285F4} 0.089}  & {\color[HTML]{4285F4} 0.8719} & {\color[HTML]{4285F4} 0.0935} \\
  
    \bottomrule
    \end{tabular}%
 \end{minipage}
\begin{minipage}{.48\linewidth}
  \setlength\tabcolsep{1pt} 
  \centering
  \scriptsize
  \caption{\textbf{\small Results on \cifars and VGG-11 w/ BN}\label{tab:cifar_vgg}}
  \begin{tabular}{l|l|l|l|l}
 \toprule
   &                          \multicolumn{2}{c|}{\textbf{confusion}}                                   & \multicolumn{2}{c|}{\textbf{bias}}                               \\
      &                          \multicolumn{2}{c|}{\textbf{model}}                                   & \multicolumn{2}{c|}{\textbf{model}}                               \\
\midrule
model      & acc                           & conf                          & acc                           & bias                          \\
\midrule
orig       & 0.9197                        & 0.081                         & 0.9197                        & 0.0675                        \\
orig-ft    & 0.9204                        & 0.076                         & 0.9213                        & 0.059                         \\
\midrule
w-aug 0.1  & 0.9129                        & 0.063                         & 0.9156                        & 0.0535                        \\
w-aug 0.3  & 0.9161                        & 0.071                         & {\color[HTML]{4285F4} 0.9186} & {\color[HTML]{4285F4} 0.0495} \\
w-aug 0.5  & 0.9165                        & 0.0735                        & 0.917                         & 0.05                          \\
w-aug 0.7  & 0.9162                        & 0.077                         & 0.9171                        & 0.0575                        \\
w-aug 0.9  & {\color[HTML]{4285F4} 0.9193} & {\color[HTML]{4285F4} 0.0725} & 0.916                         & 0.0545                        \\
\midrule
w-bn 0.1   & 0.8168                        & 0.0015                        & 0.8314                        & 0.016                         \\
w-bn 0.3   & 0.8554                        & 0.0125                        & 0.8717                        & 0.022                         \\
w-bn 0.5   & 0.8868                        & 0.0305                        & 0.9004                        & 0.0335                        \\
w-bn 0.7   & {\color[HTML]{4285F4} 0.9068} & {\color[HTML]{4285F4} 0.0575} & 0.9098                        & 0.0395                        \\
w-bn 0.9   & 0.9157                        & 0.072                         & {\color[HTML]{4285F4} 0.9186} & {\color[HTML]{4285F4} 0.044}  \\
\midrule
w-loss 0.1 & 0.1                           & 0.5                           & 0.1                           & 0                             \\
w-loss 0.3 & 0.1903                        & 0.4835                        & 0.285                         & 0.1645                        \\
w-loss 0.5 & 0.5521                        & 0.36                          & 0.75                          & 0.1465                        \\
w-loss 0.7 & 0.836                         & 0.2455                        & 0.838                         & 0.1395                        \\
w-loss 0.9 & {\color[HTML]{4285F4} 0.9153} & {\color[HTML]{4285F4} 0.088}  & {\color[HTML]{4285F4} 0.9122} & {\color[HTML]{4285F4} 0.0615} \\
\midrule
w-os 0.001 & {\color[HTML]{4285F4} 0.9059} & {\color[HTML]{4285F4} 0.0525} & {\color[HTML]{4285F4} 0.9023} & {\color[HTML]{4285F4} 0.041}  \\
w-os 0.01  & 0.9148                        & 0.071                         & 0.9144                        & 0.0535                        \\
w-os 0.1   & 0.9179                        & 0.0765                        & 0.9179                        & 0.0565                        \\
w-os 0.3   & 0.919                         & 0.0795                        & 0.9192                        & 0.0565                        \\
w-os 0.5   & 0.9193                        & 0.08                          & 0.9196                        & 0.056                         \\
w-os 0.7   & 0.9196                        & 0.0805                        & {\color[HTML]{FBBC04} 0.9198} & {\color[HTML]{FBBC04} 0.056}  \\
w-os 0.9   & {\color[HTML]{FBBC04} 0.9197} & {\color[HTML]{FBBC04} 0.0805} & 0.9197                        & 0.0555                        \\
\midrule
w-dbr 0.1  & {\color[HTML]{4285F4} 0.7035} & {\color[HTML]{4285F4} 0.0585} & 0.6136                        & 0.443                         \\
w-dbr 0.3  & 0.8954                        & 0.0865                        & 0.8145                        & 0.4405                        \\
w-dbr 0.5  & 0.9109                        & 0.075                         & 0.8323                        & 0.4415                        \\
w-dbr 0.7  & 0.9147                        & 0.0745                        & 0.8717                        & 0.2105                        \\
w-dbr 0.9  & 0.9159                        & 0.081                         & {\color[HTML]{4285F4} 0.9171} & {\color[HTML]{4285F4} 0.0535} \\
  
    \bottomrule
    \end{tabular}%
 
 \end{minipage}
 
\end{table}

\begin{table}[h]
\begin{minipage}{.48\linewidth}
  \setlength\tabcolsep{1pt} 
  \centering
  \scriptsize
  \caption{\textbf{\small Results on \cifars and MobileNetv2}\label{tab:cifar_mobilenetv2}}
  \begin{tabular}{l|l|l|l|l}
 \toprule
   &                          \multicolumn{2}{c|}{\textbf{confusion}}                                   & \multicolumn{2}{c|}{\textbf{bias}}                               \\
      &                          \multicolumn{2}{c|}{\textbf{model}}                                   & \multicolumn{2}{c|}{\textbf{model}}                               \\
\midrule
model              & acc                           & conf                          & acc                           & bias                          \\
\midrule
orig        & 0.942                                       & 0.0565                                      & 0.942                                       & 0.0415                                      \\
orig-ft     & 0.9398                                      & 0.057                                       & 0.9398                                      & 0.0395                                      \\
w-aug 0.1   & 0.9302                                      & 0.057                                       & 0.929                                       & 0.04                                        \\
w-aug 0.3   & 0.9366                                      & 0.055                                       & 0.9373                                      & 0.04                                        \\
w-aug 0.5   & {\color[HTML]{4285F4} 0.9384}               & {\color[HTML]{4285F4} 0.0545}               & {\color[HTML]{4285F4} 0.9394}               & {\color[HTML]{4285F4} 0.04}                 \\
w-aug 0.7   & 0.9397                                      & 0.0595                                      & 0.9363                                      & 0.046                                       \\
w-aug 0.9   & 0.9383                                      & 0.054                                       & 0.94                                        & 0.0485                                      \\
\midrule
w-bn 0.1    & 0.7825                                      & 0.001                                       & 0.7906                                      & 0.0008                                      \\
w-bn 0.3    & 0.8319                                      & 0.0065                                      & 0.8598                                      & 0.017                                       \\
w-bn 0.5    & 0.8804                                      & 0.0175                                      & {\color[HTML]{4285F4} 0.8976}               & {\color[HTML]{4285F4} 0.0285}               \\
w-bn 0.7    & {\color[HTML]{4285F4} 0.9139}               & {\color[HTML]{4285F4} 0.039}                & 0.9305                                      & 0.0345                                      \\
w-bn 0.9    & {\color[HTML]{4285F4} 0.939}                & {\color[HTML]{4285F4} 0.053}                & 0.9385                                      & 0.0395                                      \\
\midrule
w-loss 0.1  & 0.1764                                      & 0.433                                       & {\color[HTML]{4285F4} 0.345}                & {\color[HTML]{4285F4} 0.032}                \\
w-loss 0.3  & 0.6606                                      & 0.2835                                      & 0.7258                                      & 0.124                                       \\
w-loss 0.5  & 0.8411                                      & 0.196                                       & 0.8295                                      & 0.105                                       \\
w-loss 0.7  & 0.8899                                      & 0.132                                       & 0.8791                                      & 0.101                                       \\
w-loss 0.9  & {\color[HTML]{4285F4} 0.932}                & {\color[HTML]{4285F4} 0.0755}               & 0.9316                                      & 0.051                                       \\
\midrule
w-os 0.0001 & 0.9073                                      & 0.027                                       & {\color[HTML]{4285F4} 0.8943}               & {\color[HTML]{4285F4} 0.024}                \\
w-os 0.001  & {\color[HTML]{4285F4} 0.9226}               & {\color[HTML]{4285F4} 0.036}                & {\color[HTML]{4285F4} 0.9171}               & {\color[HTML]{4285F4} 0.033}                \\
w-os 0.01   & 0.9339                                      & 0.0435                                      & 0.9317                                      & 0.034                                       \\
w-os 0.1    & 0.9412                                      & 0.052                                       & 0.9401                                      & 0.039                                       \\
w-os 0.3    & {\color[HTML]{FBBC04} 0.942}                & {\color[HTML]{FBBC04} 0.053}                & 0.9414                                      & 0.039                                       \\
w-os 0.5    & 0.9416                                      & 0.0545                                      & 0.9413                                      & 0.04                                        \\
w-os 0.7    & 0.9421                                      & 0.056                                       & {\color[HTML]{FBBC04} 0.9415}               & {\color[HTML]{FBBC04} 0.041}                \\
w-os 0.9    & {\color[HTML]{FBBC04} 0.9421}               & {\color[HTML]{FBBC04} 0.0565}               & 0.9417                                      & 0.0415                                      \\
\midrule
w-dbr 0.1   & {\color[HTML]{4285F4} 0.6594}               & {\color[HTML]{4285F4} 0.06}                 & 0.4402                                      & 0.4675                                      \\
w-dbr 0.3   & 0.9003                                      & 0.0665                                      & 0.7939                                      & 0.392                                       \\
w-dbr 0.5   & 0.9248                                      & 0.0625                                      & 0.8477                                      & 0.465                                       \\
w-dbr 0.7   & 0.9335                                      & 0.0615                                      & 0.8667                                      & 0.4005                                      \\
w-dbr 0.9   & {\color[HTML]{4285F4} 0.9422}               & {\color[HTML]{4285F4} 0.0535}               & {\color[HTML]{4285F4} 0.9381}               & {\color[HTML]{4285F4} 0.0625}               \\
  
    \bottomrule
    \end{tabular}%
 \end{minipage}
\begin{minipage}{.48\linewidth}
  \setlength\tabcolsep{1pt} 
  \centering
  \scriptsize
  \caption{\textbf{\small Results on \cifar and ResNet34}\label{tab:cifar100_resnet}}
  \begin{tabular}{l|l|l|l|l}
 \toprule
   &                          \multicolumn{2}{c|}{\textbf{confusion}}                                   & \multicolumn{2}{c|}{\textbf{bias}}                               \\
      &                          \multicolumn{2}{c|}{\textbf{model}}                                   & \multicolumn{2}{c|}{\textbf{model}}                               \\
\midrule
model      & acc                           & conf                          & acc                           & bias                          \\
\midrule
orig       & 0.6988                                            & 0.155                                             & 0.6988                                            & 0.07                                              \\
orig-ft    & 0.7092                                            & 0.13                                              & 0.7047                                            & 0.05                                              \\
w-aug 0.1  & 0.7021                                            & 0.17                                              & 0.701                                             & 0.095                                             \\
w-aug 0.3  & {\color[HTML]{4285F4} 0.707}                      & {\color[HTML]{4285F4} 0.13}                       & 0.705                                             & 0.1                                               \\
w-aug 0.5  & {\color[HTML]{4285F4} 0.7097}                     & {\color[HTML]{4285F4} 0.17}                       & {\color[HTML]{4285F4} 0.7081}                     & {\color[HTML]{4285F4} 0.075}                      \\
w-aug 0.7  & 0.7108                                            & 0.17                                              & {\color[HTML]{4285F4} 0.7061}                     & {\color[HTML]{4285F4} 0.07}                       \\
w-aug 0.9  & 0.7055                                            & 0.17                                              & 0.7054                                            & 0.085                                             \\
\midrule
w-bn 0.1   & 0.5438                                            & 0                                                 & 0.5513                                            & 0                                                 \\
w-bn 0.3   & 0.596                                             & 0                                                 & 0.5993                                            & 0                                                 \\
w-bn 0.5   & 0.6342                                            & 0.01                                              & {\color[HTML]{4285F4} 0.6385}                     & {\color[HTML]{4285F4} 0.015}                      \\
w-bn 0.7   & {\color[HTML]{4285F4} 0.6736}                     & {\color[HTML]{4285F4} 0.035}                      & {\color[HTML]{4285F4} 0.6784}                     & {\color[HTML]{4285F4} 0.035}                      \\
w-bn 0.9   & {\color[HTML]{4285F4} 0.7039}                     & {\color[HTML]{4285F4} 0.11}                       & {\color[HTML]{FBBC04} 0.705}                      & {\color[HTML]{FBBC04} 0.045}                      \\
\midrule
w-loss 0.1 & 0.0175                                            & 0.39                                              & {\color[HTML]{4285F4} 0.0227}                     & {\color[HTML]{4285F4} 0.06}                       \\
w-loss 0.3 & 0.144                                             & 0.375                                             & 0.1399                                            & 0.175                                             \\
w-loss 0.5 & 0.5747                                            & 0.15                                              & 0.5756                                            & 0.09                                              \\
w-loss 0.7 & 0.6161                                            & 0.16                                              & 0.6209                                            & 0.105                                             \\
w-loss 0.9 & {\color[HTML]{4285F4} 0.6767}                     & {\color[HTML]{4285F4} 0.14}                       & {\color[HTML]{4285F4} 0.671}                      & {\color[HTML]{4285F4} 0.07}                       \\
\midrule
w-os 0.01  & 0.6957                                            & 0.04                                              & {\color[HTML]{4285F4} 0.6938}                     & {\color[HTML]{4285F4} 0.045}                      \\
w-os 0.1   & {\color[HTML]{4285F4} 0.6976}                     & {\color[HTML]{4285F4} 0.105}                      & {\color[HTML]{4285F4} 0.6968}                     & {\color[HTML]{4285F4} 0.095}                      \\
w-os 0.3   & 0.6984                                            & 0.12                                              & 0.6988                                            & 0.085                                             \\
w-os 0.5   & 0.6984                                            & 0.15                                              & 0.6985                                            & 0.095                                             \\
w-os 0.7   & {\color[HTML]{FBBC04} 0.6987}                     & {\color[HTML]{FBBC04} 0.155}                      & 0.6991                                            & 0.085                                             \\
w-os 0.9   & 0.6987                                            & 0.155                                             & 0.699                                             & 0.075                                             \\
\midrule
w-dbr 0.1  & {\color[HTML]{FBBC04} 0.0208}                     & {\color[HTML]{FBBC04} 0.37}                       & {\color[HTML]{4285F4} 0.0643}                     & {\color[HTML]{4285F4} 0.035}                      \\
w-dbr 0.3  & 0.5157                                            & 0.125                                             & 0.5459                                            & 0.175                                             \\
w-dbr 0.5  & {\color[HTML]{4285F4} 0.5677}                     & {\color[HTML]{4285F4} 0.22}                       & 0.6347                                            & 0.14                                              \\
w-dbr 0.7  & 0.6695                                            & 0.21                                              & 0.6661                                            & 0.195                                             \\
w-dbr 0.9  & {\color[HTML]{4285F4} 0.6752}                     & {\color[HTML]{4285F4} 0.11}                       & 0.6758                                            & 0.125                                             \\
  
    \bottomrule
    \end{tabular}%
 
 \end{minipage}
 
\end{table}

\begin{table}[h]
\begin{minipage}{.48\linewidth}
  \setlength\tabcolsep{1pt} 
  \centering
  \scriptsize
  \caption{\textbf{\small Results on \coco and ResNet-50}\label{tab:coco_resnet}}
  \begin{tabular}{l|l|l|l|l}
 \toprule
   &                          \multicolumn{2}{c|}{\textbf{confusion}}                                   & \multicolumn{2}{c|}{\textbf{bias}}                               \\
      &                          \multicolumn{2}{c|}{\textbf{model}}                                   & \multicolumn{2}{c|}{\textbf{model}}                               \\
\midrule
model              & acc                           & conf                          & acc                           & bias                          \\
\midrule
orig       & 0.6604                        & 0.2381                        & 0.6604                        & 0.2366                        \\
orig-ft    & 0.6613                        & 0.2298                        & 0.6613                        & 0.2283                        \\
\midrule
w-aug 0.1  & 0.6536                        & 0.2983                        & 0.653                         & 0.3483                        \\
w-aug 0.3  & 0.6567                        & 0.3126                        & 0.6563                        & 0.3019                        \\
w-aug 0.5  & 0.6581                        & 0.2793                        & 0.6588                        & 0.2759                        \\
w-aug 0.7  & 0.6601                        & 0.2896                        & 0.6605                        & 0.2749                        \\
w-aug 0.9  & {\color[HTML]{4285F4} 0.6609} & {\color[HTML]{4285F4} 0.2687} & {\color[HTML]{4285F4} 0.6607} & {\color[HTML]{4285F4} 0.2456} \\
\midrule
w-bn 0.1   & 0.6559                        & 0.0119                        & 0.6569                        & 0.121                         \\
w-bn 0.3   & 0.658                         & 0.0266                        & {\color[HTML]{4285F4} 0.658}  & {\color[HTML]{4285F4} 0.1453} \\
w-bn 0.5   & 0.6597                        & 0.0549                        & 0.6588                        & 0.1933                        \\
w-bn 0.7   & 0.6607                        & 0.1266                        & 0.6593                        & 0.2244                        \\
w-bn 0.9   & {\color[HTML]{4285F4} 0.6612} & {\color[HTML]{4285F4} 0.1694} & 0.6594                        & 0.2692                        \\
\midrule
w-loss 0.1 & 0.661                         & 0.0201                        & 0.661                         & 0.0194                        \\
w-loss 0.3 & 0.661                         & 0.0261                        & 0.661                         & 0.0253                        \\
w-loss 0.5 & 0.661                         & 0.0357                        & 0.6611                        & 0.0347                        \\
w-loss 0.7 & 0.6611                        & 0.048                         & 0.6611                        & 0.0474                        \\
w-loss 0.9 & {\color[HTML]{4285F4} 0.6613} & {\color[HTML]{4285F4} 0.1246} & {\color[HTML]{4285F4} 0.6613} & {\color[HTML]{4285F4} 0.1238} \\
\midrule
w-os 0.1   & 0.6557                        & 0                             & 0.6557                        & 0                             \\
w-os 0.3   & 0.6582                        & 0                             & 0.6582                        & 0                             \\
w-os 0.5   & 0.6599                        & 0                             & 0.6599                        & 0                             \\
w-os 0.7   & {\color[HTML]{4285F4} 0.6602} & {\color[HTML]{4285F4} 0.1159} & {\color[HTML]{4285F4} 0.6602} & {\color[HTML]{4285F4} 0.1151} \\
w-os 0.9   & {\color[HTML]{FBBC04} 0.6604} & {\color[HTML]{FBBC04} 0.203}  & {\color[HTML]{FBBC04} 0.6604} & {\color[HTML]{FBBC04} 0.2015} \\
\midrule
w-dbr 0.1  & 0.6548                        & 0.4108                        & {\color[HTML]{4285F4} 0.6542} & {\color[HTML]{4285F4} 0.2171} \\
w-dbr 0.3  & 0.6567                        & 0.0978                        & 0.6563                        & 0.2729                        \\
w-dbr 0.5  & 0.6575                        & 0.0476                        & 0.6586                        & 0.3047                        \\
w-dbr 0.7  & 0.6597                        & 0.0126                        & 0.6594                        & 0.3369                        \\
w-dbr 0.9  & {\color[HTML]{4285F4} 0.6607} & {\color[HTML]{4285F4} 0.0077} & 0.66                          & 0.3347                   \\
  
    \bottomrule
    \end{tabular}%
 \end{minipage}
\begin{minipage}{.48\linewidth}
  \setlength\tabcolsep{1pt} 
  \centering
  \scriptsize
  \caption{\textbf{\small Results on \coco gender and ResNet-50}\label{tab:cocogender_resnet}}
  \begin{tabular}{l|l|l|l|l}
 \toprule
   &                          \multicolumn{2}{c|}{\textbf{confusion}}                                   & \multicolumn{2}{c|}{\textbf{bias}}                               \\
      &                          \multicolumn{2}{c|}{\textbf{model}}                                   & \multicolumn{2}{c|}{\textbf{model}}                               \\
\midrule
model      & acc                           & conf                          & acc                           & bias                          \\
\midrule
orig       & 0.6701                        & 0.0402                        & 0.6701                        & 0.2521                        \\
orig-ft    & 0.6709                        & 0.0382                        & 0.6708                        & 0.2614                        \\
\midrule
w-aug 0.1  & 0.6614                        & 0.1435                        & 0.6589                        & 0.3185                        \\
w-aug 0.3  & 0.6676                        & 0.0656                        & 0.666                         & 0.3045                        \\
w-aug 0.5  & 0.67                          & 0.0629                        & 0.6686                        & 0.2574                        \\
w-aug 0.7  & {\color[HTML]{4285F4} 0.6696} & {\color[HTML]{4285F4} 0.0594} & {\color[HTML]{4285F4} 0.6694} & {\color[HTML]{4285F4} 0.2631} \\
w-aug 0.9  & {\color[HTML]{4285F4} 0.6711} & {\color[HTML]{4285F4} 0.0422} & {\color[HTML]{4285F4} 0.6712} & {\color[HTML]{4285F4} 0.2541} \\
\midrule
w-bn 0.1   & 0.6662                        & 0                             & {\color[HTML]{4285F4} 0.6641} & {\color[HTML]{4285F4} 0.063}  \\
w-bn 0.3   & 0.6679                        & 0                             & 0.6658                        & 0.0781                        \\
w-bn 0.5   & 0.6691                        & 0.0025                        & 0.6671                        & 0.1031                        \\
w-bn 0.7   & {\color[HTML]{4285F4} 0.6699} & {\color[HTML]{4285F4} 0.0039} & 0.6681                        & 0.1419                        \\
w-bn 0.9   & {\color[HTML]{4285F4} 0.6705} & {\color[HTML]{4285F4} 0.0092} & {\color[HTML]{4285F4} 0.6686} & {\color[HTML]{4285F4} 0.1768} \\
\midrule
w-loss 0.1 & 0.6695                        & 0.0349                        & 0.6698                        & 0.0944                        \\
w-loss 0.3 & 0.6704                        & 0.019                         & 0.67                          & 0.0875                        \\
w-loss 0.5 & 0.6705                        & 0.0179                        & 0.6702                        & 0.1037                        \\
w-loss 0.7 & {\color[HTML]{4285F4} 0.6707} & {\color[HTML]{4285F4} 0.0203} & {\color[HTML]{4285F4} 0.6704} & {\color[HTML]{4285F4} 0.1223} \\
w-loss 0.9 & 0.6708                        & 0.0331                        & {\color[HTML]{4285F4} 0.6707} & {\color[HTML]{4285F4} 0.1501} \\
\midrule
w-os 0.1   & 0.6687                        & 0                             & 0.6657                        & 0                             \\
w-os 0.3   & 0.6692                        & 0                             & 0.668                         & 0                             \\
w-os 0.5   & {\color[HTML]{4285F4} 0.6695} & {\color[HTML]{4285F4} 0}      & {\color[HTML]{4285F4} 0.6693} & {\color[HTML]{4285F4} 0}      \\
w-os 0.7   & {\color[HTML]{4285F4} 0.6698} & {\color[HTML]{4285F4} 0.0044} & {\color[HTML]{FBBC04} 0.6699} & {\color[HTML]{FBBC04} 0.2122} \\
w-os 0.9   & 0.67                          & 0.0314                        & 0.67                          & 0.2368                        \\
\midrule
w-dbr 0.1  & 0.6663                        & 0.0476                        & 0.6673                        & 0.0277                        \\
w-dbr 0.3  & 0.668                         & 0.0654                        & {\color[HTML]{4285F4} 0.6678} & {\color[HTML]{4285F4} 0.1019} \\
w-dbr 0.5  & 0.6686                        & 0.0642                        & 0.6682                        & 0.2284                        \\
w-dbr 0.7  & {\color[HTML]{4285F4} 0.6692} & {\color[HTML]{4285F4} 0.0686} & 0.6688                        & 0.3267                        \\
w-dbr 0.9  & {\color[HTML]{4285F4} 0.6704} & {\color[HTML]{4285F4} 0.077}  & 0.6696                        & 0.3926                         \\
  
    \bottomrule
    \end{tabular}%
 
 \end{minipage}
 
\end{table}

\begin{table}[h]
\begin{minipage}{.48\linewidth}
  \setlength\tabcolsep{1pt} 
  \centering
  \scriptsize
  \caption{\textbf{\small Results on \cifars and ResNet-18 on two pairs}\label{tab:cifar_resnet_multipair}}
  \begin{tabular}{l|l|l}
 \toprule
   &                          \multicolumn{2}{c|}{\textbf{confusion}}                    \\
      &                          \multicolumn{2}{c|}{\textbf{model}}                     \\
\midrule
model              & acc                           & conf                \\
\midrule
orig       & 0.8747                        & 0.067                         \\
orig-ft    & 0.8771                        & 0.0698                        \\
\midrule
w-aug 0.1  & 0.8307                        & 0.0683                        \\
w-aug 0.3  & 0.8656                        & 0.0648                        \\
w-aug 0.5  & 0.8713                        & 0.071                         \\
w-aug 0.7  & {\color[HTML]{4285F4} 0.8769} & {\color[HTML]{4285F4} 0.0668} \\
w-aug 0.9  & 0.8783                        & 0.0683                        \\
\midrule
w-bn 0.1   & 0.7839                        & 0.03                          \\
w-bn 0.3   & 0.8192                        & 0.041                         \\
w-bn 0.5   & {\color[HTML]{4285F4} 0.8466} & {\color[HTML]{4285F4} 0.0498} \\
w-bn 0.7   & 0.8654                        & 0.0605                        \\
w-bn 0.9   & {\color[HTML]{FBBC04} 0.8765} & {\color[HTML]{FBBC04} 0.0663} \\
\midrule
w-loss 0.1 & 0.5628                        & 0.185                         \\
w-loss 0.3 & 0.7045                        & 0.1402                        \\
w-loss 0.5 & 0.8144                        & 0.093                         \\
w-loss 0.7 & 0.8591                        & 0.0735                        \\
w-loss 0.9 & {\color[HTML]{4285F4} 0.8739} & {\color[HTML]{4285F4} 0.0698} \\
\midrule
w-os 0.001 & 0.7788                        & 0.02125                       \\
w-os 0.01  & {\color[HTML]{4285F4} 0.8289} & {\color[HTML]{4285F4} 0.037}  \\
w-os 0.1   & 0.8628                        & 0.05275                       \\
w-os 0.3   & 0.8722                        & 0.05975                       \\
w-os 0.5   & 0.8735                        & 0.06275                       \\
w-os 0.7   & 0.8743                        & 0.065                         \\
w-os 0.9   & {\color[HTML]{FBBC04} 0.8749} & {\color[HTML]{FBBC04} 0.066}  \\
\midrule
w-dbr 0.1  & 0.7727                        & 0.0728                        \\
w-dbr 0.3  & 0.8498                        & 0.0602                        \\
w-dbr 0.5  & 0.8672                        & 0.0592                        \\
w-dbr 0.7  & 0.8739                        & 0.0633                        \\
w-dbr 0.9  & {\color[HTML]{4285F4} 0.8792} & {\color[HTML]{4285F4} 0.0622}      \\
  
    \bottomrule
    \end{tabular}%
 \end{minipage}
\begin{minipage}{.48\linewidth}
  \setlength\tabcolsep{1pt} 
  \centering
  \scriptsize
  \caption{\textbf{\small Results on \coco and ResNet-50 on two pairs}\label{tab:coco_resnet_multipair}}
  \begin{tabular}{l|l|l}
 \toprule
   &                          \multicolumn{2}{c|}{\textbf{confusion}}            \\
      &                          \multicolumn{2}{c|}{\textbf{model}}                     \\
\midrule
model      & acc                           & conf                 \\
\midrule
orig       & 0.6604                                            & 0.2013                                            \\
orig-ft    & 0.6613                                            & 0.2298                                            \\
\midrule
w-aug 0.1  & 0.6521                                            & 0.2937                                            \\
w-aug 0.3  & 0.6564                                            & 0.2628                                            \\
w-aug 0.5  & 0.6586                                            & 0.2697                                            \\
w-aug 0.7  & {\color[HTML]{4285F4} 0.66}                       & {\color[HTML]{4285F4} 0.2484}                     \\
w-aug 0.9  & {\color[HTML]{4285F4} 0.6607}                     & {\color[HTML]{4285F4} 0.2141}                     \\
\midrule
w-bn 0.1   & 0.6548                                            & 0.0593                                            \\
w-bn 0.3   & 0.6572                                            & 0.0835                                            \\
w-bn 0.5   & {\color[HTML]{4285F4} 0.6592}                     & {\color[HTML]{4285F4} 0.0961}                     \\
w-bn 0.7   & {\color[HTML]{4285F4} 0.6607}                     & {\color[HTML]{4285F4} 0.1249}                     \\
w-bn 0.9   & {\color[HTML]{FBBC04} 0.6614}                     & {\color[HTML]{FBBC04} 0.175}                      \\
\midrule
w-loss 0.1 & 0.6605                                            & 0.0388                                            \\
w-loss 0.3 & 0.6605                                            & 0.0442                                            \\
w-loss 0.5 & 0.6608                                            & 0.0512                                            \\
w-loss 0.7 & 0.6611                                            & 0.0605                                            \\
w-loss 0.9 & {\color[HTML]{4285F4} 0.6615}                     & {\color[HTML]{4285F4} 0.1189}                     \\
\midrule
w-os 0.1   & 0.6477                                            & 0                                                 \\
w-os 0.3   & {\color[HTML]{4285F4} 0.6535}                     & {\color[HTML]{4285F4} 0}                          \\
w-os 0.5   & 0.6575                                            & 0                                                 \\
w-os 0.7   & {\color[HTML]{4285F4} 0.6595}                     & {\color[HTML]{4285F4} 0.0968}                     \\
w-os 0.9   & 0.6602                                            & 0.1725                                            \\
\midrule
w-dbr 0.1  & {\color[HTML]{FBBC04} 0.6552}                     & {\color[HTML]{FBBC04} 0.2957}                     \\
w-dbr 0.3  & 0.6571                                            & 0.2019                                            \\
w-dbr 0.5  & 0.6578                                            & 0.1828                                            \\
w-dbr 0.7  & 0.6583                                            & 0.1624                                            \\
w-dbr 0.9  & {\color[HTML]{4285F4} 0.6588}                     & {\color[HTML]{4285F4} 0.1607}                     \\
  
    \bottomrule
    \end{tabular}%
 
 \end{minipage}
 
\end{table}

\end{document}